\def\year{2020}\relax
\pdfoutput=1
\documentclass[letterpaper]{article} 
\usepackage{aaai20}  
\usepackage{times}  
\usepackage{helvet} 
\usepackage{courier}  
\usepackage[hyphens]{url}  
\usepackage{graphicx} 
\usepackage{graphicx}  
\usepackage{booktabs}
\usepackage{import}
\usepackage{amsmath}
\usepackage[toc]{appendix}
\DeclareMathOperator*{\argmax}{arg\,max}
\DeclareMathOperator*{\argmin}{arg\,min}

\title{WATTNet: Learning to Trade FX via Hierarchical Spatio-Temporal Representation of Highly Multivariate Time Series}

\author{Michael Poli\textsuperscript{\rm{1,2}}, Jinkyoo Park\textsuperscript{\rm 2}, Ilija Ilievski\textsuperscript{\rm 1} \\ 
\textsuperscript{\rm 1}Neuri Pte Ltd,
Singapore, Singapore\\ \textsuperscript{\rm 2}Department of Industrial \& Systems Engineering, KAIST, Daejeon, South Korea \\
michael.poli@neuri.ai, jinkyoo.park@kaist.ac.kr, ilija139@neuri.ai 
}

\begin{document}
\maketitle

\begin{abstract}
Finance is a particularly challenging application area for deep learning models due to low noise-to-signal ratio, non-stationarity, and partial observability. \textit{Non-deliverable-forwards} (NDF), a derivatives contract used in foreign exchange (FX) trading, presents additional difficulty in the form of long-term planning required for an effective selection of start and end date of the contract. In this work, we focus on tackling the problem of NDF tenor selection by leveraging high-dimensional sequential data consisting of spot rates, technical indicators and expert tenor patterns. To this end, we construct a dataset from the Depository Trust \& Clearing Corporation (DTCC) NDF data that includes a comprehensive list of NDF volumes and daily spot rates for 64 FX pairs. We introduce \textit{W}aveA\textit{TT}ention\textit{Net} (WATTNet), a novel temporal convolution (TCN) model for spatio-temporal modeling of highly multivariate time series, and validate it across NDF markets with varying degrees of dissimilarity between the training and test periods in terms of volatility and general market regimes. The proposed method achieves a significant positive return on investment (ROI) in all NDF markets under analysis, outperforming recurrent and classical baselines by a wide margin. Finally, we propose two orthogonal interpretability approaches to verify noise stability and detect the driving factors of the learned tenor selection strategy.

\end{abstract}

\section{Introduction}
\noindent Following recent trends of successful AI adoption, the financial world has seen a significant surge of attempts at leveraging deep learning and reinforcement learning techniques across various application areas. Slowing down progress in this field are the particular properties of financial data: low signal-to-noise ratio ~\cite{guhr2003new}, partial observability, and irregular sampling. Furthermore, AI breakthroughs in finance often go unpublished due to monetary incentives. Additional challenges are caused by the scarcity of datasets available, which are often limited in scope, difficult to acquire or for some application areas missing altogether.

As an attempt to alleviate some of these concerns, we release both a curated dataset and a novel model for foreign exchange (FX) futures trading. We focus our attention on a particular class of FX trading methods, \textit{non-deliverable-forward} (NDF) contracts, which constitute an important open problem in finance and can serve as a challenging benchmark for supervised or reinforcement learning models. We formulate the learning problem as an optimal selection problem in which the model is tasked with selecting the end date of the forward contract (tenor) from a rich input containing past human trade patterns as well as spot rates and technical indicators. In particular, tenor selection is cast into a \textit{direct imitation learning} ~\cite{judah2012active} framework, where the model learns policy $\pi$ directly from a set of execution trajectories of a demonstration policy $\pi^*$ without receiving a reward signal from the environment. The demonstrations are derived in a greedy fashion from spot rate data and the resulting input-output tuple is utilized to perform standard supervised learning. 

A key difference of our approach compared to existing FX trading algorithms lies in the type of data relied upon for learning, which includes expert tenor patterns in addition to standard technical indicators. Such patterns are extracted from a large dataset containing trades from competitive market players assumed to be informed about market state and to act rationally in order to achieve higher returns. Leveraging this additional information allows the models to differentiate between profitable and non-profitable market conditions with improved accuracy, ultimately leading to higher returns.

Fundamentally important for finance are models capable of capturing inter and intra-dependencies in highly multivariate time series. Many, if not most, of such interaction terms are nonlinear and thus challenging to analyze with standard statistical approaches. A direct consequence has been the new-found popularity of data-driven models for financial forecasting tasks, in particular recurrent neural networks (RNN) and their variants. Recurrent models, while offering an intuitive approach to time series modeling, lack an explicit module to capture inter-dependencies and perform relational reasoning ~\cite{santoro2018relational}. A different approach to time series modeling relies on temporal convolutions (TCN) as its fundamental computational block. Particularly successful in this area of research is WaveNet ~\cite{oord2016wavenet}, originally developed as a generative model for speech data. However, vanilla WaveNet and its derivative models are primarily designed to handle univariate time series and thus are ill-suited for highly multivariate financial time series. To bridge this gap, we introduce a new TCN model called \textit{WaveATTentionNet} (WATTNet) that incorporates computationally efficient dilated convolutions for temporal learning of autoregressive effects and self-attention modules to learn spatial, inter-time series interaction terms. 

We summarize our main contributions as follows:
\begin{itemize}
    \item We curate, analyze, and release a new dataset containing spot rates for 64 FX currencies, along with technical indicators and hourly frequency NDF contract trade data spanning the period from $2013$ to $2019$. Several models, including classical baselines (Momentum-1, Momentum-90) and recurrent baselines (GRUs, LSTMs) are evaluated against expert NDF data.
    \item We introduce \textit{WATTNet}, a novel \textit{temporal convolution} (TCN) architecture for spatio-temporal modeling. WATTNet is designed to extend WaveNet models to settings with highly multivariate time series data.
    \item We provide two orthogonal approaches to evaluate noise stability and explain driving factors of the learned trading strategy, along with examples to highlight their efficacy.
\end{itemize}

\section{Related Work and Background}

\paragraph{Deep Learning for FX trading}
Earlier attempts at utilizing the expressivity of neural networks in \textit{forex} (FX) trading have been carried out in ~\cite{chan1995enhancing}, which predicts technical indicators via shallow fully-connected neural networks. ~\cite{yu2005designing} designs a hybrid trading system capable of providing suggestions based on qualitative expert knowledge and price forecasting data obtained from a neural network. More recently ~\cite{czekalski2015ann}, ~\cite{galeshchuk2017deep} and ~\cite{petropoulos2017stacked} leverage fully-connected neural networks, CNNs and classical autoregressive modeling techniques.
However, these approaches focus on regular forex markets and short-term predictions and rely only on spot rates and technical indicators as informative features. Incorporating additional sources of data has been explored in  ~\cite{nassirtoussi2015text}, ~\cite{vargas2017deep} ~\cite{hu2018listening}, in which additional textual information obtained from financial news articles or online discussion is included in the input features.

While the literature has no shortage of works in which reinforcement learning is applied to portfolio management ~\cite{yu2019model} and optimal trade execution, the FX markets remain comparatively unexplored. ~\cite{carapucco2018reinforcement} develops a short-term spot trading system based on reinforcement learning and obtains positive ROI in the EURUSD market. ~\cite{sornmayura2019robust} offers an analysis of deep Q-learning (DQN) performance on two FX instruments. We are not aware of any published work where deep learning or reinforcement systems are introduced to tackle FX trading in an NDF setting. 

\paragraph{Spatio temporal modeling}
WaveNet ~\cite{oord2016wavenet} is an autoregressive model based on dilated temporal convolutions (TCN) in which the joint probability of the input sequence is modeled as a factorized product of conditional probabilities. SNAIL ~\cite{mishra2017simple} obtains improvements over vanilla WaveNet by adding a temporal attention layer between dilated convolutions. However, both vanilla WaveNet and SNAIL are originally designed to process univariate time series data and are thus unable to learn interaction terms between time series. ConvLSTM ~\cite{xingjian2015convolutional} introduce a convolution operation inside the LSTM cell to capture spatiotemporal information. A weakness of ConvLSTMs is given by the prior assumption of structure in the spatial domain where features closer together are prioritized by the convolution operation, as is the case for example with video data. In general applications, the time series are arbitrarily concatenated as input data and locality assumptions do not hold. Long-Short Term Network (LSTNet)~\cite{lai2018modeling} extracts local features in temporal and spatial domain with convolutions and adds a recurrent layer for longer-term dependencies. Similarly to ConvLSTM, LSTNet assumes spatial locality.
A more recent approach to spatio-temporal modeling based on Graph Neural Networks (GNNs) is Spatio-Temporal Graph Convolutional Network (STCGN) \cite{yu2018spatio} which utilizes graph convolution to carry out learning in both spatial and temporal dimensions.

\subsection{Background}
We briefly introduce the necessary background regarding different types of \textit{forex trading}.
\paragraph{Foreign Exchanges}
Trading in \textit{forex} (FX) markets is generally done via \textit{spot exchanges} or \textit{forward exchanges}, where \textit{spot rate} indicates the present expected buying rate. The \textit{spot market} can be volatile and is affected by news cycles, speculation, and underlying market dynamics. On the other hand, forward exchanges contain a long-term planning component: two parties fix a binding amount and date of exchange and the profits are calculated by comparing currency rates at the start date and fix date. The difference between start date and fix date is commonly referred to as \textit{tenor}.

\paragraph{Non-Deliverable-Forward}
An NDF operates similarly to \textit{forward exchange} contracts and exists as a replacement to forward FX trades in emerging markets. NDF markets are \textit{over-the-counter}, meaning they operate directly between involved parties without supervision, and are generally more volatile due to limited market-depth. In NDF trades the parties agree on notional amounts of primary and secondary currency (e.g. dollar USD and korean won KRW) which define the \textit{forward rate}. The currency amounts are not exchanged at the end of the contract: instead, NDF trades are cash-settled in USD, and the cash flow is computed as $R_{t,a} = (x_{t+a} - x_{t}) v_{t}$ where $x_t$ is the spot rate at time $t$, $a$ is the tenor and $v$ is the notional amount. Longer tenors are generally more profitable at the expense of a higher volatility, commonly referred to as \textit{risk-premia}. A successful trading agent thus has to find a difficult balance between risky, high return and safer, low return actions.
\section{NDF Dataset}
\paragraph{Notation}
A multivariate time series of length T and dimension M is indicated as $\{\mathbf{X}\}$. We use $\{\mathbf{x}_{i}\}$ for individual time series indexed by $i$. $x_{i,t}$ selects a scalar element of time series $i$ at time index $t$ . In particular, we indicate a slice across all time series at time $t$ with $\{x_{1,t} \dots x_{M,t}\}$.
Whenever the operations on $\{\mathbf{X}\}$ are batched, we add superscript $j$ for single samples in the batch: $\{\mathbf{X}^j\}$. With batch size $N$, the resulting tensor $\{\mathbf{X}^1\} \dots \{\mathbf{X}^N\}$ has dimensions $N \times T \times M$. We refer to tenors as $a$ and to the set of admissible tenor choices as $\mathcal{A}$.

\begin{table}
\centering
\setlength\tabcolsep{3pt}
\begin{tabular}[h!]{lr}  
\toprule
NDF contracts & 7,580,814  \\
Trading hours & 35,712  \\
Trading days & 1,488  \\
Number of features per trade hour & 1,123  \\
Number of FX spot rates & 64 \\
\bottomrule
\end{tabular}
\caption{Summary of dataset statistics. Additional details included in Appendix A.}
\label{tab:datastats}
\end{table}

\paragraph{Expert benchmarks} The NDF trade records have been collected from the The Depository Trust \& Clearning Corporation (DTCC) database. These records contain start and end dates of each NDF contract, along with currency amounts. For each trading day $t$ and admissible tenor $a \in \mathcal{A}$ we obtain trading volumes $v_{t,a}$. We refer to \textit{Expert} as a trading agent that chooses tenors corresponding to maximum volumes $a_t = \argmax_{a \in \mathcal{A}}{v_{t,a}}$. In addition to \textit{Expert} we obtain a fictitious agent based on NDF records which is assumed to have partial future knowledge of the market dynamics, which we refer to as \textit{Expert oracle}. \textit{Expert oracle} is a filtered version of 
\textit{Expert}: at each trading day $t$ it selects the shortest tenor with positive return:
\begin{equation}
    a_t = \argmin_{a \in \mathcal{A}}{\{ a | (x_{t+a} - x_t) > 0}\}
\end{equation}
In particular, \textit{Expert oracle} is designed to select shortest tenors to avoid a perfect accuracy exploitation of risk-premia which would set an unrealistic benchmark for any model.
\textit{Expert} and \textit{Expert oracle} are used as human trader benchmarks.
\paragraph{Multivariate input data}
In order to learn how to effectively choose NDF tenors, the models have access to time series that can be broadly categorized in three groups:
\textit{FX spot rates}, \textit{technical indicators}, and \textit{NDF tenor volumes}. Daily FX spot rates serve as contextual market information and provide a frame of reference that aids the model in identifying profitable states. We include spot rates of 64 major and minor FX currency pairs, with a complete list provided in Appendix A.
Raw financial market data is often augmented with hand-crafted features to help combat noise and non-stationarity ~\cite{ntakaris2019feature}. To this end, we choose the following technical indicators and include details in Appendix A:
\begin{itemize}
    \item Simple moving average (SMA)
    \item Exponential moving average (EMA)
    \item Moving Average Convergence Divergence (MACD)
    \item Rolling standard deviation (RSD)
    \item Bollinger Bands (BB)
    \item ARIMA 1-day spot rate forecast
\end{itemize}
The last category of input features, \textit{NDF tenor volumes}, is obtained from DTCC NDF records. For each NDF pair under consideration and each admissible tenor $a \in \mathcal{A}$, we generate a time series of volumes $v_{t,a}$ which includes a summation over all NDF records at a specific day $t$. In particular, given a choice of maximum tenor of 90 days, each NDF pair contributes with a 90-dimensional multivariate volume time series to the input, which further emphasizes the need for a model capable of processing and aggregating information across highly multivariate time series. The code for downloading and preprocessing the data will be released after publication.

\begin{figure*}[!h]
   \centering
    \includegraphics[scale=0.50]{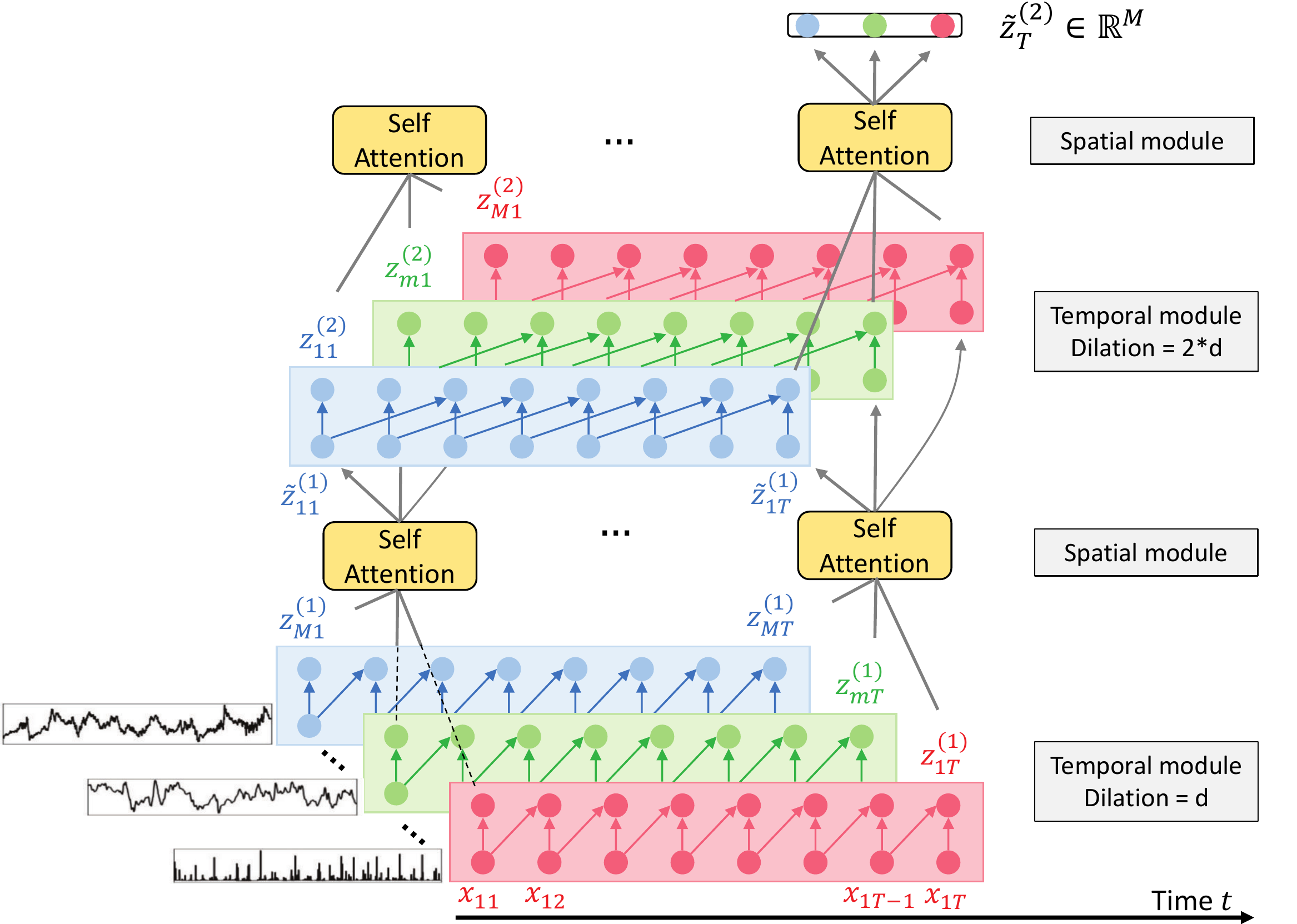}
    \caption{WATTNet overview: dilated TCNs are independently applied to each univariate input time series. A single dot-product attention head subsequently aggregates information across slices $\{z_{1,t} \dots z_{M,t}\}$ and the result $\{\tilde{\textbf{Z}}\}$ is passed to the next WATTBlock or used directly for the task at hand.}
    \label{fig:watt}
\end{figure*}

\section{Model}
\paragraph{Spatio-temporal modeling with WATTNet}
\textit{WaveATTentionNet} (WATTNet) is a novel model designed for highly multivariate time series inputs. WATTNet includes \textit{temporal modules}, tasked with independently aggregating information across time steps of univariate time series $\{\mathbf{x}\} \in \{\mathbf{X}\}$ and \textit{spatial modules} which aggregate features across slices of all time series at a specific time $t$ $\{x_{1,t}, \dots x_{M,t}\}$. Temporal and spatial modules are alternated and allow for learning a hierarchical spatio-temporal representation. An overview of the model is given in Figure~\ref{fig:watt}. 

\paragraph{Temporal learning}
Temporal learning is achieved by applying temporal dilated convolutions (TCN) to univariate time series $\{\mathbf{x}\} \in \{\mathbf{X}\}$. In particular, given a convolution with kernel size $k$ and dilation coefficient $d$, we compute the output at time $t$ of a dilated convolution of $\{\mathbf{x}\}$ as:
\begin{equation}
    z_t = \sum_{i=1}^{k} w_i * x_{t-i*d}
    \label{eq:dil}
\end{equation}
where $w_i$ is the $i^{\text{th}}$ weight of the convolutional kernel. Each univariate time series has access to its own set of convolutional weights $\mathbf{w}$ as temporal convolution operations are carried on independently. We note that independence between convolutions is necessary to provide the model with enough flexibility to treat time series with different characteristics. The outputs of the TCN operation are then concatenated as to form a multivariate latent time series $\{\mathbf{Z}\}$. In particular, WATTNet includes gated convolutions, a standard architectural component for sequential data. Two dilated TCNs are applied to $\{\mathbf{X}\}$ and the results $\{\mathbf{Z}\}_{\alpha}$, $\{\mathbf{Z}\}_{\beta}$ are passed to non-linear activation functions and then multiplied element-wise:
\begin{equation}
    \{\mathbf{Z}\} = \sigma(\{\mathbf{Z}\}_{\alpha}) \odot  \text{tanh}(\{\mathbf{Z}\}_{\beta})
    \label{eq:dil}
\end{equation}
where $\sigma$ indicates a sigmoid activation. The output $\{\mathbf{Z}\}$ is then fed into a spatial learning module after which the process repeats for a number of times depending on WATTNet's layer depth. 

\paragraph{Spatial learning}
A single-head scaled-dot product attention mechanism ~\cite{vaswani2017attention} is placed between dilated TCN layers and allows the model to exchange information across different input time series at a specific time slice. We compute key $\mathbf{K}$, query $\mathbf{Q}$ and value $\mathbf{V}$ by considering a slice $\{z_{1,t} \dots z_{M,t}\} \in \{\mathbf{Z}\}$ of latent time series at time $t$ as the input of learnable linear transformation of the type $\psi(\{z_{1,t} \dots z_{M,t}\})$ with weights $\mathbf{W}_k$, $\mathbf{W}_q$, $\mathbf{W}_v$. The resulting matrices $\mathbf{K}$, $\mathbf{Q}$, and $\mathbf{V}$ are then used in the standard scaled-dot product attention to return M weighted averages of values $\mathbf{V}$:
\begin{equation}
\{\tilde{z}_{1,t} \dots \tilde{z}_{M,t}\} = \text{softmax}(\frac{\mathbf{Q}\mathbf{K}^T}{\sqrt{d_k}}\mathbf{V})
\end{equation}
where $d_k$ is a scaling factor given by the second dimension of $\mathbf{K}$. 
The process is repeated for latent feature slices $\{z_{1,t} \dots z_{M,t}\}$, $t = 1, \dots T$ and the results are concatenated into $\{\tilde{\mathbf{Z}}\}$, a spatio-temporal latent representation of input data $\{\mathbf{X}\}$. Weights $\mathbf{W}_q, \mathbf{W}_k, \mathbf{W}_v$ are shared across the entire sequence length $T$, allowing the attention head to capture time-invariant features that incorporate information from multiple time series. Output $\{\tilde{\mathbf{Z}}\}$ can be used directly for different tasks to perform decision making conditioned on multivariate time series data; alternatively, if the task at hand benefits from deeper models, $\{\tilde{\mathbf{Z}}\}$ can instead be passed to the following TCN layer to perform additional cycles of temporal and spatial learning. 

\paragraph{Hierarchical representation}
A single temporal and spatial module constitute a full WATTNet layer of computation and is referred to as \textit{WATTBlock}. WATTBlocks can be stacked, in which case output $\{\tilde{\mathbf{Z}}\}$ becomes a hierarchical spatio-temporal representation of $\{\mathbf{X}\}$. As is the case with other TCN-based models, the dilation coefficient is doubled each temporal module as to provide an increasing receptive field which allows for a computationally inexpensive way to model long sequences. An additional benefit of the gradual dilation increase is the slow introduction of interaction terms between time series which include less lagged values for early WATTBlocks and more for later ones. At layer $i$, the dilated TCN for scalar output $z_t$ has a receptive field of $2^{i}k$, with $k$ being the size of the convolutional kernel. During spatial learning, the information flow across a slice of latent TCN output $\{\mathbf{Z}\}$ at time $t$ $\{z_{1,t}, \dots z_{M,t}\}$ is thus limited to $2^{i}k$ lagged values of the raw $\{\mathbf{X}\}$, given by $\{x_{1,[t-2^{i}k:t]}, \dots x_{M,[t-2^{i}k:t]}\}$. We observe that gradually increasing the size of this interaction window is key in learning a hierarchical representation of the data that strongly intertwines spatial and temporal causal effects. 

\paragraph{Computational requirements}
WATTNet can be used as a general lightweight tool for spatio-temporal modeling. The temporal modules are fully parallelizable due to complete independence of inputs and convolutional weights across different univariate time series.
To leverage an already existing fast CUDA implementation of parallel TCNs in PyTorch, we utilize \textit{grouped-convolutions}. In particular, the $M$ dimension of input time series $\{\mathbf{X}\}$ becomes the channel dimension for the TCN, and different convolutional kernels are applied to each input channel to obtain the corresponding output.
On the spatial learning front, the attention modules have a computational cost of $O(TM^2)$, which is comparable to the standard quadratic attention cost of $O(M^2)$ when $T << M$.

\section{NDF Tenor Selection}
Selecting a profitable tenor is challenging since it burdens the model with a choice of short tenors with smaller returns or long, risky tenors with a potentially greater return. One approach to training a tenor selection model is performing imitation learning on \textit{Expert} or \textit{Expert oracle} labels. Both have advantages and disadvantages; training on \textit{Expert} allows for daily online training and thus reduces the need for the model to extrapolate to periods further into the future. This aspect can be particularly beneficial for turbulent markets that display frequent regime switches. \textit{Expert oracle} labels, on the other hand, require information from up to $N$ days in the future, with $N$ being the maximum allowed tenor, since positive return filtering can only be performed by leveraging spot rate data. \textit{Expert oracle} labels can be advantageous since they teach the model to be \textit{risk-averse}; however, both approaches potentially include unwanted human-bias in the learned strategy. We propose an alternative approach in which WATTNet learns from optimal greedy tenor labels obtained directly from market data. Given the spot rate value for target FX pair $\{\mathbf{y}\}$, we extract the optimal tenor label $\bar{a}_t$ at time $t$ as:
\begin{equation}
    \bar{a}_t = \argmax_{a\in\mathcal{A}}{(y_{t+a} - y_{t})}
    \label{e}
\end{equation}
\textit{Policy divergence} ~\cite{ross2011reduction} is a performance degrading issue often present in imitation learning where the agent accumulates small expert imitation errors along the rollout and ends up in unexplored regions of the state space. To sidestep this issue, we base our strategy on a conditional independence assumption between tenor $a_t$ and state dynamics $s_{t+1}$: $p(s_{t+1}|s_{t}) = p(s_{t+1}|s_{t},a_{t})$.
The training procedure is then carried out as follows. Input time series $\{\mathbf{X}\}$ is split into overlapping length $T$ slices which are then shuffled into a training dataset. At each trading day $t$, the model $\phi$ is trained via standard gradient-descent methods to minimize the cross-entropy loss of outputs $\phi(\{\mathbf{X}_{[t-T:t]}\})$ and tenor labels $\bar{a}_t$. 

\begin{table*}[h!]
\small
\centering
\setlength\tabcolsep{3pt}
\begin{tabular}{lrrrrrrrrr}
\toprule
 && USDCNY &&& USDKRW &&& USDIDR & \\
\cmidrule(r){1-10}
Model & ROI & opt.acc & nn.acc & ROI & opt.acc. & nn.acc. & ROI & opt.acc. & nn.acc. \\
\midrule
Optimal & 759.8 & 100 & 100 & 844.2 & 100 & 100 & 1260.0 & 100 & 100 \\
Expert (oracle) & 77.6 & 21.6 & 100 & 139.4 & 13.7 & 100 & 152 & 3.6 & 100 \\
\midrule
Expert & 0.0 & 1.4 & 47.8 & 12.7 & 1.9 & 49.9 & 230 & 0.4 & 67.0 \\
Momentum-1 & 14.6 & 1.4 & 48.3 & 10.7 & 0.9 & 49.7 & 201 & 0.7 & 66.5 \\
Momentum-90 & 4.9 & 6.7& 54.1 & 31.9 & 2.6 & 56.2 & $\textbf{338}$ & $\textbf{1.9}$ &69.2 \\
GRU-I  & $26.7.\pm48.5$ & $4.9\pm0.7$ & $54.7\pm1.0$ & $-98.7\pm35.4$ & $1.5\pm0.5$ & $52.1\pm1.4$ & $83.5\pm33.1$ & $0.6\pm0.1$ & $62.9\pm2.6$  \\
LSTM-I & $74.3\pm37.3$ & $3.7\pm0.9$ & $58.7\pm2.2$ & $74.6\pm30.0$ & $2.6\pm0.6$ & $56.0\pm2.5$ & $146.4\pm40.4$ & $1.1\pm0.3$ & $66.5\pm1.1$ \\
WATTNet & $\textbf{219.1}\pm25.5$ & $\textbf{6.7}\pm0.7$ & $\textbf{59.5}\pm1.6$ & $\textbf{142.4}\pm16.9$ & $\textbf{2.7}\pm0.2$ & $\textbf{59.5}\pm1.0$ & $280.2\pm39.9$ & $1.3\pm0.3$ & $\textbf{69.5}\pm0.9$ \\
\bottomrule
\end{tabular}
\caption{Test results in percentages (average and standard error). Best performance is indicated in bold.}
\label{tab:allresone}
\end{table*}

\begin{table}
\small
\centering
\setlength\tabcolsep{3pt}
\begin{tabular}[h!]{lrrrr}  
\toprule
Market & $\mu_{\text{train}}$ & $\sigma_{\text{train}}$ & $\mu_{\text{val}}$ & $\sigma_{\text{val}}$\\
\midrule
USDCNY & 11.33 & 231.63 & -2.96 & 232.88 \\
USDIDR & 22.04 & 506.17 & 3.48 & 335.75 \\
USDINR & 2.88 & 297.40 & 8.19 & 286.79 \\
USDKRW & 0.50 & 471.25 & -12.75 & 387.49 \\
USDPHP & 10.44 & 271.37 & -0.79 & 240.37 \\
USDTWD & -3.73 & 307.28 & -2.09 & 235.29 \\
\bottomrule
\end{tabular}
\caption{Statistics of 1-step (daily) percent returns. Units reported are scaled up by $1e^{5}$ for clarity.}
\label{tab:prmet}
\end{table}
\begin{figure}[t]
    \centering
    \includegraphics[scale=0.19]{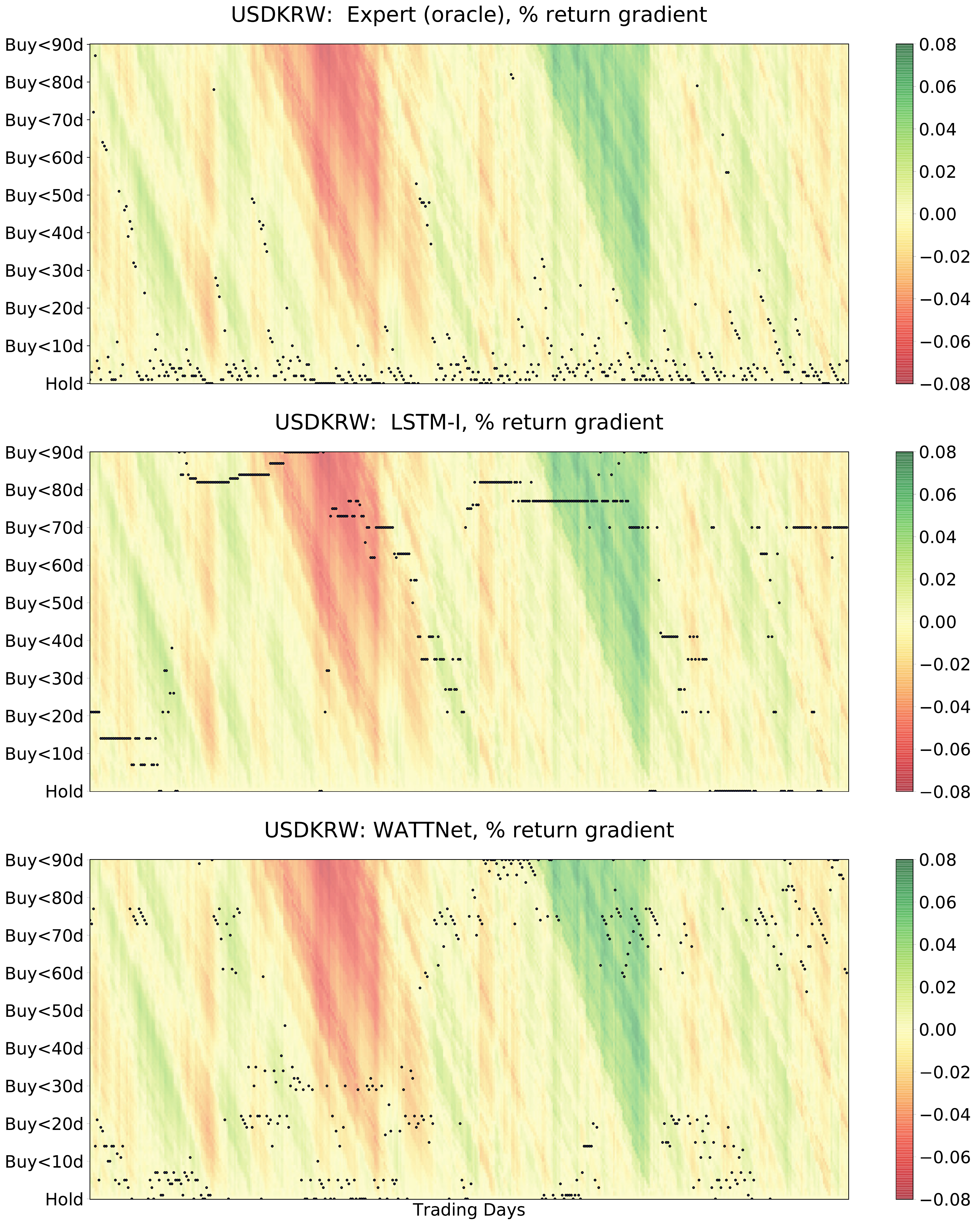}
    \caption{Tenor actions across USDKRW test set. Background gradient is used
    to indicate the ROI of different tenors. The gradient is
    slanted since the raw return at day $t$ with a tenor of 90 days
    is the same as the return from trading day $t + i$ with tenor of
    $90 - i$ days. WATTNet learns to correctly exploit long periods of positive trend and places less actions than LSTM-I in negative (red) return regions.}
    \label{fig:actions}
\end{figure}
\section{Experimental Results}
\paragraph{NDF Markets}
The experimental evaluation covers the following 6 major NDF markets: \textit{Chinese Yuan} (USDCNY), \textit{Indonesian Rupiah} (USDIDR), \textit{Indian Rupee} (USDINR), \textit{Korean Won} (USDKRW), \textit{Philippine Peso} (USDPHP), \textit{Taiwan Dollar} (USDTWD). We elaborate on the results for USDCNY, USDKRW, USDIDR and include discussion of the remaining 3 markets as supplementary material (Appendix C). The selection has been carried out to test the proposed method on markets with different characteristics as shown in Table \ref{tab:prmet}.

\paragraph{Recurrent and classical baselines}
Two-layer stacked GRU and LSTMs are used as additional baselines for tenor and referred to as GRU-I and LSTM-I. Both recurrent baselines and WATTnet are connected to a fully-connected head which takes as input the latent representation of $\{\mathbf{X}\}$ produced by the model. A probability distribution over tenor actions is then obtained via softmax. Additionally, we include the following classical trading baselines:
\begin{itemize}
    \item \textit{Momentum-1}: 1-day lag of expert tenor actions. Effective in markets where monotonic behavior in the spot rate is frequent and the monotonic sequences span several trading periods.
    \item \textit{Momentum-90}: best performing tenor from 90 days prior. Effective in markets with trends whose duration is longer compared to the maximum tenor.
\end{itemize}

\paragraph{Training setup} The models are implemented in PyTorch and trained using Adam \cite{kingma2014adam} and a learning rate cosine decay schedule from $6e^{-4}$ down to $3e^{-4}$. To avoid overfitting uninformative noisy patterns in stale data input sequence length is set to $30$ days. In addition, to enable a fair comparison and avoid additional overfitting we employ an early stopping scheme based on training loss that is motivated by different convergence times of different models. We use a static testing approach with a long period of $446$ out-of-sample trading days to test stability of the learned trading strategy under turbulent market conditions and a wider distributional shift between in-sample and out-of-sample data. PyTorch code for models and training procedure is included in the supplementary material. 

\paragraph{Metrics}
The following metrics are used to benchmark the performance of trading models and baselines:
\begin{itemize}
    \item \textit{Return on investment} (ROI): given a tenor action $a$ at time $t$ and spot rate value $x_t$, the percent ROI is calculated as $\text{ROI}_t = 100 \left( \frac{x_{t+a} - x_t}{x_t}\right)$
    \item \textit{Optimal accuracy}: standard supervised learning accuracy of model outputs $\hat{y}_t$ versus \textit{optimal} tenor labels $y_t$.
    \item \textit{Non-negative return accuracy}: accuracy of model outputs $\hat{y_t}$ compared to tenor actions with positive or zero return. At time $t$, there are generally multiple tenor actions with non-negative return, thus rendering \textit{non-negative accuracy} a less strict metric compared to \textit{optimal accuracy}. It should be noted that it is possible for a high ROI trading model to show poor optimal accuracy but competitive positive return accuracy since non-negative accuracy also captures positive ROI strategies that differ from optimal tenor labels.
\end{itemize}

\paragraph{Discussion of results}
We characterize the 6 NDF markets under evaluation based on mean $\mu$ and standard deviation $\sigma$ of their 1-day returns and performance of classical baselines. Mean-variance statistics given in Table \ref{tab:prmet} show easier markets with similar volatility in training and test periods (e.g. USDINR) as well as markets that appear more erratic and thus challenging to trade profitably in. From Table \ref{tab:allresone}, we determine USDCNY and USDKRW to be challenging for \textit{Momentum} and recurrent baselines, in addition to being barely profitable for expert traders. GRU-I is unable to consistently get positive ROI and we suspect this is caused by its insufficient ability to exchange information between time series. LSTM-I, on the other hand, fares better by leveraging its memory module to perform rudimentary spatial learning ~\cite{santoro2018relational}. USDIDR appears to be highly profitable for \textit{Momentum-90}, a phenomenon highlighting a longer positive trend and a higher density of positive return trades in the test data. 
\begin{figure*}[h!]
   \centering
    \includegraphics[scale=0.25]{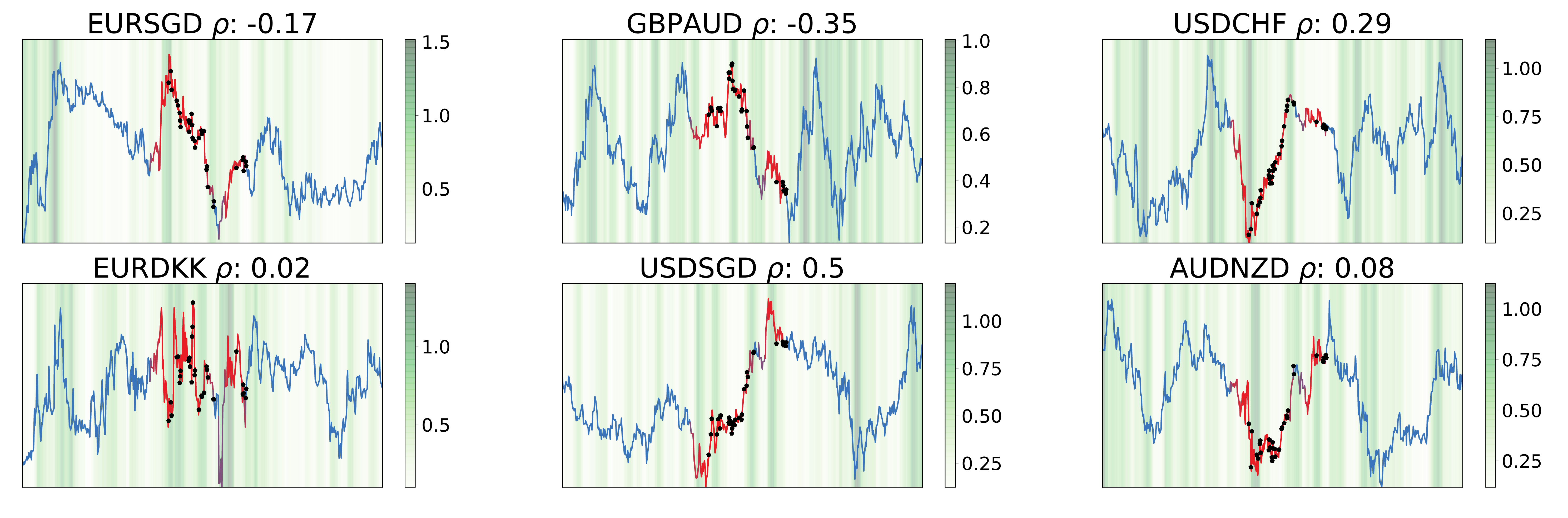}
    \caption{Highest impact spot rate features behind 90 day tenor actions in USDCNY testing period. Background color shows $20$ day rolling standard deviation.}
    \label{inputgrad}
\end{figure*}
\begin{figure}[h!]
   \centering
    \includegraphics[scale=0.34]{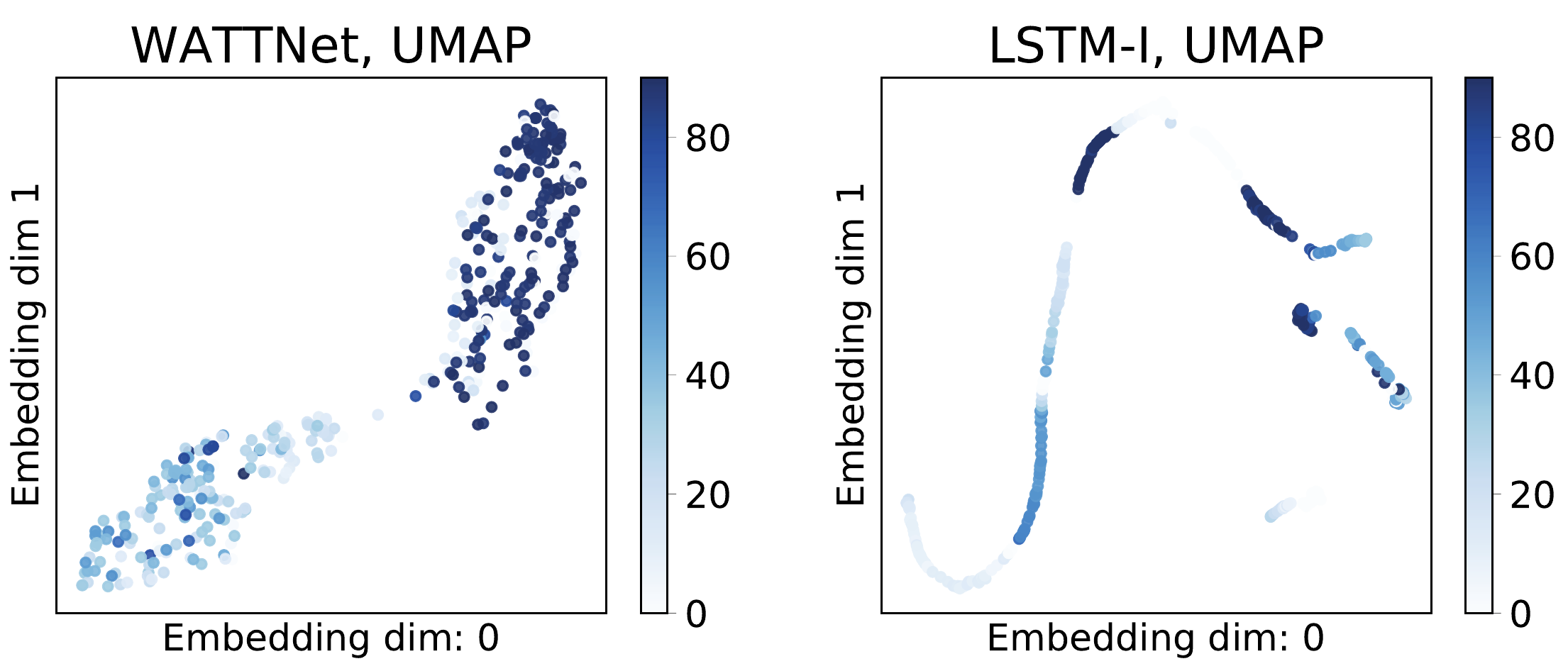}
    \caption{UMAP embedding of model latents. The points are labeled according to the final tenor output of the model.}
    \label{fig:umap}
\end{figure}

\begin{itemize}
    \item \textbf{USDCNY:} The results in Table \ref{tab:allresone} show that all classical baselines, including expert actions, perform poorly in terms of ROI and optimal tenor accuracy. LSTM and WATTNet, on the other hand, are able to generalize better, with WATTNet surpassing ROI of oracle trades.
    \item \textbf{USDKRW:} The USDKRW train-test split presents the most challenging trading period across all 6 NDF markets considered in this work. The return statistics in Table \ref{tab:prmet} show a decrease in mean return as well as a significant change in volatility. Traditional baselines perform poorly whereas WATTNet surpasses \textit{Expert oracle} ROI. Figure~\ref{fig:actions} highlights the ability of WATTNet to adjust its tenor actions depending on trading spot rate trends.
    \item \textbf{USDIDR:} USDIDR has been chosen to evaluate the baselines under a profitable trading period. All classical baselines trade with positive ROI, even surpassing oracle ROI due to their propensity to trade at longer tenors. WATTNet achieves performance competitive with \textit{Momentum} baselines, showing that it is capable of fully exploiting trading periods with long positive trends and a wide shift in volatility between training and test data.
\end{itemize}
\subsection{Explainability}
Model explainability is particularly important in application areas where the models are tasked with critical decision making, as is the case for algorithmic trading. Understanding driving factors behind a trading decision is necessary to properly assess the risks involved. We tackle this issue by proposing two orthogonal approaches for evaluating the tenor selection strategy in terms of noise stability and driving factors. 
\paragraph{Feature importance by input gradients}
To pinpoint the driving factors of trades at different tenors we propose sorting the features by their gradient magnitude. In the case of tenor selection, each input feature carries a specific meaning which can be leveraged by domain experts to confirm whether the model outputs actions consistent with market dynamics.

Given a batch of $N$ multivariate input sequences $\{\mathbf{X}^1\} \dots \{\mathbf{X}^N\}$ with tenor labels equal to $a$, we compute the cross-entropy loss of model $\mathcal{L}$ and derive the empirical expectation for the absolute value of time series $\{\mathbf{x}_j\}$ input gradient as follows:
\begin{equation}
 \mathcal{G}_j = \frac{1}{T} \left|\sum_{t=1}^{T}\frac{\partial \mathcal{L}}{\partial x_{jt}}\right|
\end{equation}
To illustrate the effectiveness of this approach we select $6$ spot rate features with highest absolute gradient values for tenor actions of $90$ days in the USDCNY test data: EURSGD, GBPAUD, USDCHF, EURDKK, USDSGD, AUDNZD (Figure~\ref{inputgrad}). The Pearson's correlation coefficient $\rho$ between USDCNY and each of the above-listed features is computed with training and test sets. In the background, 20 day rolling standard deviation highlights regions of low and high volatility. Input sequences $\{\mathbf{x}\} \in \{\mathbf{X}\}$ which are mapped by the model to $90$ tenor actions are colored in red, whereas the actions themselves are indicated as black dots. The model learns to trade on long tenors when currencies that are positively correlated with USDCNY, such as USDSGD, undergo periods of growth. The degree to which such trends affect the model is directly reflected in $\rho$: USDCHF, still positively correlated with USDCNY, shows a less decisive positive trend, with additional ups and downs. Moreover, the model learns to favor trading periods with low volatility.
This analysis can be extended by domain experts to additional input features, such as technical indicators or past tenor actions, and can boost confidence in the decisions made by the model.

\paragraph{Latent space representation}
Desired properties of the learned trading strategy are \textit{input coherence} and \textit{stability}.
Input coherence is achieved by a model that outputs similar tenors for similar states. Stability, on the other hand, is concerned with how much noise perturbation is required to cause a tenor switch from a certain state. 
We perform a visual inspection of these properties via a \textit{uniform manifold approximation and projection} (UMAP) which excels at capturing both local and global structure of the high-dimensional data ~\cite{mcinnes2018umap}. For each model, latent vectors of their last layer are embedded into two-dimensional space. UMAP outputs compact clusters of labels for input coherent models and more volumetric clusters for stable models. From Figure~\ref{fig:umap} we observe that WATTNet learns a coherent latent representation that clusters low and high tenors correctly and covers a larger volume of the embedding space. Instability of GRU-I and LSTM-I observed in the results of Table \ref{tab:allresone} can in part be explained by noticing that their learned representation lies on thin lower-dimensional manifolds with mixed tenor labels. As a result, small noise perturbations can cause wide jumps in tenor actions, potentially causing a drop in performance.

\section{Conclusion}
We introduced a challenging imitation learning problem, tenor selection for \textit{non-deliverable-forward} (NDF) contracts. With the goal of promoting further research in this direction, we constructed and released a comprehensive NDF dataset and designed \textit{WaveATTentioNet} (WATTNet), a novel model for spatio-temporal data which outperforms expert benchmarks and traditional baselines across several NDF markets. Finally, we employed two explainability techniques to determine driving factors and noise stability of the learned tenor strategy. Future work includes defining and solving \textit{order sizing} of an NDF contract, as well as augmenting the tenor selection model with a reinforcement learning agent head. Reminiscent of world models ~\cite{ha2018recurrent}, such an approach would sacrifice explainability for additional flexibility; the agent could for example be equipped with a memory module consisting of a recurrent network to keep track of its previous trades. A different direction would be exploring the use of multi-head attention instead of single-head as an effective modification to WATTNet in the case of long input sequences or multiple relations between features.

{\small
\selectfont
\bibliographystyle{aaai}
\bibliography{root_arxiv}}
\pagebreak

\clearpage
\appendix

\begin{table}
\centering
\begin{tabular}[!h]{lrr}
\toprule
\multicolumn{2}{c}{6 context FX instruments} \\
\cmidrule(r){1-3}
\cmidrule(r){1-3}
Currency 1 & Currency 2 & Name \\
\midrule
United States Dollar & Chinese Yuan & USDCNY \\
United States Dollar & Indonesian Rupiah & USDIDR \\
United States Dollar & Indian Rupee & USDINR \\
United States Dollar & Korean Won & USDKRW \\
United States Dollar & Philippine Peso & USDPHP \\
United States Dollar & Taiwan Dollar & USDTWD \\
\bottomrule
\end{tabular}
\label{tab:ndfspot}
\caption{NDF FX instruments}
\end{table}
\begin{table}
\centering
\begin{tabular}[t]{lrr}
\toprule
\multicolumn{2}{c}{58 context FX instruments} \\
\cmidrule(r){1-3}
\cmidrule(r){1-3}
Currency 1 & Currency 2 & Name \\
\midrule
Euro & United States Dollar & EURUSD \\
Pound Sterling & United States Dollar & GBPUSD \\
United States Dollar & Canadian Dollar & USDCAD \\
United States Dollar & Swiss Franc & USDCHF \\
United States Dollar & Japanese Yen & USDJPY \\
Euro & Pound Sterling & EURGBP \\
Euro & Swiss Franc & EURCHF \\
Austrial Dollar & United States Dollar & AUDUSD \\
Austrial Dollar & Canadian Dollar & AUDCAD \\
Euro & Japanese Yen & EURJPY \\
Pound Sterling & Japanese Yen & GBPJPY \\
Euro & Australian Dollar & EURAUD \\
Euro & Czech Koruna & EURCZK \\
Euro & Hungarian Forint & EURHUF \\
Euro & New Zealand Dollar & EURNZD \\
Euro & Swedish Krona & EURSEK \\
Euro & Singapore Dollar & EURSGD \\
Euro & Canadian Dollar & EURCAD \\
Euro & Danish Krone & EURDKK \\
Euro & Norwegian Krone & EURNOK \\
Euro & Polish Zloty & EURPLN \\
Euro & Turkish Lira & EURTRY \\
Euro & South African Rand & EURZAR \\
United States Dollar & Danish Krone & USDDKK \\
United States Dollar & Hungarian Forint & USDHUF \\
United States Dollar & Mexican Peso & USDMXN \\
United States Dollar & Poland Zloty & USDPLN \\
United States Dollar & Swedish Krona & USDSEK \\
United States Dollar & Thai Baht & USDTHB \\
United States Dollar & South African Rand & USDZAR \\
United States Dollar & Czech Koruna & USDCZK \\
United States Dollar & Hong Kong Dollar & USDHKD \\
United States Dollar & Norwegian Krone & USDNOK \\
United States Dollar & Saudi Riyal & USDSAR \\
United States Dollar & Singapore Dollar & USDSGD \\
United States Dollar & Turkish Lira & USDTRY \\
Pound Sterling & Australian Dollar & GBPAUD \\
Pound Sterling & Swiss Franc & GBPCHF \\
Pound Sterling & South African Rand & GBPZAR \\
Pound Sterling & Singapore Dollar & GBPSGD \\
Australian Dollar & Japanese Yen & AUDJPY \\
Australian Dollar & Singapore Dollar & AUDSGD \\
Canadian Dollar & Japanese Yen & CADJPY \\
Swiss Franc & Japanese Yen & CHFJPY \\
New Zealand Dollar & Canadian Dollar & NZDCAD \\
New Zealand Dollar & United States Dollar & NZDUSD \\
Singapore Dollar & Japanese Yen & SGDJPY \\
South African Rand & Japanese Yen & ZARJPY \\
Pound Sterling & Canadian Dollar & GBPCAD \\
Pound Sterling & New Zealand Dollar & GBPNZD \\
Pound Sterling & Poland Zloty & GBPPLN \\
Australian Dollar & New Zealand Dolar & AUDNZD \\
Canadian Dollar & Swiss Franc & CADCHF \\
Canadian Dollar & Singapore Dollar & CADSGD \\
Swiss Franc & South African Rand & CHFZAR \\
New Zealand Dollar & Japanese Yen & NZDJPY \\
New Zealand Dollar & Singapore Dollar & NZDSGD \\
Turkish Lira & Japanese Yen & TRYJPY \\
United States Dollar & Malaysian Ringgit & USDMYR \\
\bottomrule
\end{tabular}
\label{context}
\caption{58 FX instruments used as context in the dataset}
\end{table}

\section{Dataset details}
Here we report more information about the dataset. 
\begin{itemize}
    \item Period: 2013-09-10 to 2019-06-17
    \item Number of features ($M$ dimension of $\{\textbf{X}\}$): 1123. 64 FX pairs, 519 technical indicators, 540 NDF volume features. More specifically, 90 NDF volume features per NDF pair (90 * 6) 
\end{itemize}
The list of context NDF spot rates is given in Table 1, and context FX spot rates in Table 2. Context spot rates have been obtained via the \textit{Oanda Developer API}. 
\paragraph{Technical Indicators}
Here we include more details about the technical indicators, including their dimensionality as number of features.
\begin{itemize}
    \item Simple moving average (SMA): SMA is a trend-following indicators that filters out high frequency oscillations by smoothing spot rate time series. We use both 7 day and 21 day averages (i.e $N = 7$, $N = 21$). Included for all 64 FX pairs (total: 64 * 2 = 128 features).
    \item Exponential moving average (EMA): similarly to SMA, EMA is a lagging indicator. The weight for each values exponentially decreases, with bigger weight assigned to more recent values of the time series. We include 12 day and 26 day EMAs (total: 128 features).
    \begin{equation}
        \mu_t = \Big\{ \begin{array}{ll}
            x_t & t = 1 \\
            \alpha x_t + (1 - \alpha) \mu_{t-1} & t > 1
        \end{array}
    \end{equation}
    \item Moving Average Convergence Divergence (MACD): A filtering function with a bigger time constant is subtracted to another with a smaller one in order to estimate the derivative of a time series. In particular, a common choice of filtering functions are 12 day and 26 day exponential moving averages: $\text{MACD} = \text{EMA}_{12} - \mbox{EMA}_{26}$ (total: 64 features).
    \item Rolling standard deviation (RSD): a window of 20 days is used to compute standard deviation at time $t = 20$ (total: 64 features).
    \item Bollinger Bands (BB): commonly used to characterize volatility over time. We include both an upper band derived and a lower band as features $\text{BB} = \text{SMA} \pm \text{RSD}$ (total: 64 * 2 = 128 features)
\begin{table}
\centering
\setlength\tabcolsep{3pt}
\begin{tabular}[!h]{lrrr}  
\toprule
Layer & Input dim. & Output dim. \\
\midrule
Recurrent-1 & 1123 & 512 \\
Recurrent-2 & 512 & 512  \\
FC-1 & 512 & 128 \\
FC-1 & 128 & 91 \\
\bottomrule
\end{tabular}
\caption{Layer dimensions for recurrent models GRU-I and LSTM-I}
\label{tab:rec}
\end{table}
    \begin{table}
\centering
\setlength\tabcolsep{3pt}
\begin{tabular}[t]{lrrrr}  
\toprule
Layer & \textit{M}-in & \textit{M}-out & \textit{T}-in & \textit{T}-out\\
\midrule
FC-cmp & 1123 & 90 & 30 & 30 \\
WATTBlock-1 & 90 & 90 & 30 & 27\\
WATTBlock-2 & 90 & 90 & 27 & 23 \\
WATTBlock-3 & 90 & 90 & 23 & 21 \\
WATTBlock-4 & 90 & 90 & 21 & 17 \\
WATTBlock-5 & 90 & 90 & 17 & 15 \\
WATTBlock-6 & 90 & 90 & 15 & 11 \\
WATTBlock-7 & 90 & 90 & 11 & 9 \\
WATTBlock-8 & 90 & 90 & 9 & 5 \\
FC-1 & 5*90 & 512 & 1 & 1\\
FC-2 & 512 & 91 & 1 & 1 \\
\bottomrule
\end{tabular}
\caption{Layer dimensions (\textit{M} and \textit{T}) for WATTNet}
\label{tab:watt}
\end{table}
    \item ARIMA spot rate forecast: an ARIMA model is trained to perform 1-day forecasts of spot rates using data from periods preceding the start of the training set to avoid information leakage. The forecasts are added as time series features. Included for NDF pairs and for USDMYR (total: 7 features)
\end{itemize}
\paragraph{Training hyperparameters}
The data is normalized as $x_t = \frac{x_t - \mu}{\sigma}$ where $\mu$ and $\sigma$ are 60-day rolling mean and standard deviation respectively.
The dataset is split into ovelapping sequences of 30 trading days. Batch size is set to 32.

\begin{table*}[t]
\small
\centering
\setlength\tabcolsep{3pt}
\begin{tabular}{lrrrrrrrrr}
\toprule
 && USDINR &&& USDPHP &&& USDTWD & \\
\cmidrule(r){1-10}
Model & ROI & opt.acc & nn.acc & ROI & opt.acc. & nn.acc. & ROI & opt.acc. & nn.acc. \\
\midrule
Optimal & 1288.7 & 100 & 100 & 900.0 & 100 & 100 & 612.0 & 100 & 100 \\
Expert (oracle) & 123.2 & 6.7 & 100 & 114.1 & 6.7 & 100 & 91.4 & 9.8 & 100 \\
\midrule
Expert & 216.1 & 0.7 & 62.7 & 117.1 & 0.2 & 57.2 & 43.3 & 0.9 & 50.2 \\
Momentum-1 & 218.1 & 2.6 & 62.5 & 110.9 & 0.7 & 58.9 & 44.0 & 0.5 & 50.0 \\
Momentum-90 & 300.9 & 2.2 & 57.4 & 134.9 & 1.2 & 58.9 & 10.0 & 1.4 & 55.2 \\
GRU-I  & $273.5\pm29.0$ & $2.0\pm0.5$ & $63.2\pm2.3$ & $244.2\pm27.5$ & $2.3\pm0.1$ & $62.5\pm1.4$ & $46.6\pm34.9$ & $2.5\pm0.3$ & $57.6\pm1.2$  \\
LSTM-I & $239.9\pm67.3$ & $2.4\pm0.5$ & $61.6\pm2.0$ & $307.1\pm16.8$ & $1.8\pm0.2$ & $\textbf{71.3}\pm1.0$ & $46.8\pm27.9$ & $2.8\pm0.6$ & $59.6\pm1.4$ \\
WATTNet & $\textbf{313.0}\pm28.1$ & $\textbf{3.7}\pm0.2$ & $\textbf{65.9}\pm1.0$ & $\textbf{329.7}\pm31.2$ & $\textbf{3.3}\pm0.6$ & $70.0\pm0.7$ & $\textbf{162.9}\pm13.9$ & $\textbf{2.8}\pm0.3$ & $\textbf{59.8}\pm1.4$ \\
\bottomrule
\end{tabular}
\label{allrestwo}
\caption{Test results in percentages (average and standard error). Best performance is indicated in bold.}
\end{table*}  

\section{Architectural hyperparameters}
We herein provide detailed information about the model design. 
\paragraph{Recurrent Models}
GRU-I and LSTM-I share the same structure given in Table~\ref{tab:rec}. Layer depth has been chosen as the best performing in the range 1 to 6. Latent spatio-temporal representation $\{\tilde{\mathbf{Z}}\}$ of input time series $\{\mathbf{X}\}$ is obtained as the output of \textit{Recurrent-2}. $\{\tilde{\mathbf{Z}}\}$ is the tensor transformed via UMAP and shown in Appendix D.

\paragraph{WATTNet}
Details about the WATTNet used are found in Table~\ref{tab:watt}. We employ fully-connected layer (\textit{FC-emp}) for compression in order to constrain GPU memory usage to less than 6GB. Improved results can be obtained by lifting this restriction and increasing the WATTBlock $M$-dimension. $\{\tilde{\mathbf{Z}}\}$ is the output of \textit{WATTBlock-8}. Due to the relatively short input sequence length, the dilation is scheduled for reset every 2 WATTBlocks. More specifically, the dilation coefficients for temporal learning are 2, 4, 8, 16, 2, 4, 8, 16 for WATTBlock-1 to 8. As is common in other dilated TCN models, dilation introduces a reduction of $T$ dimension for deeper layers. This effect is observable in Table~\ref{tab:watt}. We utilize a residual architecture for the attention module where the output is the summation of pre and post attention tensors, i.e:
\begin{equation}
    \{\tilde{\mathbf{Z}}_{out}\} = \sigma(\{\tilde{z}_{1,t} \dots \tilde{z}_{M,t}\}) + (\{x_{1,t} \dots x_{M,t}\})
\end{equation}

\section{Additional results}
We provide additonal results and discussion of USDINR, USDPHP and USDTWD. Training and test periods for all NDF pairs are visualized in Figures 1-2 for reference. Figure \ref{fig:dists} shows \textit{Expert} and \textit{Expert oracle} tenor action distributions. The results are given in Table 8.
\paragraph{USDINR}
USDINR shows positive trend in mean return and its volatility remains constant between training and test. All baselines achieve high ROI in this scenario. Of particular interest is GRU-I, which surpasses LSTM-I likely due to the relative simplicity of USDINR dynamics. With a stable, positive trend GRU-I is able to focus on the spot rates directly, thus circumventing its inability to perform proper spatial learning.
\paragraph{USDPHP} USDPHP is slightly more challenging due to its minimal volatility shift. WATTNet outperforms other models in terms of ROI and optimal accuracy.
\paragraph{USDTWD} USDTWD has negative mean return and shows a reduction in volatility in the test set. WATTNet is able to exploit this phenomenon and significantly outperforms all other baselines. 

\paragraph{Rolling testing and online training}
In general, static testing can turn out to be particularly challenging for financial data with longer test sets, since the models are tasked with extrapolating for long periods, potentially under non-stationary condition of the market. A rolling testing approach can be beneficial; however, the size of window and test periods require ad-hoc tuning for each FX currency. A key factor influencing the optimal choice of window size is the average length of \textit{market regimes}. We leave the discussion on optimal window selection for rolling testing as future work.
Figure~\ref{fig:dists} shows distributions of tenor labels and highlights the difference in average tenor length.

\section{Tenor actions and latent UMAP embeddings}
In this section we provide a complete collection of tenor action plots and UMAP embeddings for all NDF markets under consideration. Background gradient is used
to indicate the ROI of different tenors. The gradient is
slanted since the raw return at day t with a tenor of 90 days
is the same as the return from trading day $t + i$ with tenor of $90 - i$ days. 
\begin{figure}[!h]
   \centering
   \includegraphics[scale=0.33]{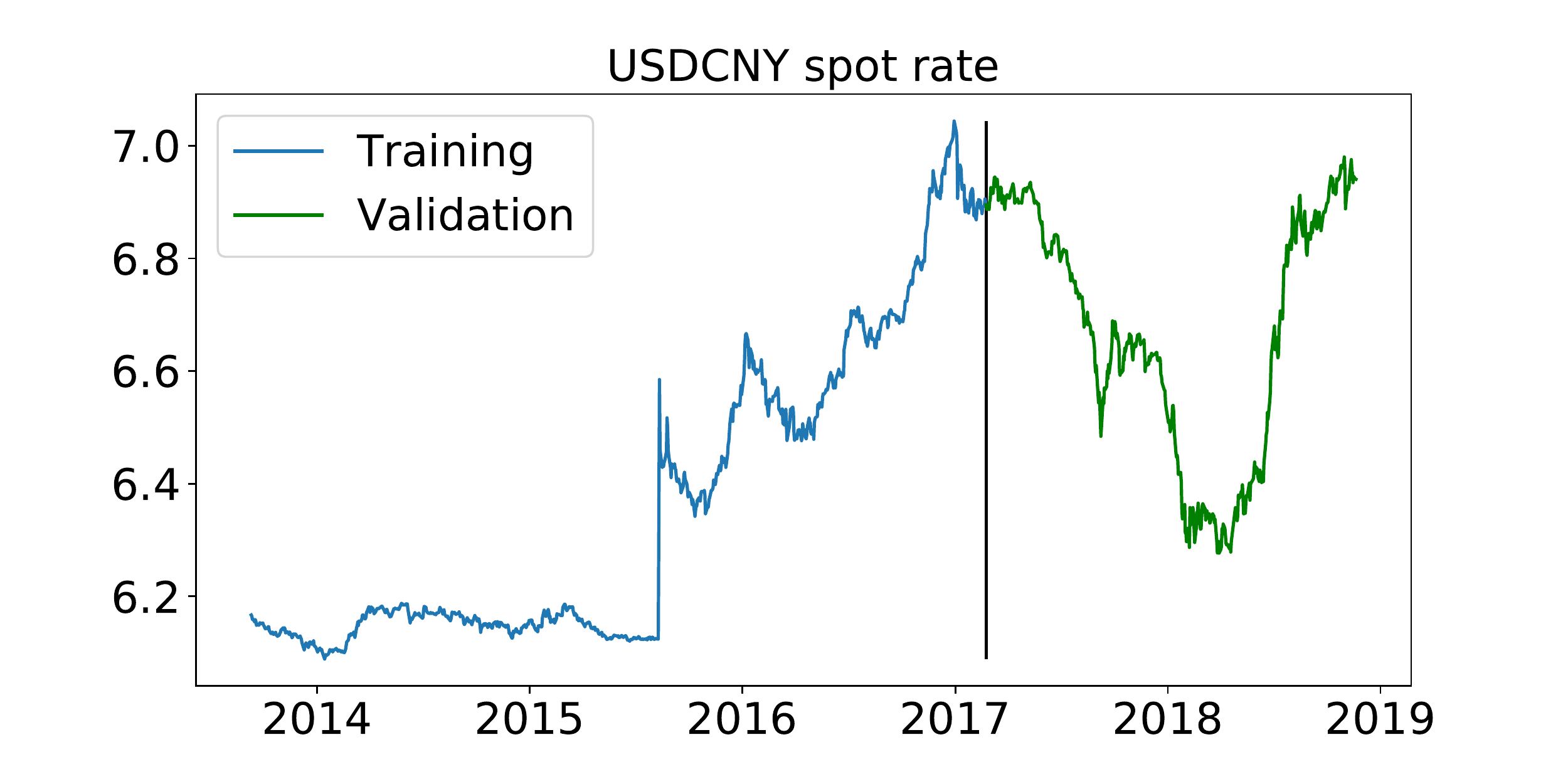}
    \includegraphics[scale=0.33]{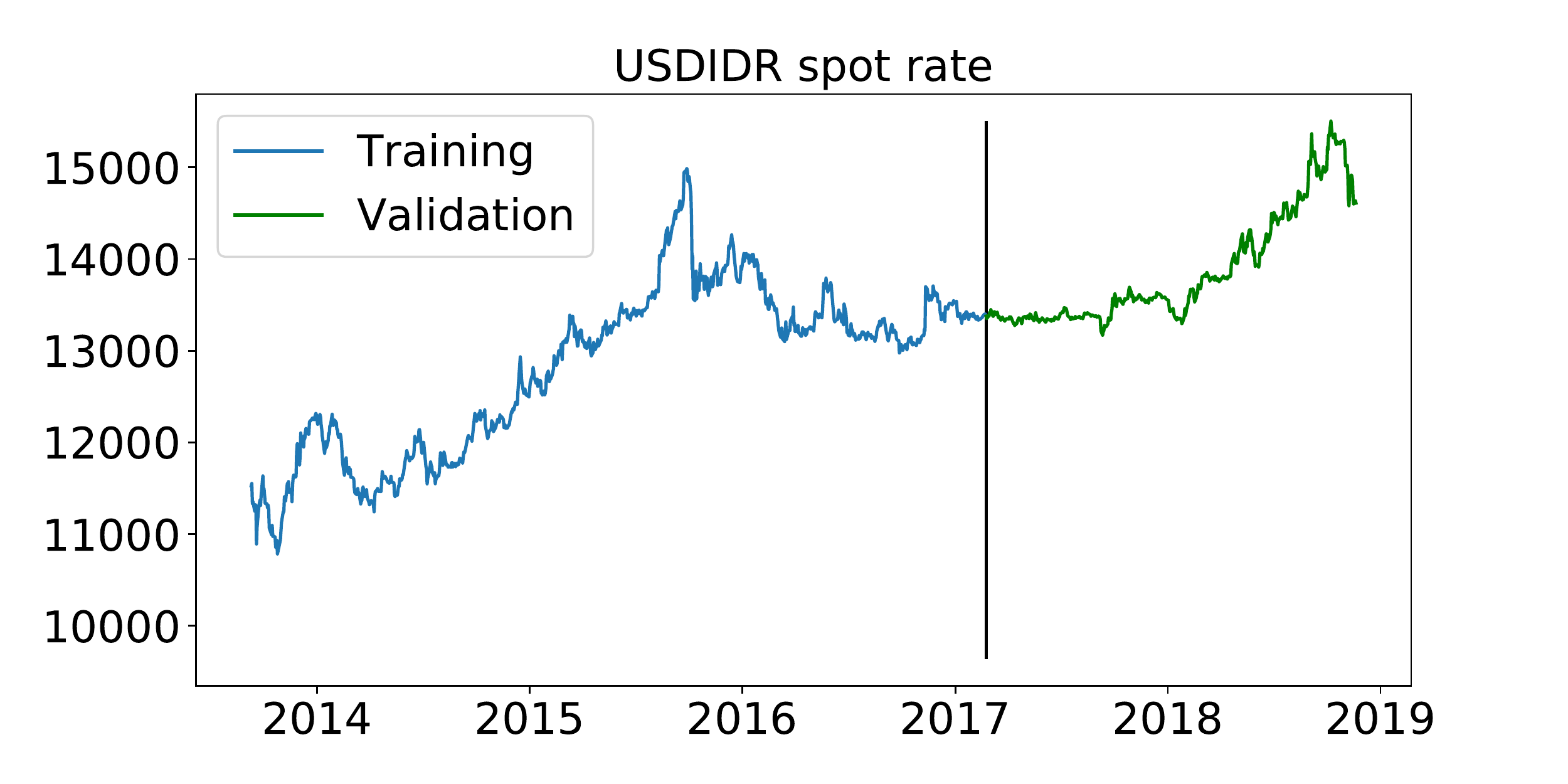}
    \includegraphics[scale=0.33]{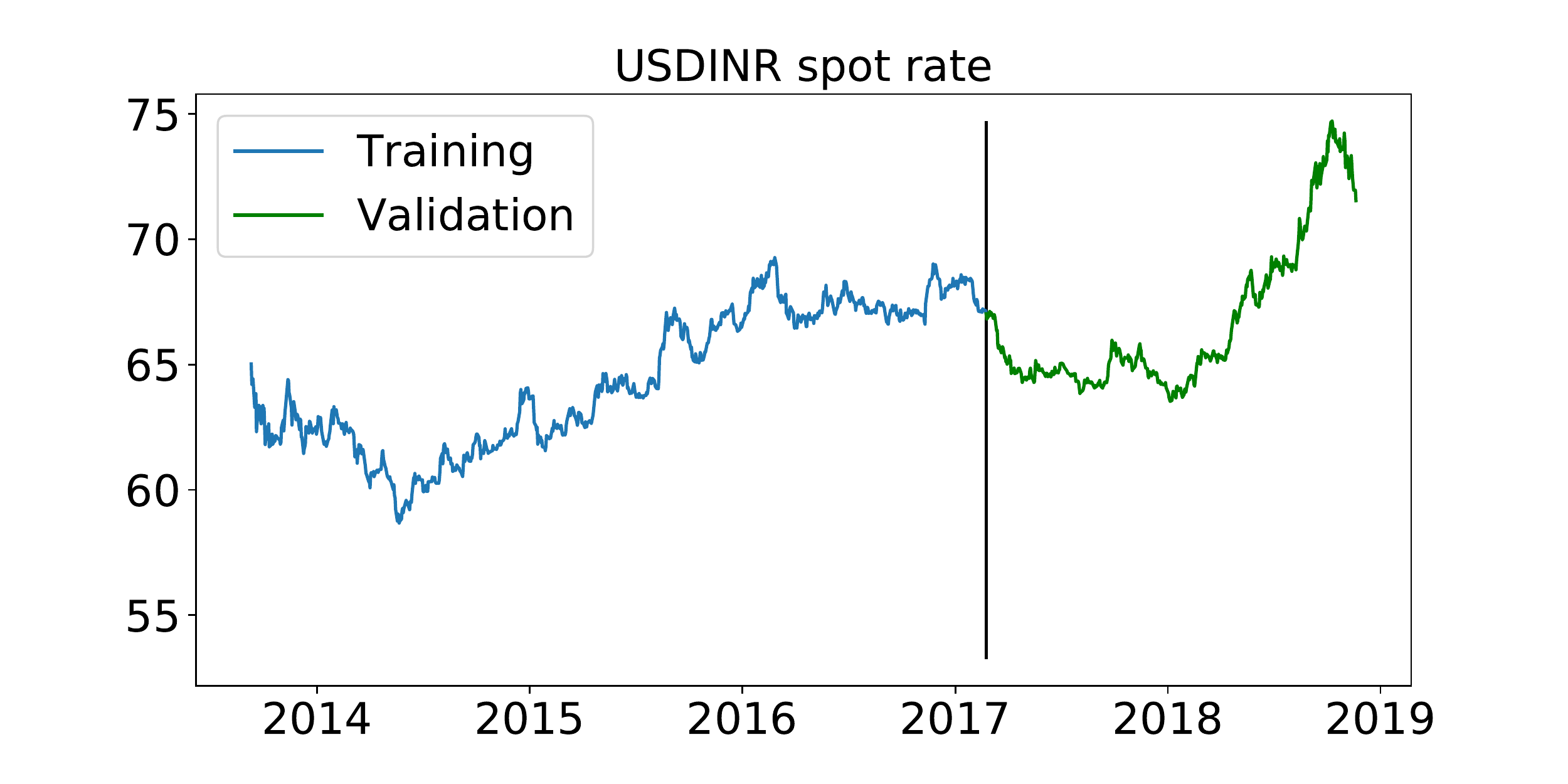}
    \label{fig:spots}
    \caption{Training and test periods for NDF pairs}
\end{figure}

\begin{figure}[!h]
   \centering
       
     \includegraphics[scale=0.33]{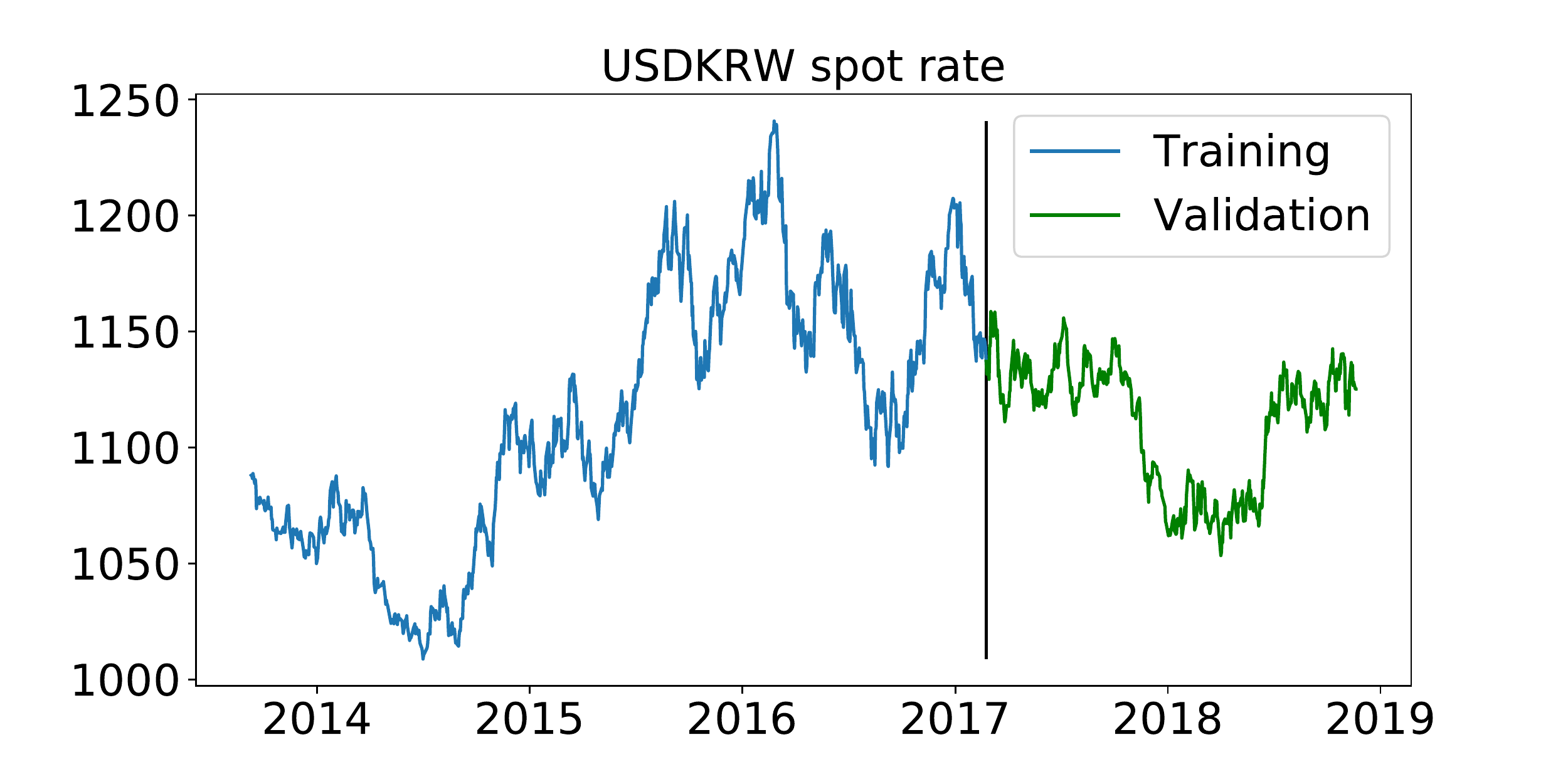}
    \includegraphics[scale=0.33]{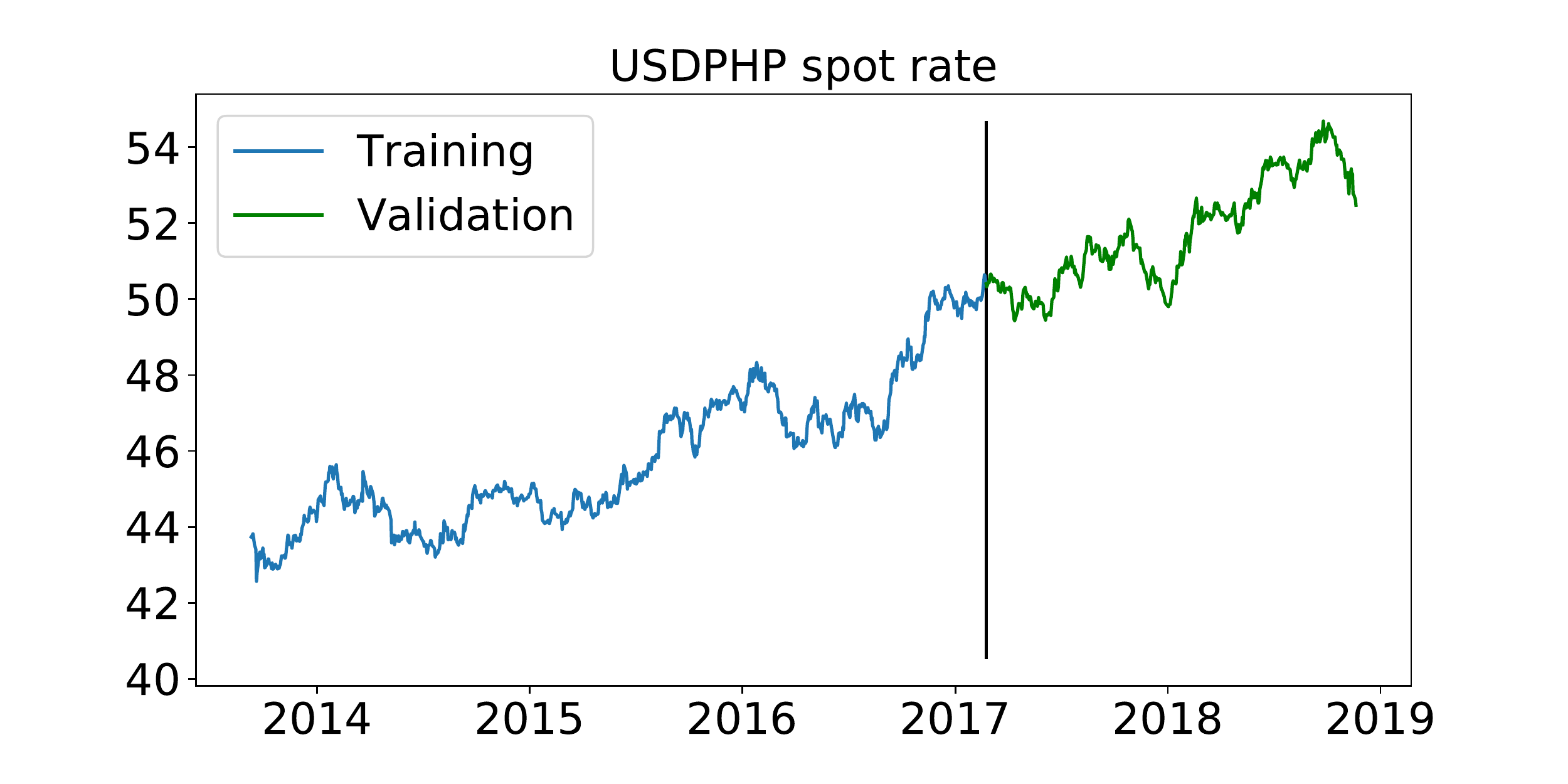}
    \includegraphics[scale=0.33]{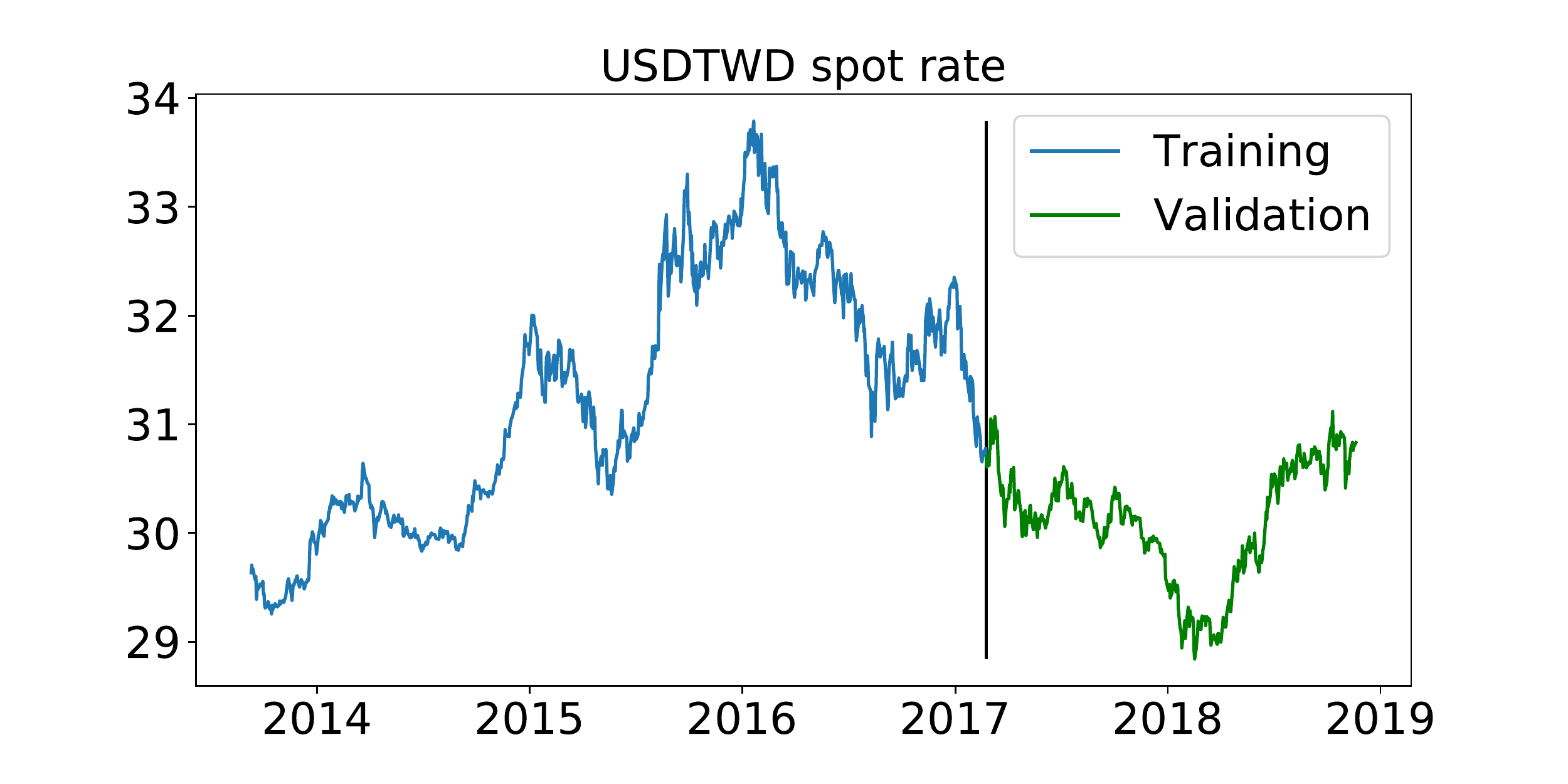}
     \label{fig:spots}
    \caption{Training and test periods for NDF pairs}
\end{figure}  

\begin{figure}[!h]
    \centering
    \includegraphics[scale=0.21]{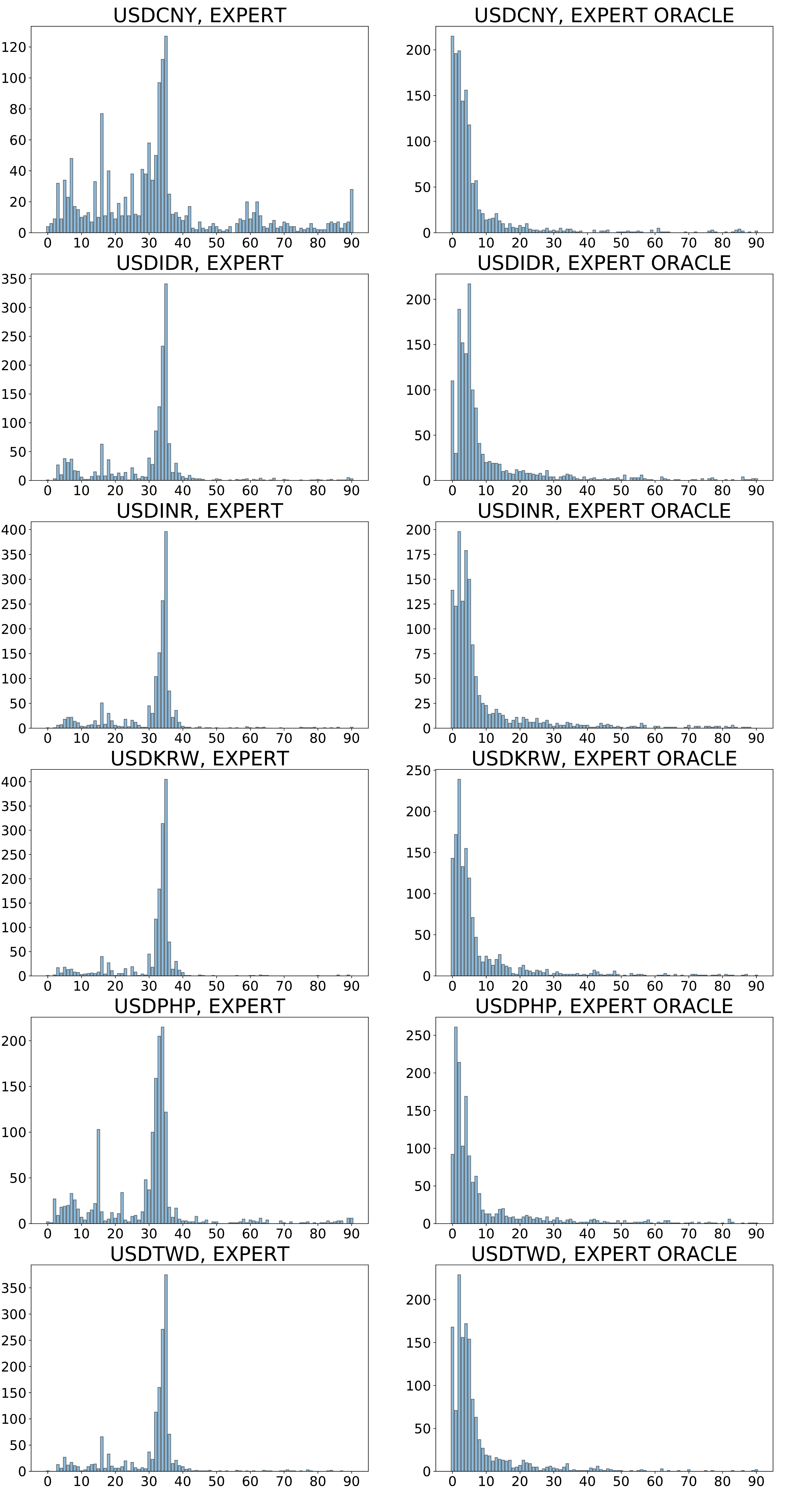}
    \caption{Distribution of tenor actions for \textit{Expert} and \textit{Expert Oracle}.}
    \label{fig:dists}
\end{figure}

\begin{figure}[!h]
   \centering
   \includegraphics[scale=0.2]{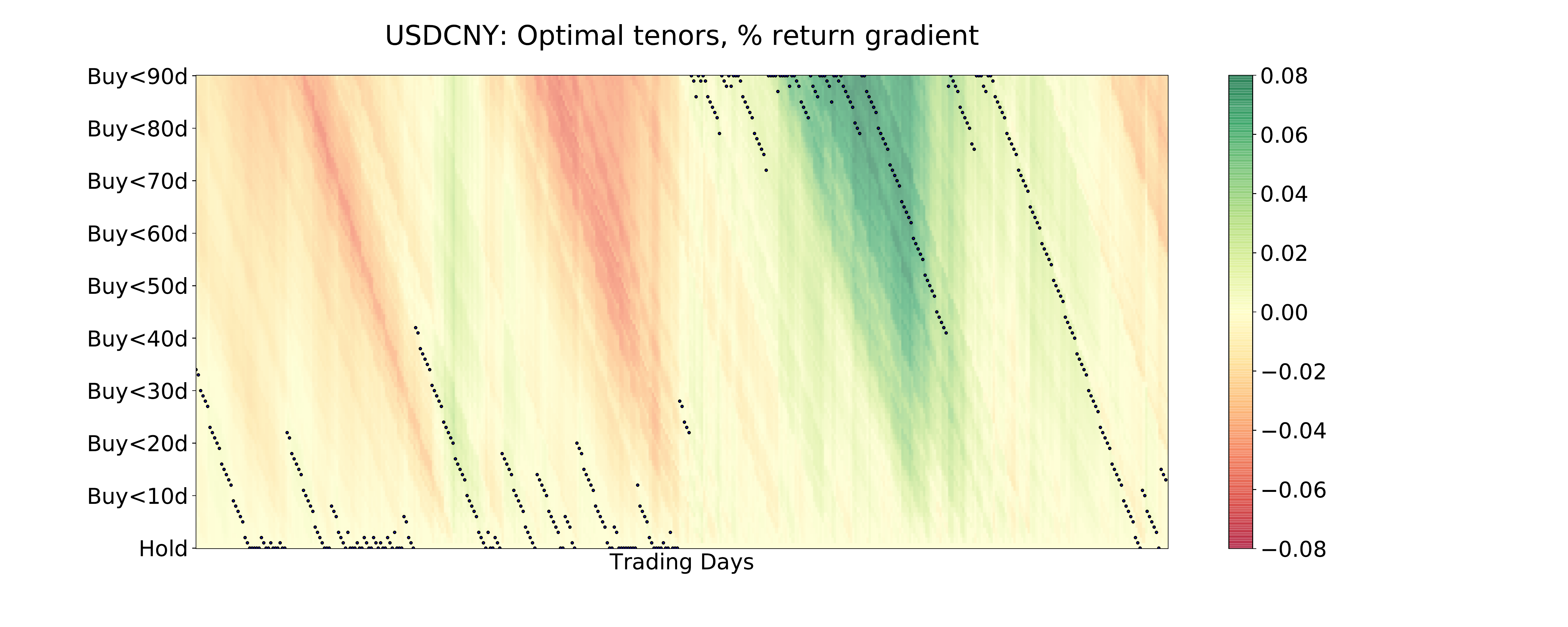}
    \includegraphics[scale=0.2]{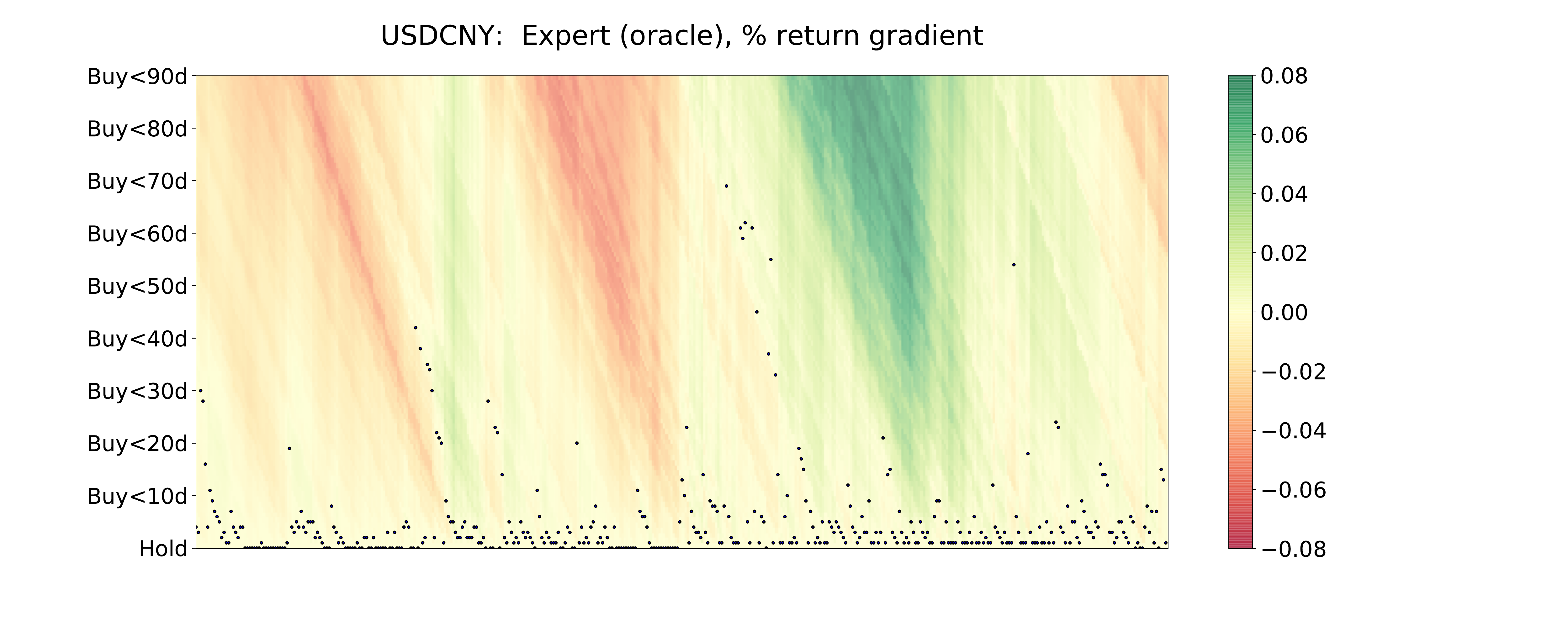}
    \includegraphics[scale=0.2]{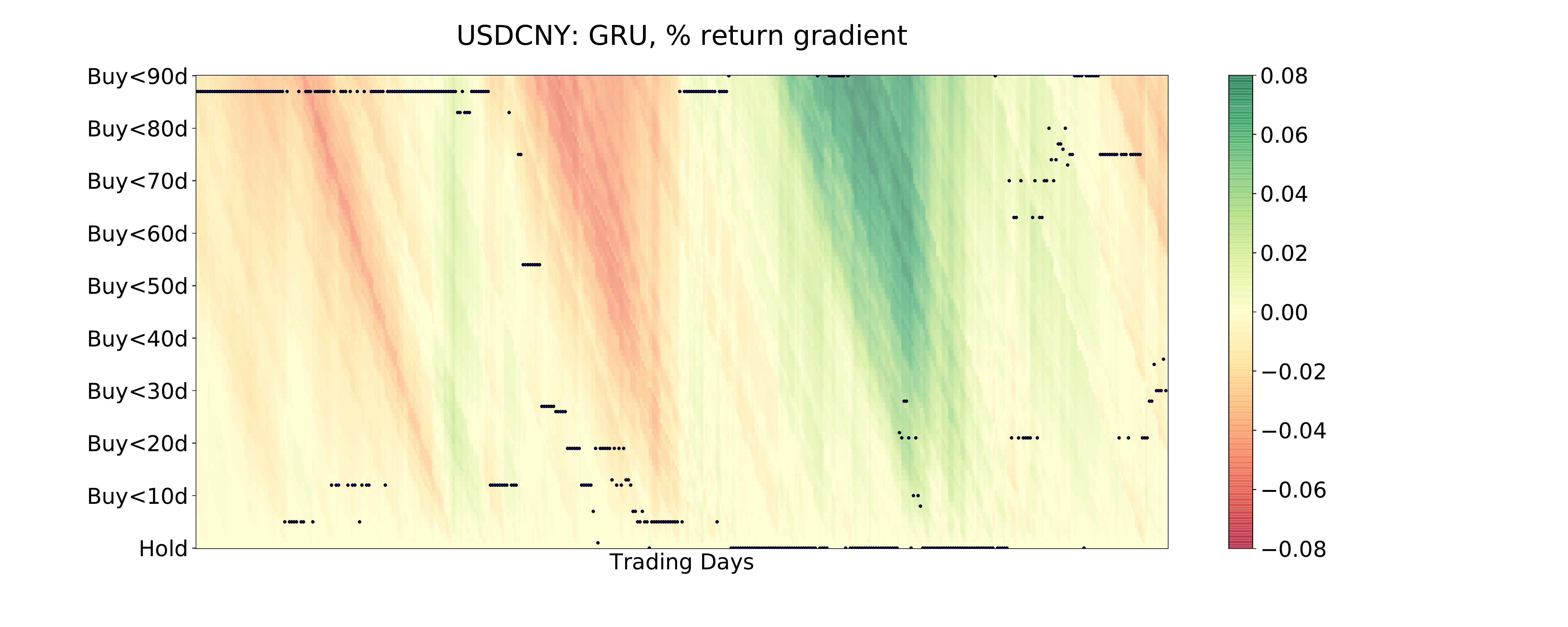}
    \includegraphics[scale=0.2]{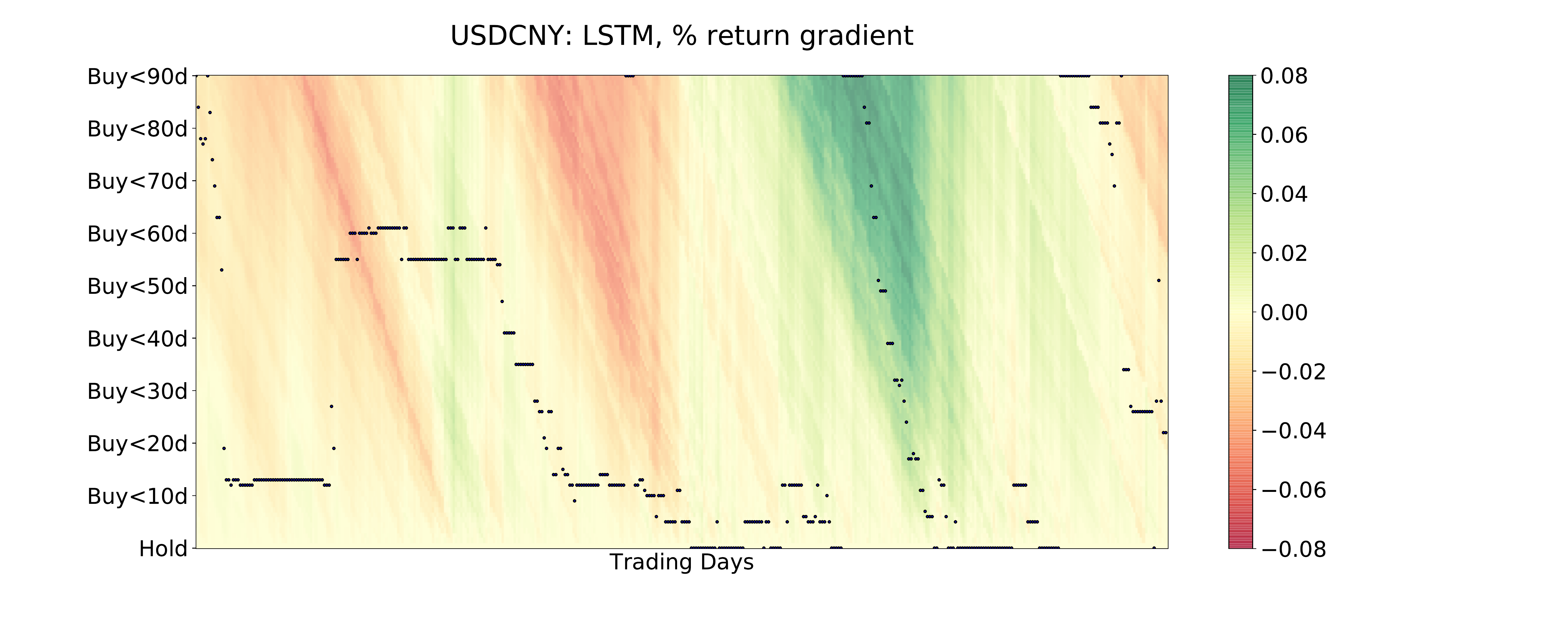}
    \includegraphics[scale=0.2]{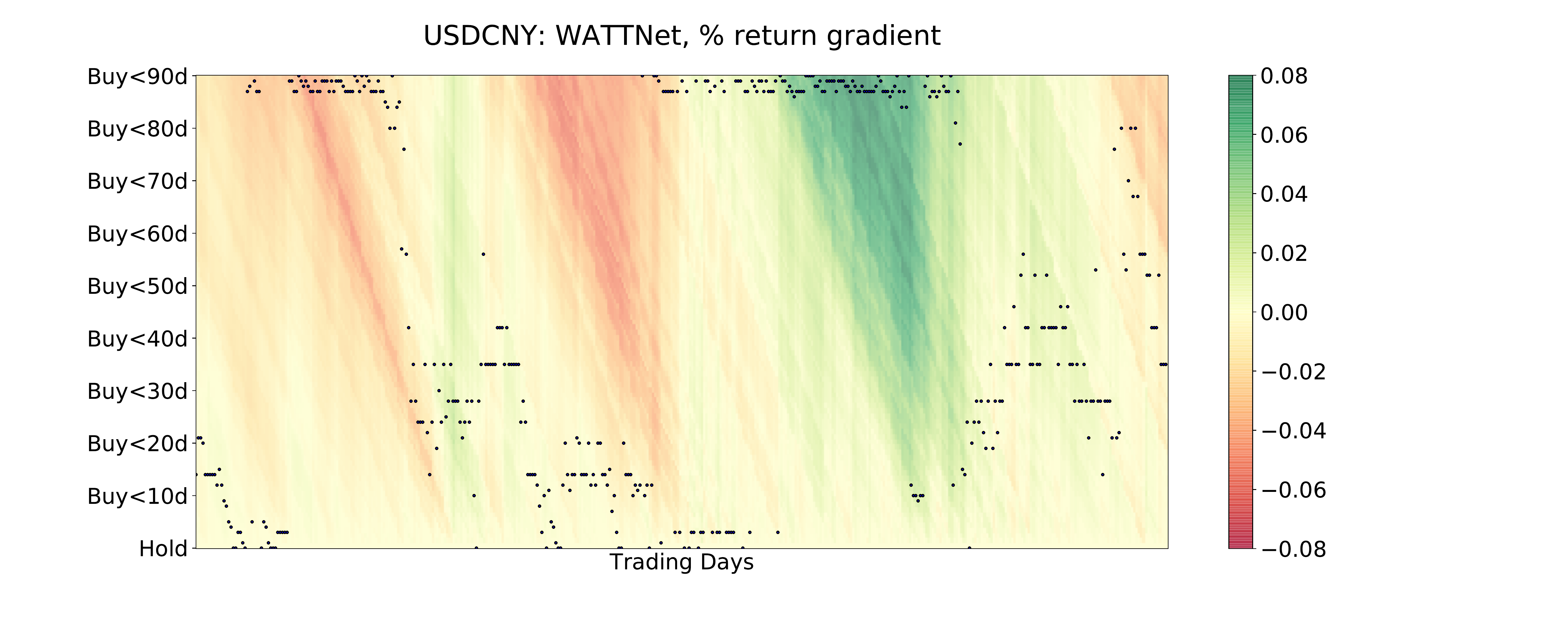}
    \label{fig:umap}
    \caption{USDCNY tenor actions}
\end{figure}
\begin{figure}[!h]
   \centering
   \includegraphics[scale=0.2]{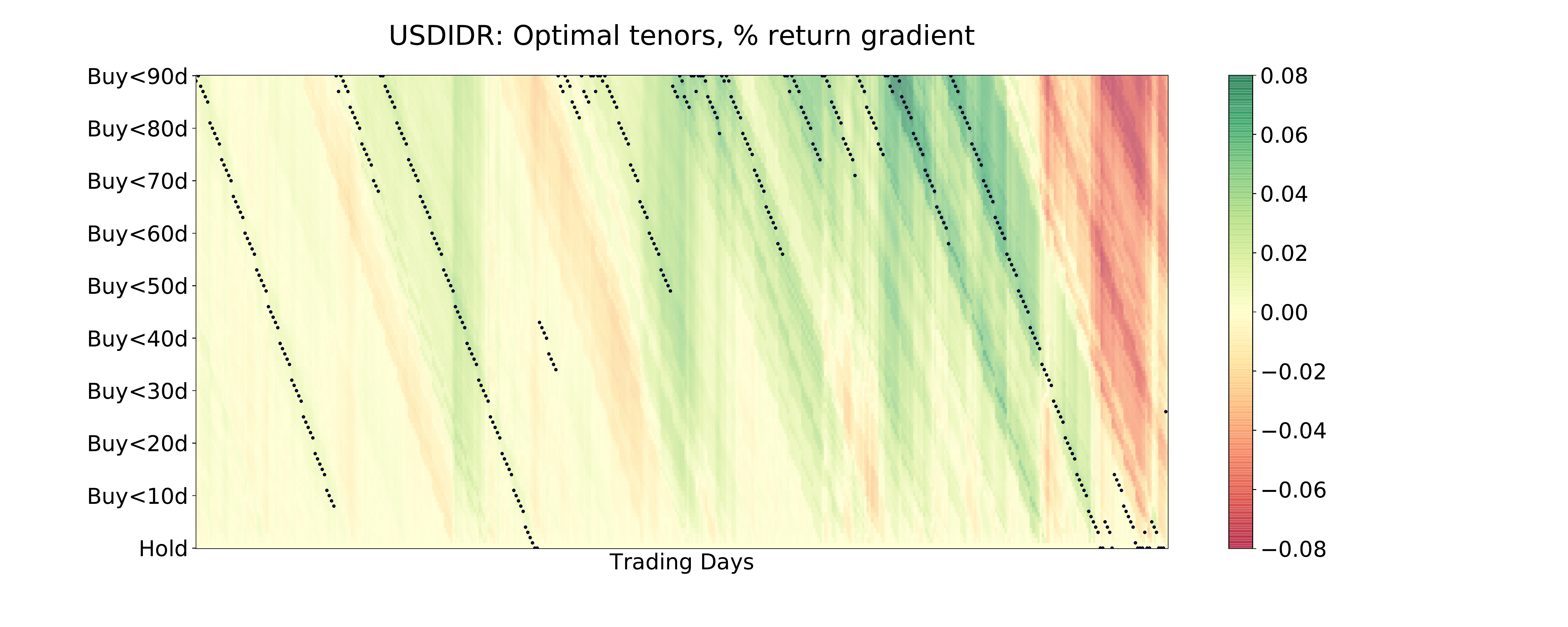}
    \includegraphics[scale=0.2]{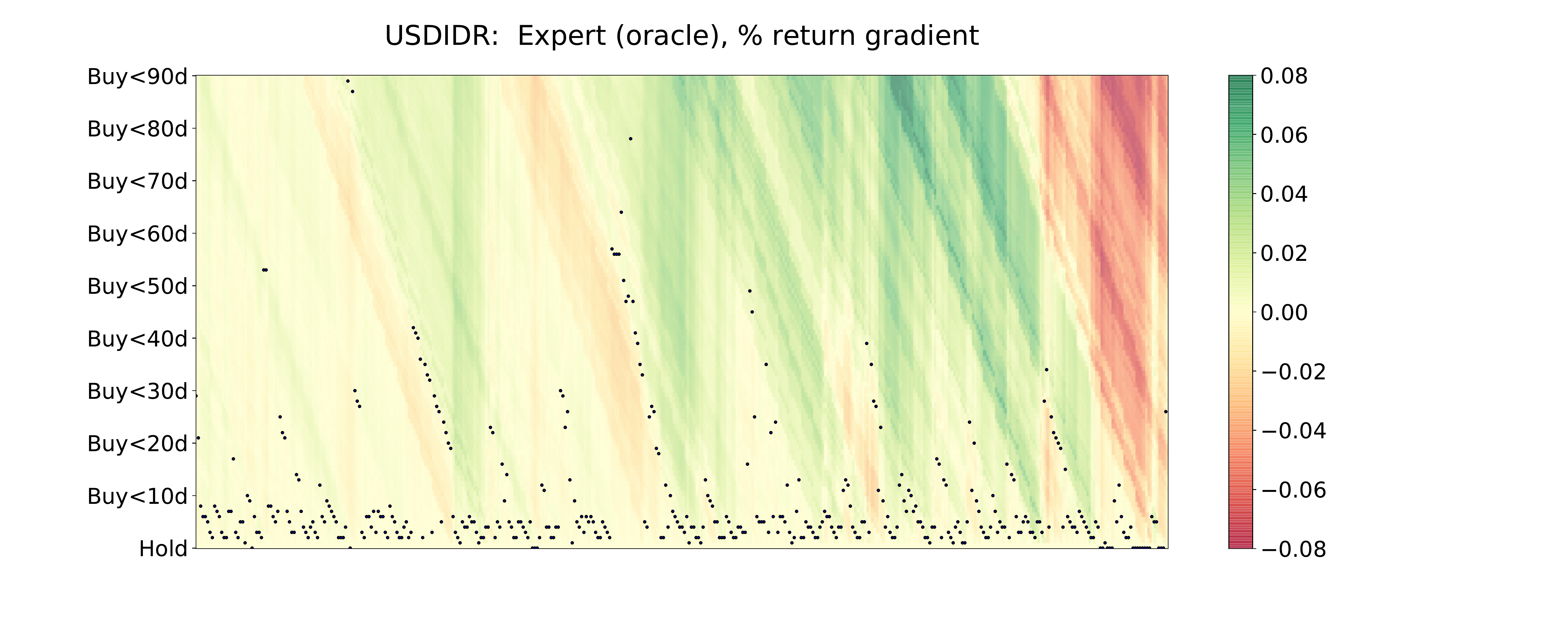}
    \includegraphics[scale=0.2]{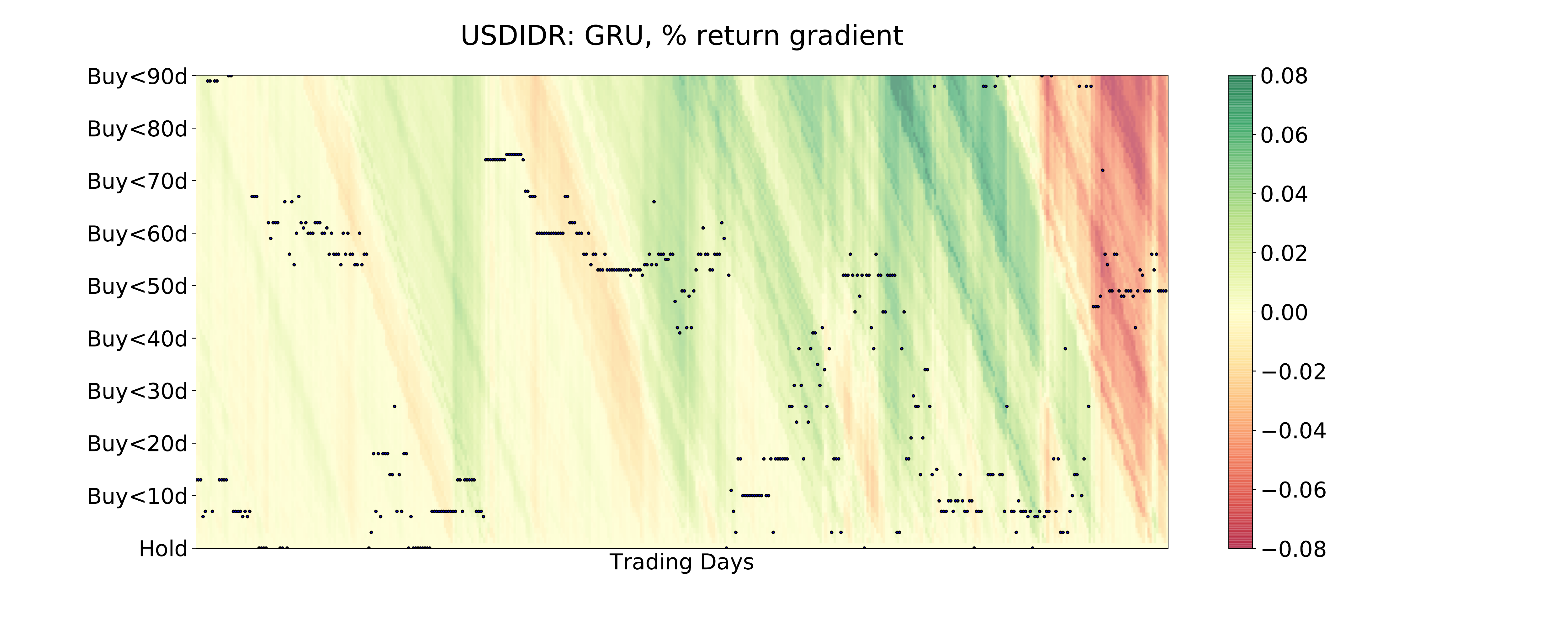}
    \includegraphics[scale=0.2]{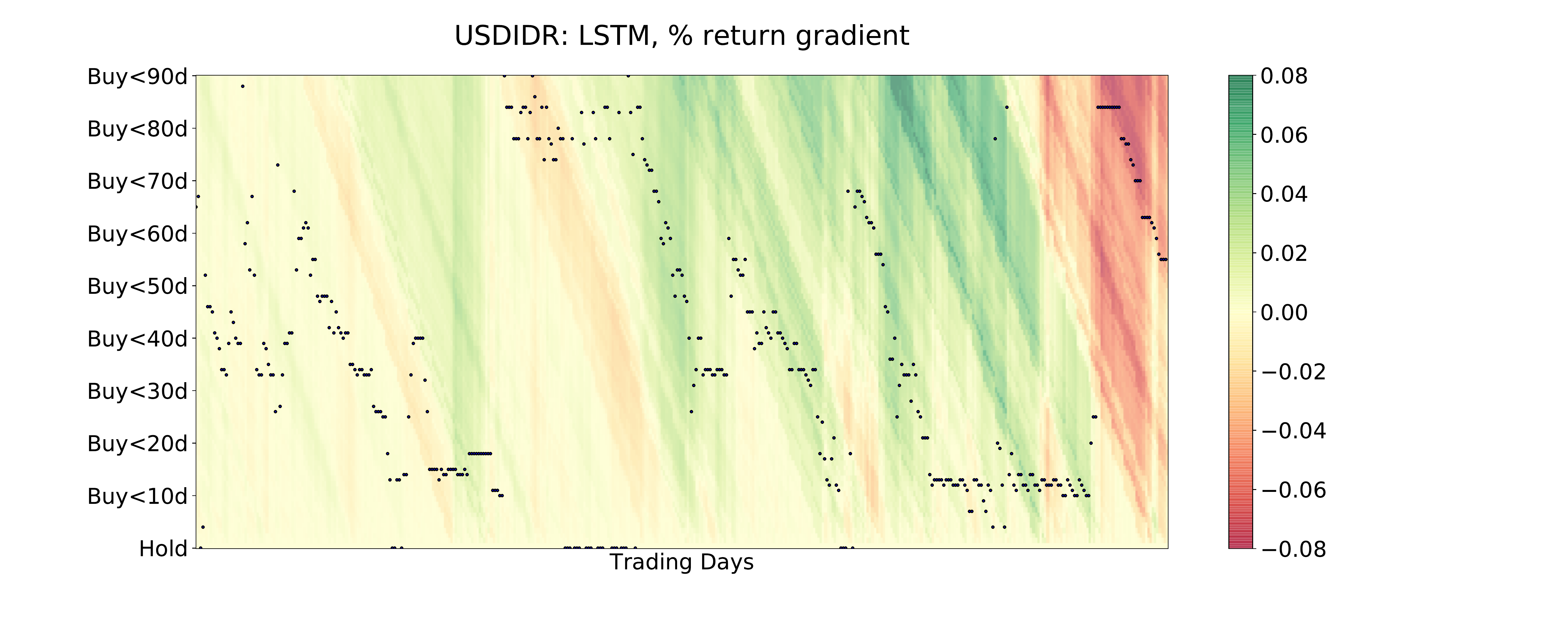}
    \includegraphics[scale=0.2]{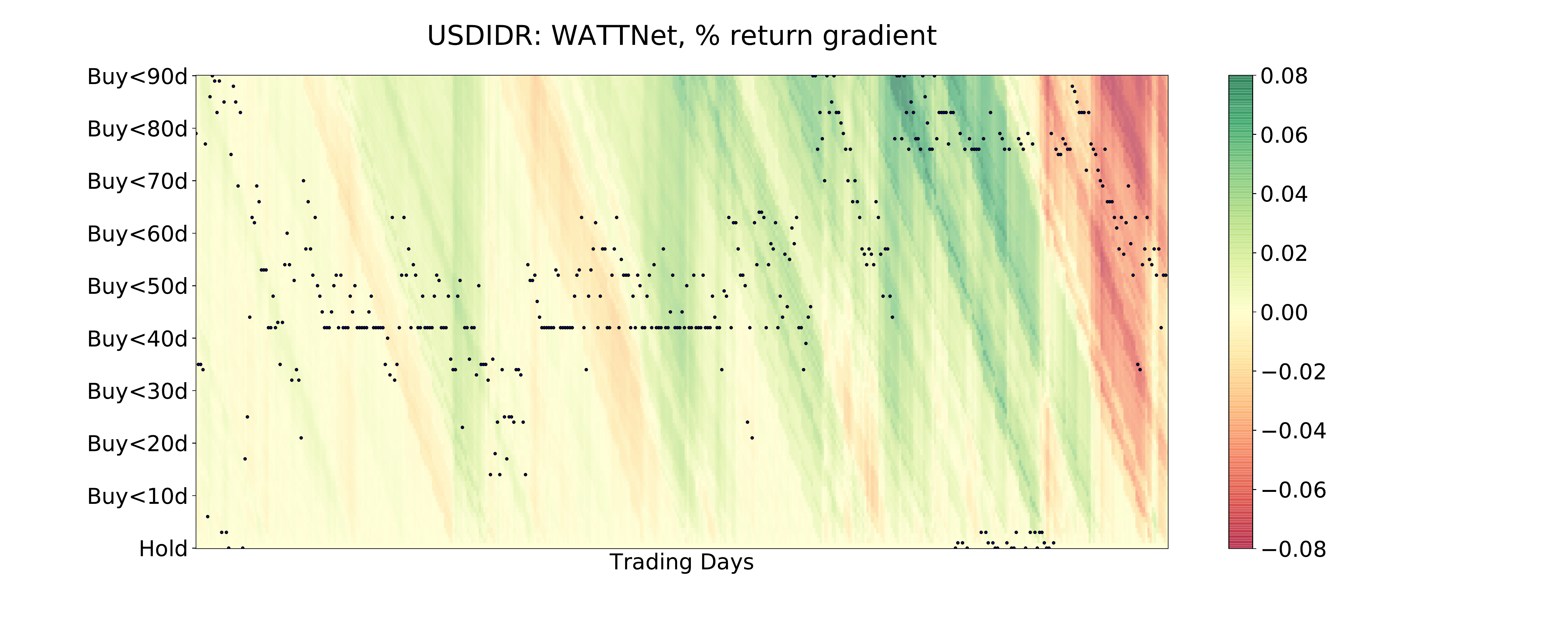}
    \label{fig:umap}
    \caption{USDIDR tenor actions}
\end{figure}
\begin{figure}[!h]
   \centering
   \includegraphics[scale=0.2]{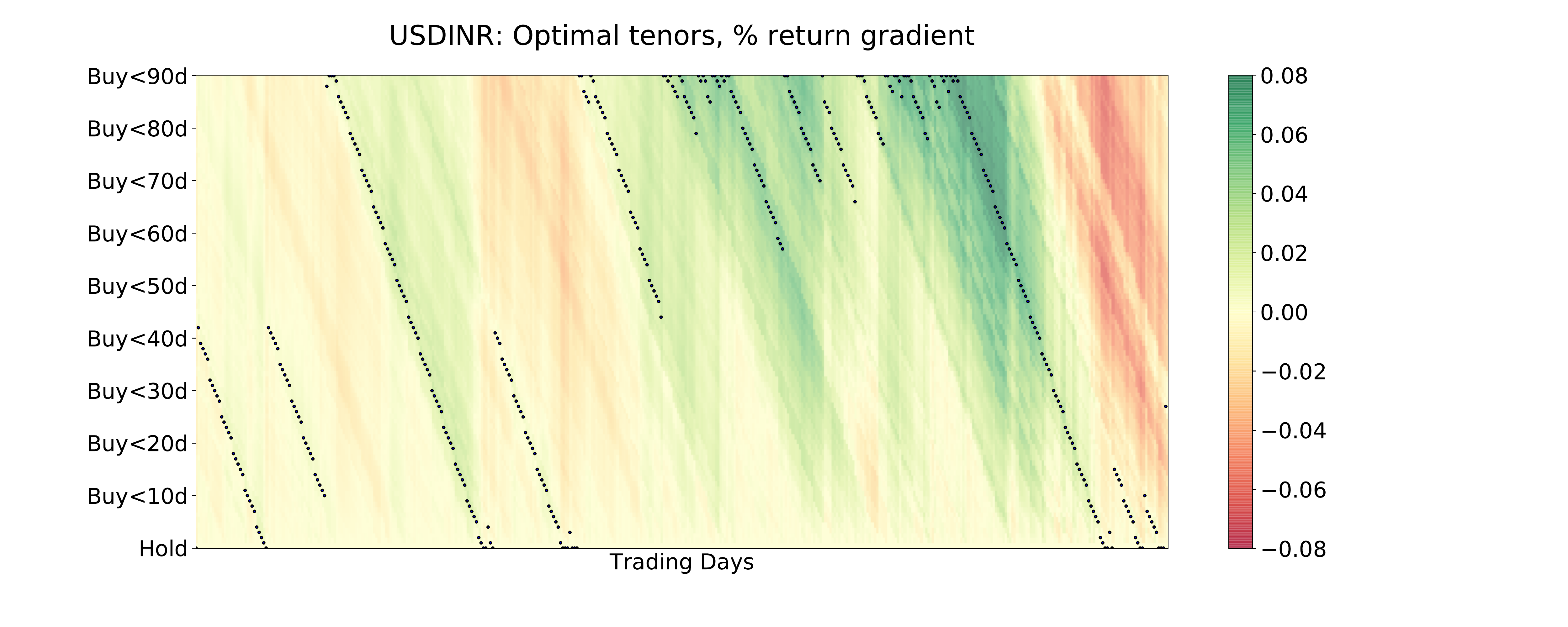}
    \includegraphics[scale=0.2]{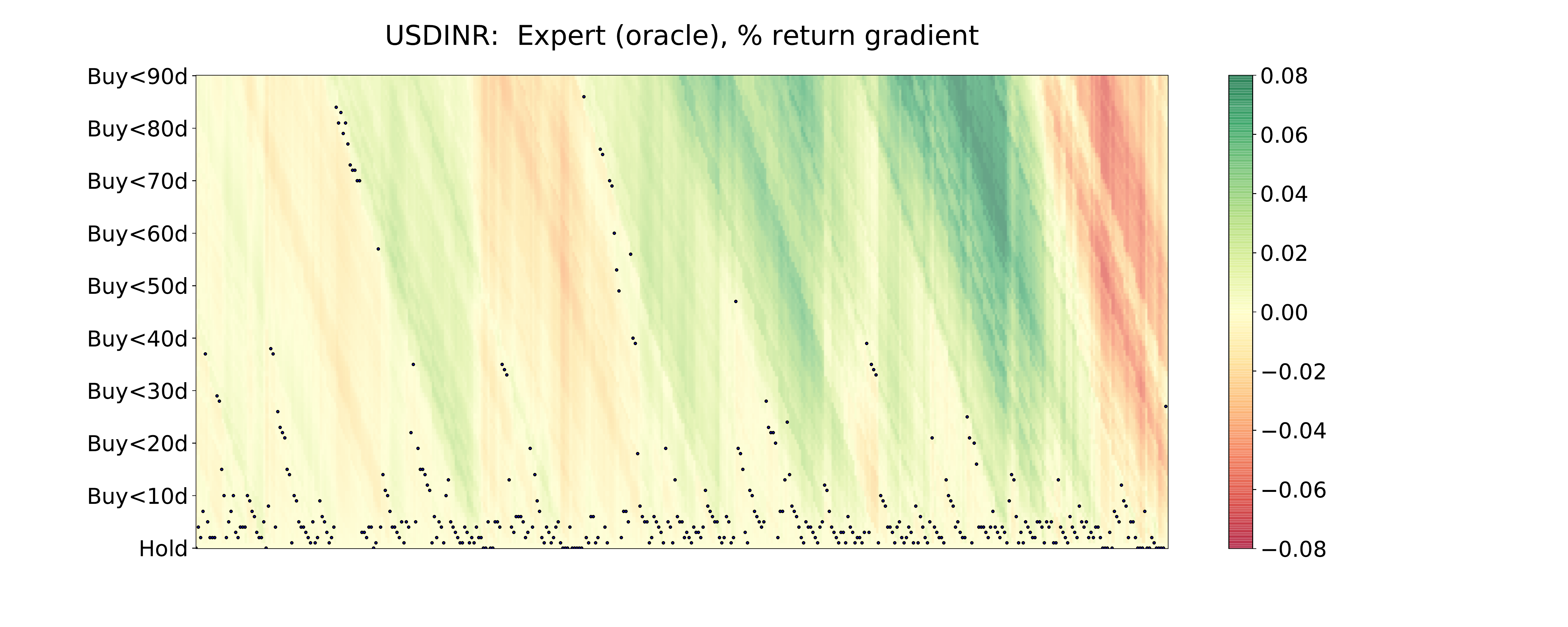}
    \includegraphics[scale=0.2]{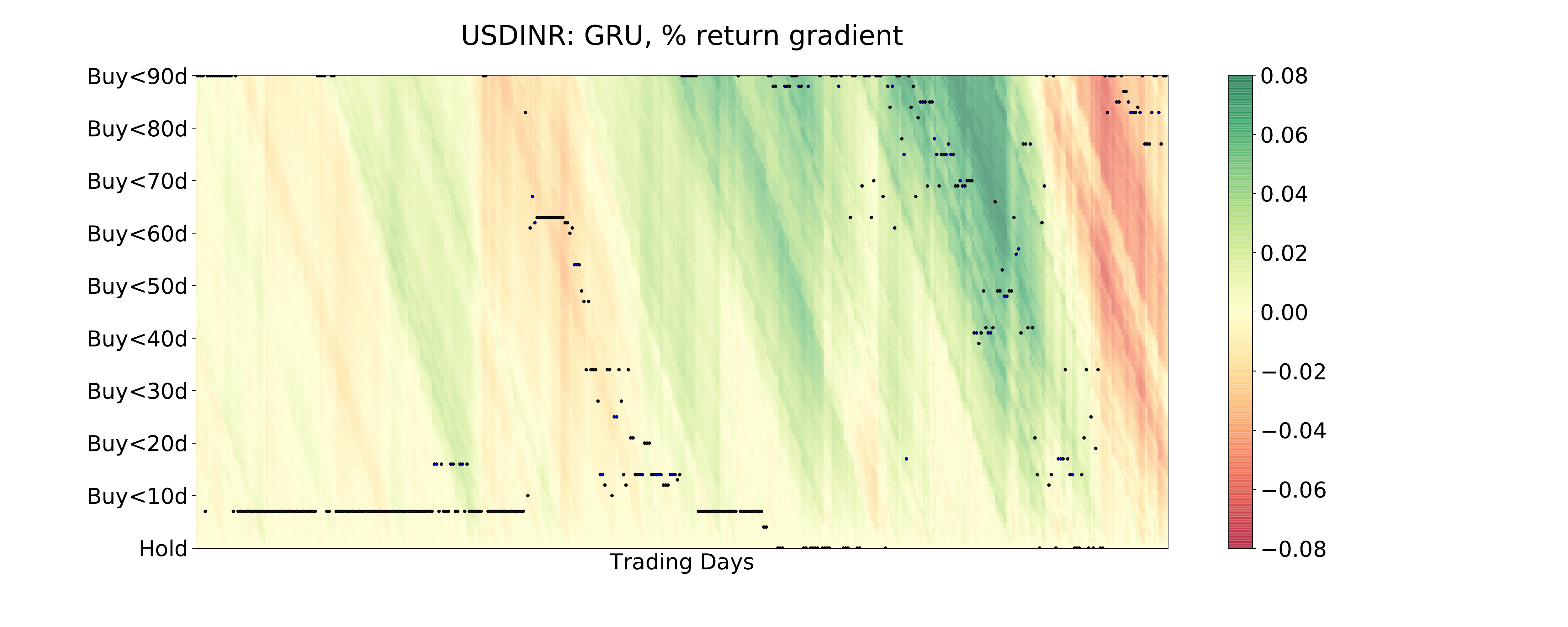}
    \includegraphics[scale=0.2]{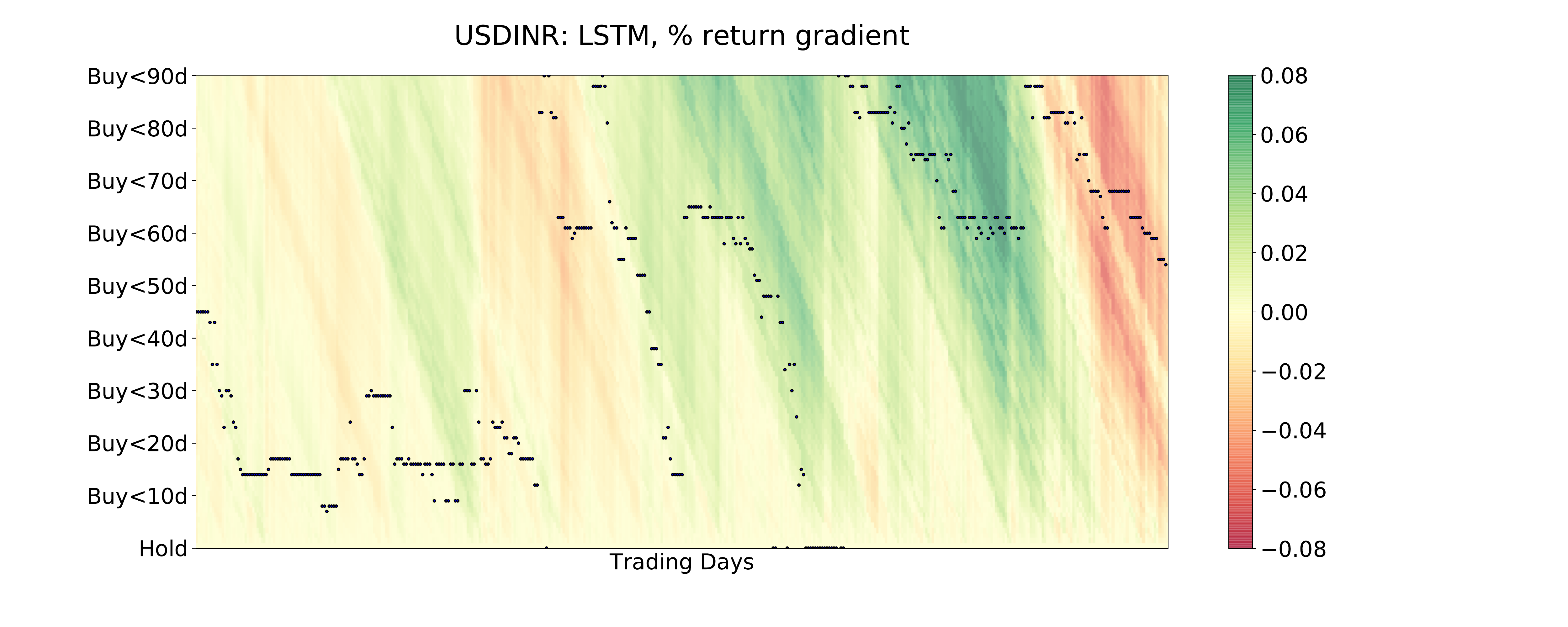}
    \includegraphics[scale=0.2]{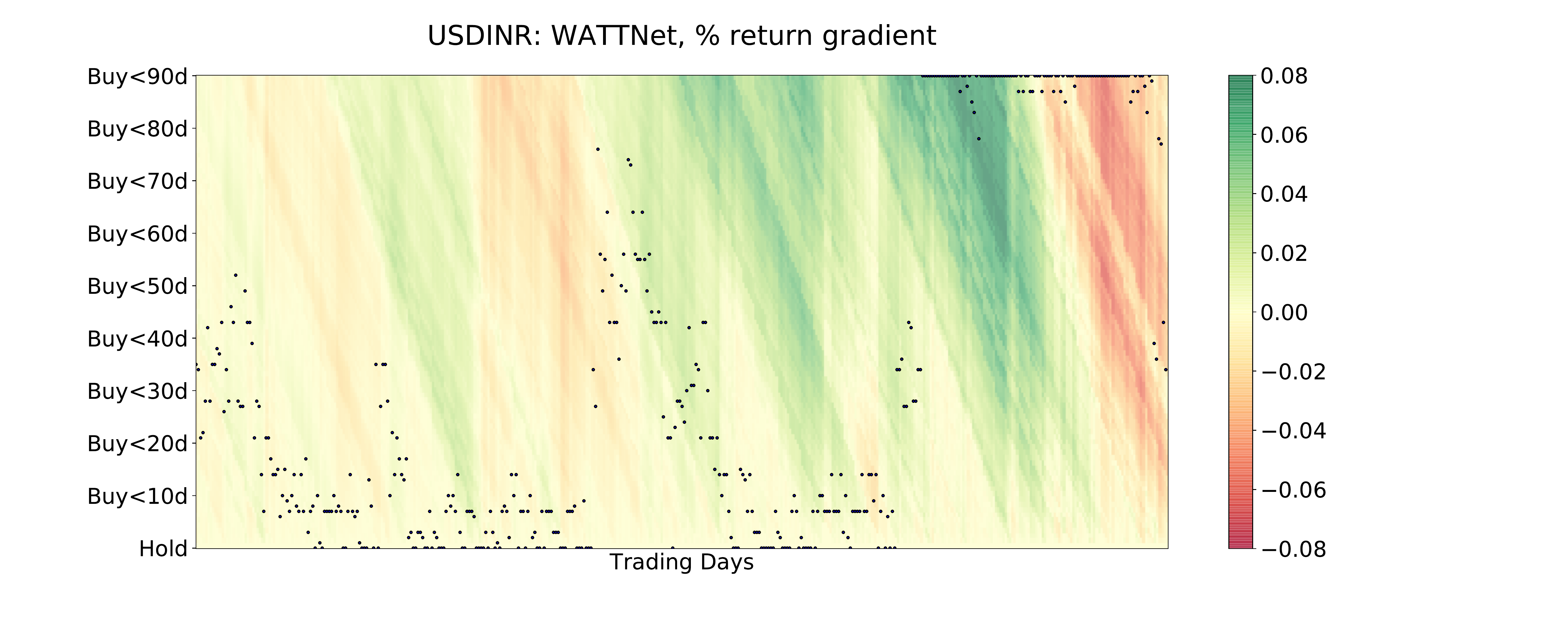}
    \label{fig:umap}
    \caption{USDINR tenor actions}
\end{figure}
\begin{figure}[!h]
   \centering
   \includegraphics[scale=0.2]{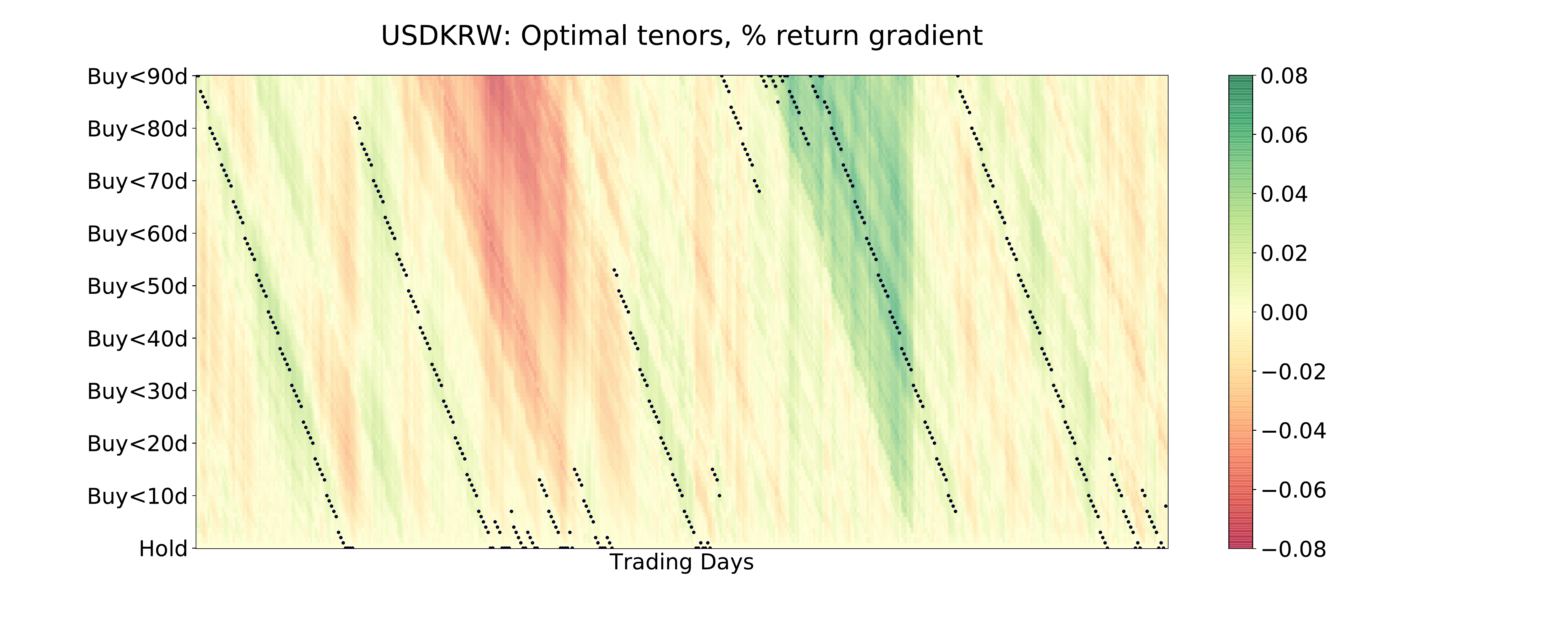}
    \includegraphics[scale=0.2]{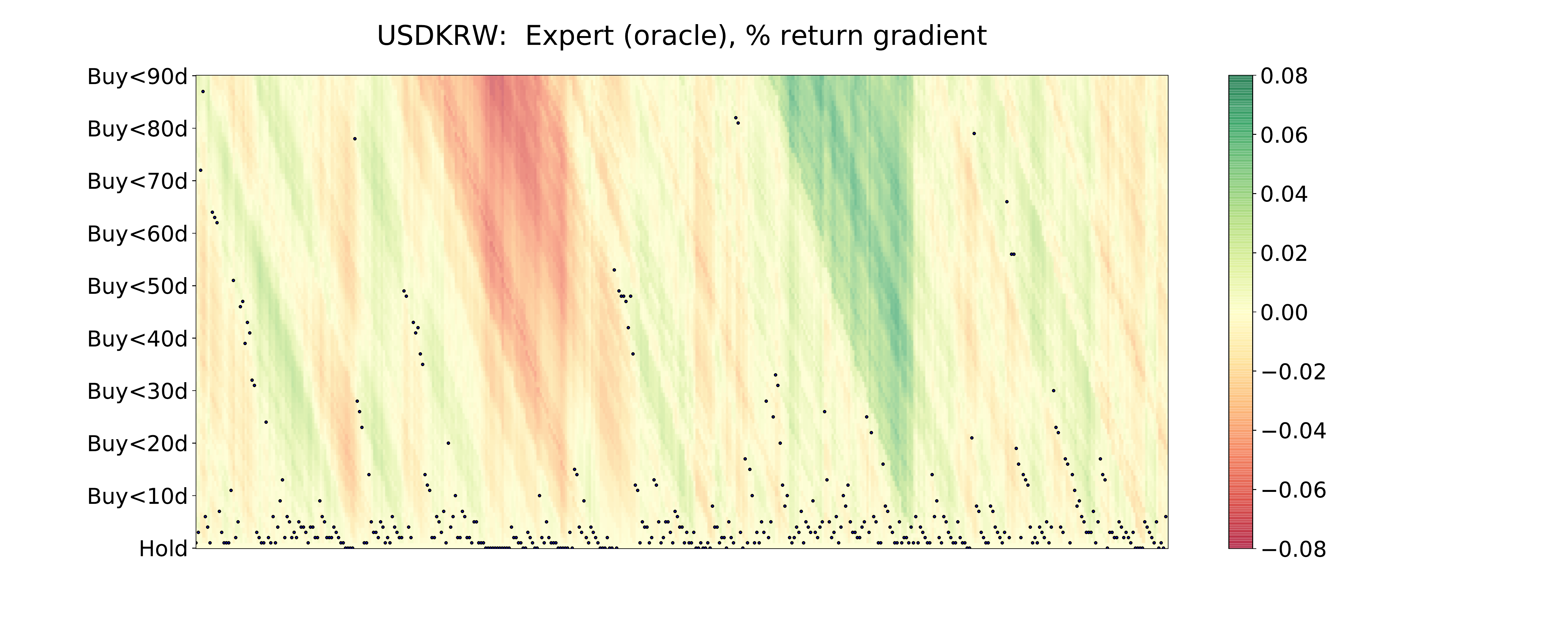}
    \includegraphics[scale=0.2]{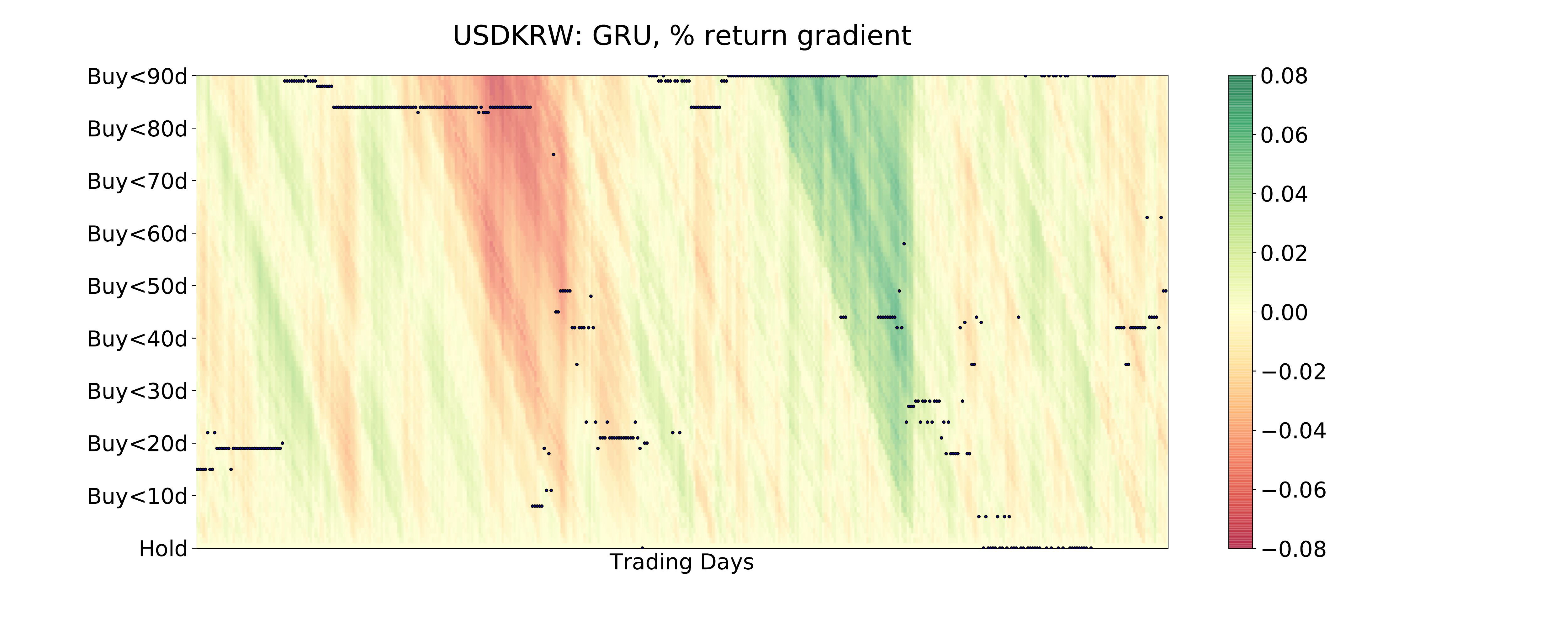}
    \includegraphics[scale=0.2]{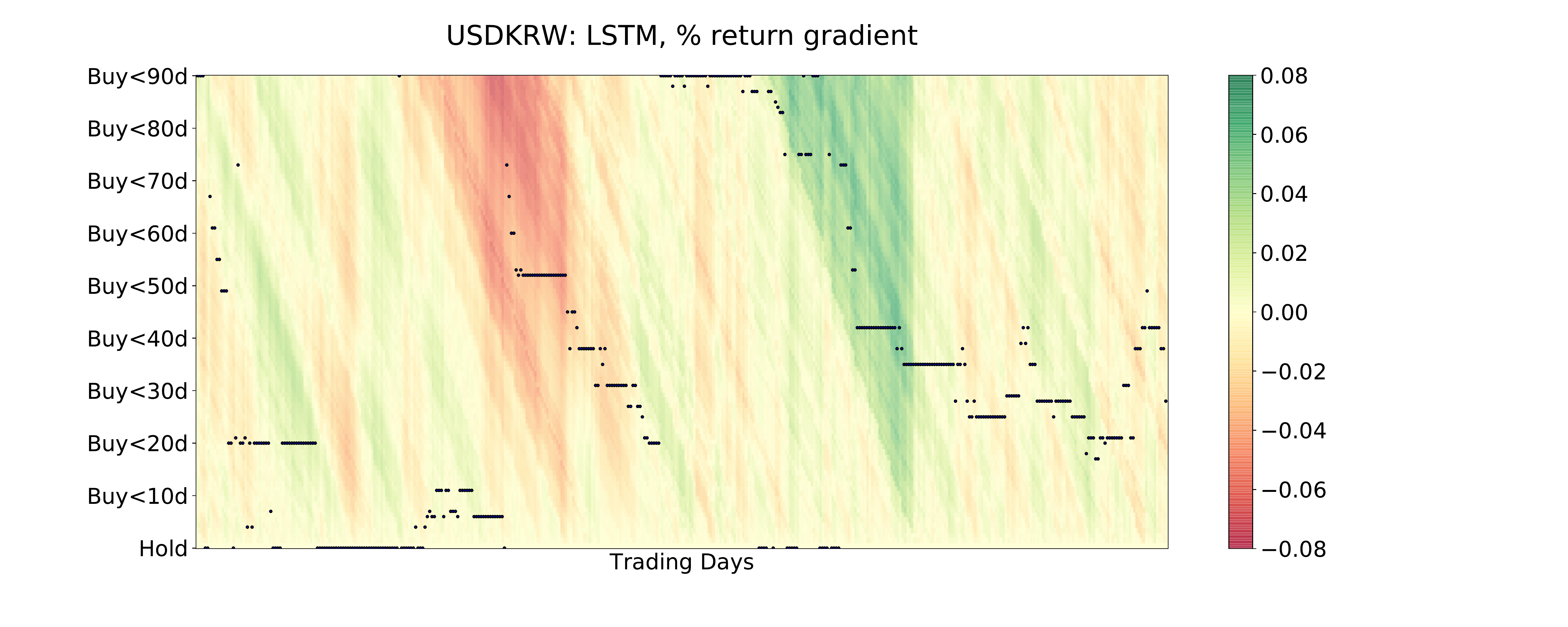}
    \includegraphics[scale=0.2]{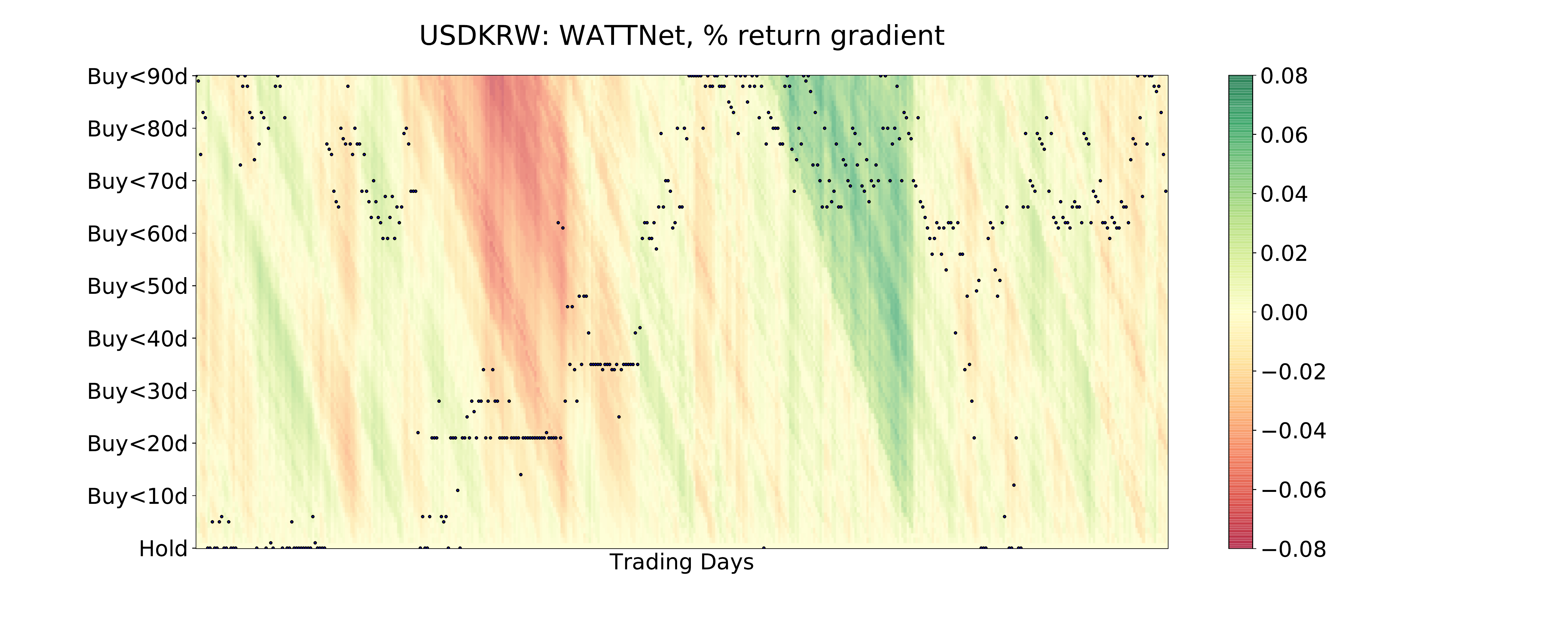}
    \label{fig:umap}
    \caption{USDKRW tenor actions}
\end{figure}
\begin{figure}[!h]
   \centering
   \includegraphics[scale=0.2]{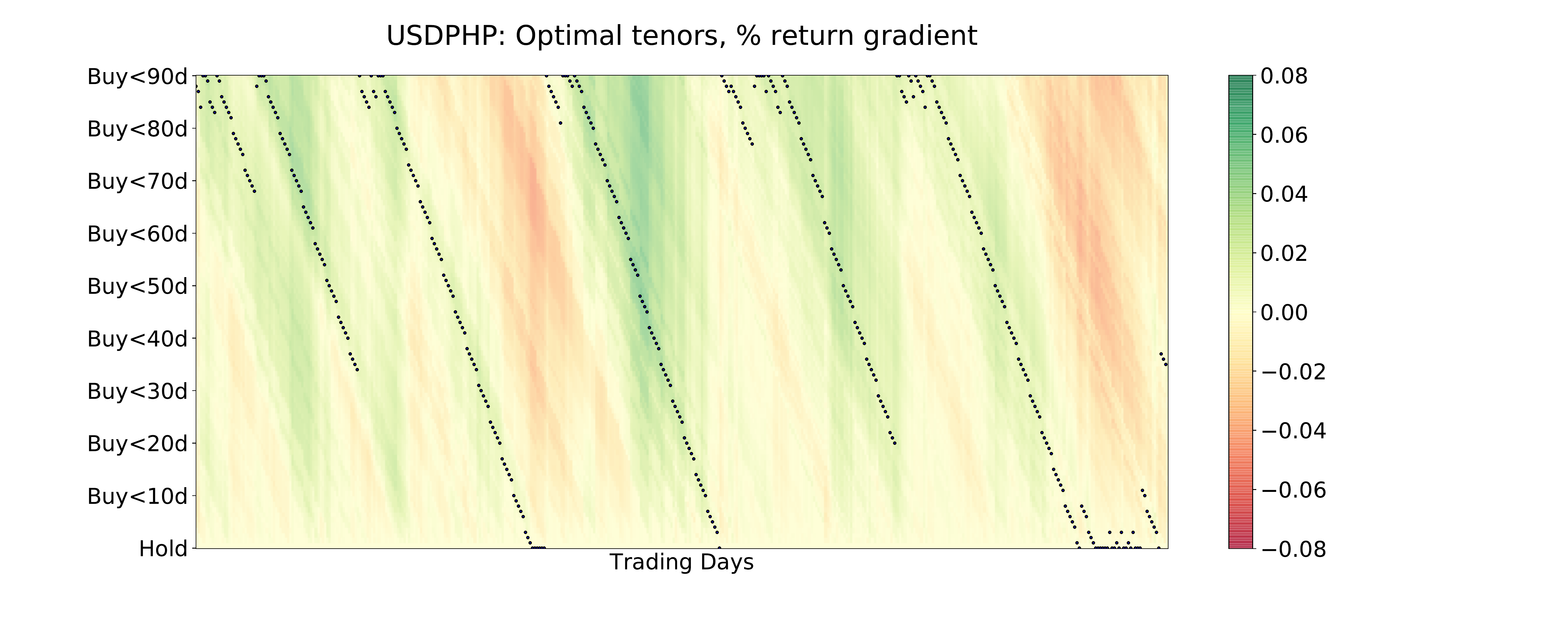}
    \includegraphics[scale=0.2]{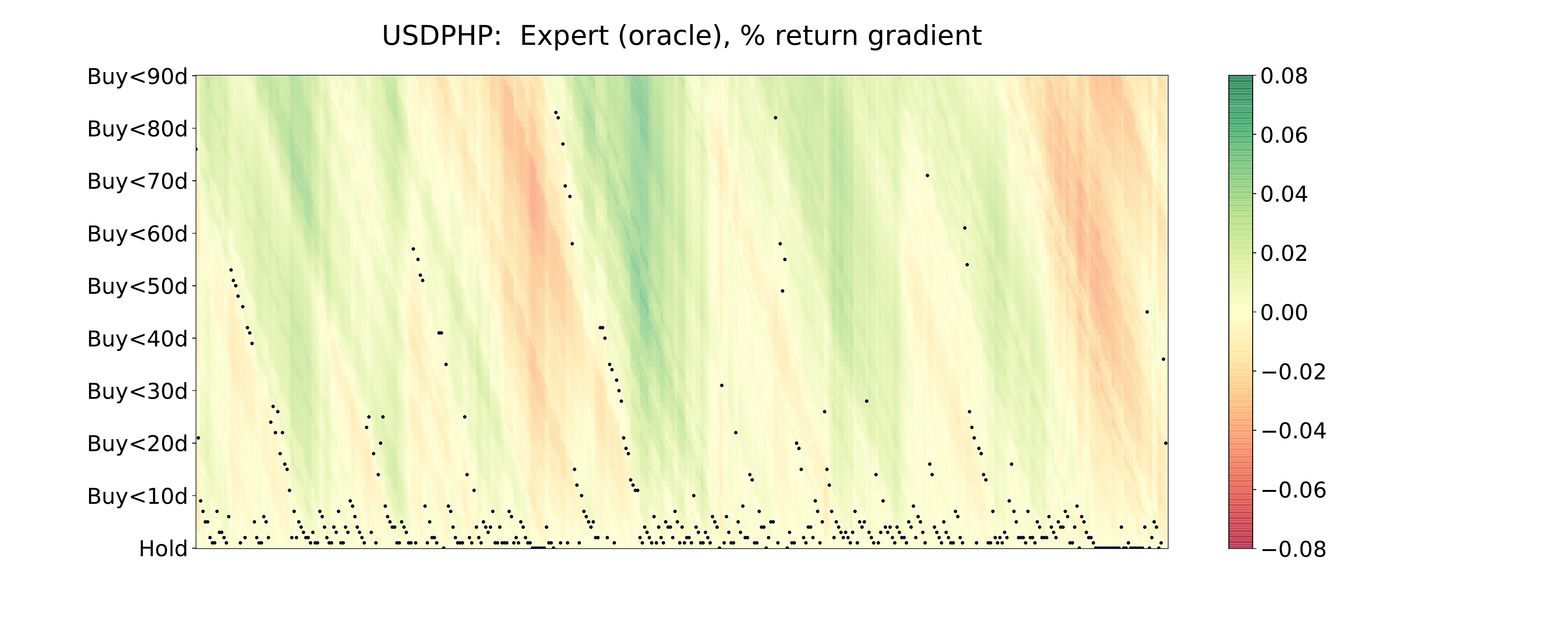}
    \includegraphics[scale=0.2]{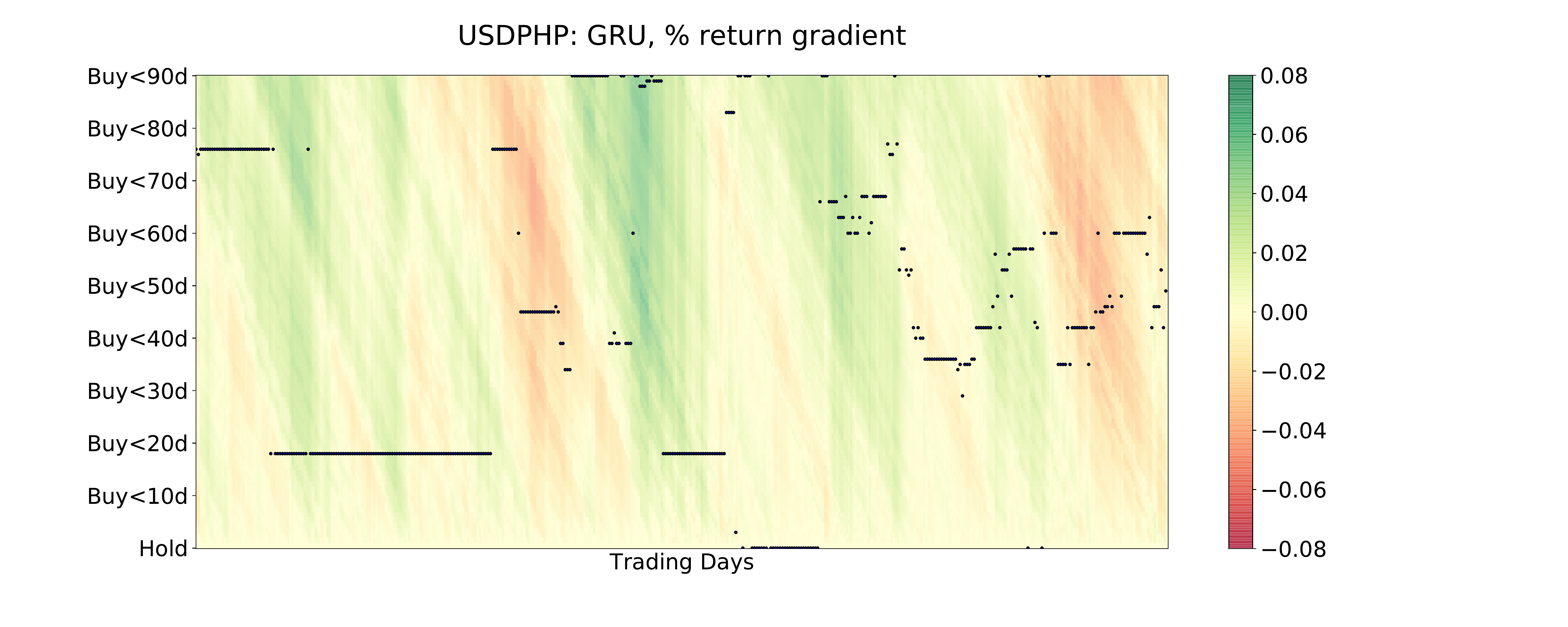}
    \includegraphics[scale=0.2]{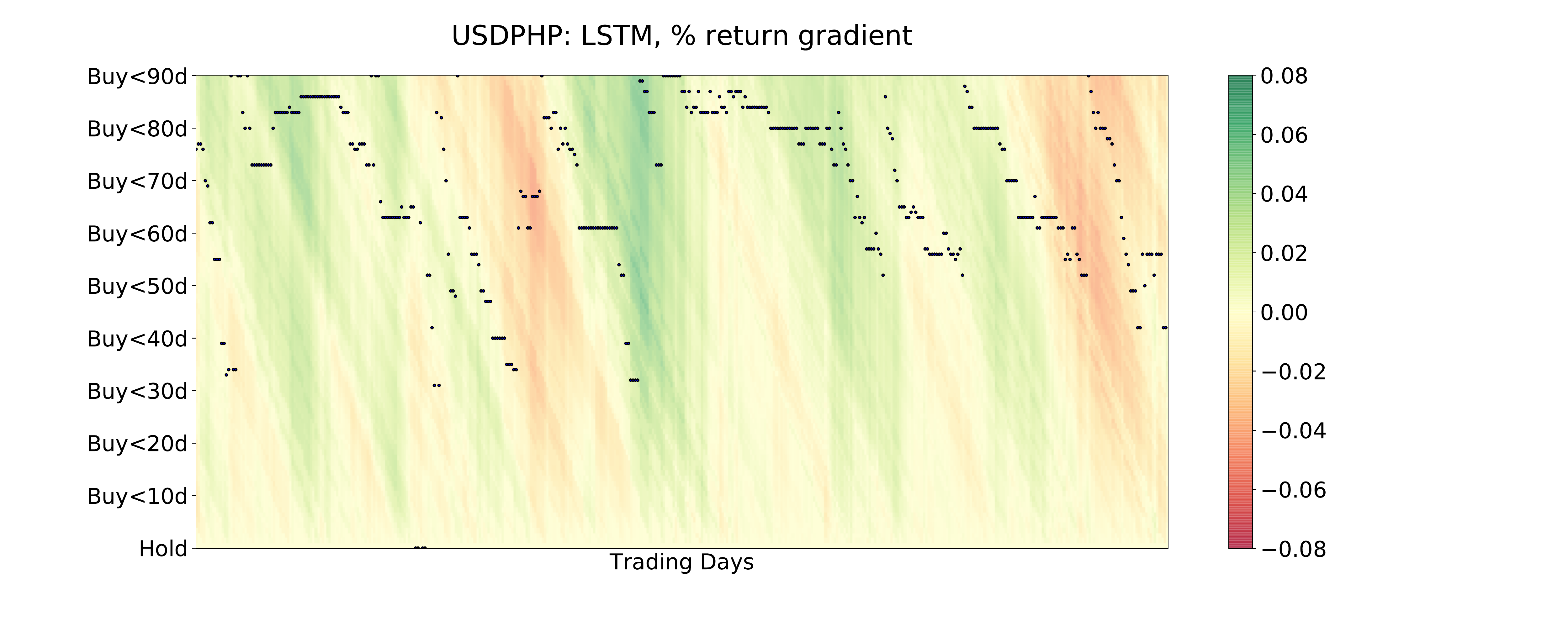}
    \includegraphics[scale=0.2]{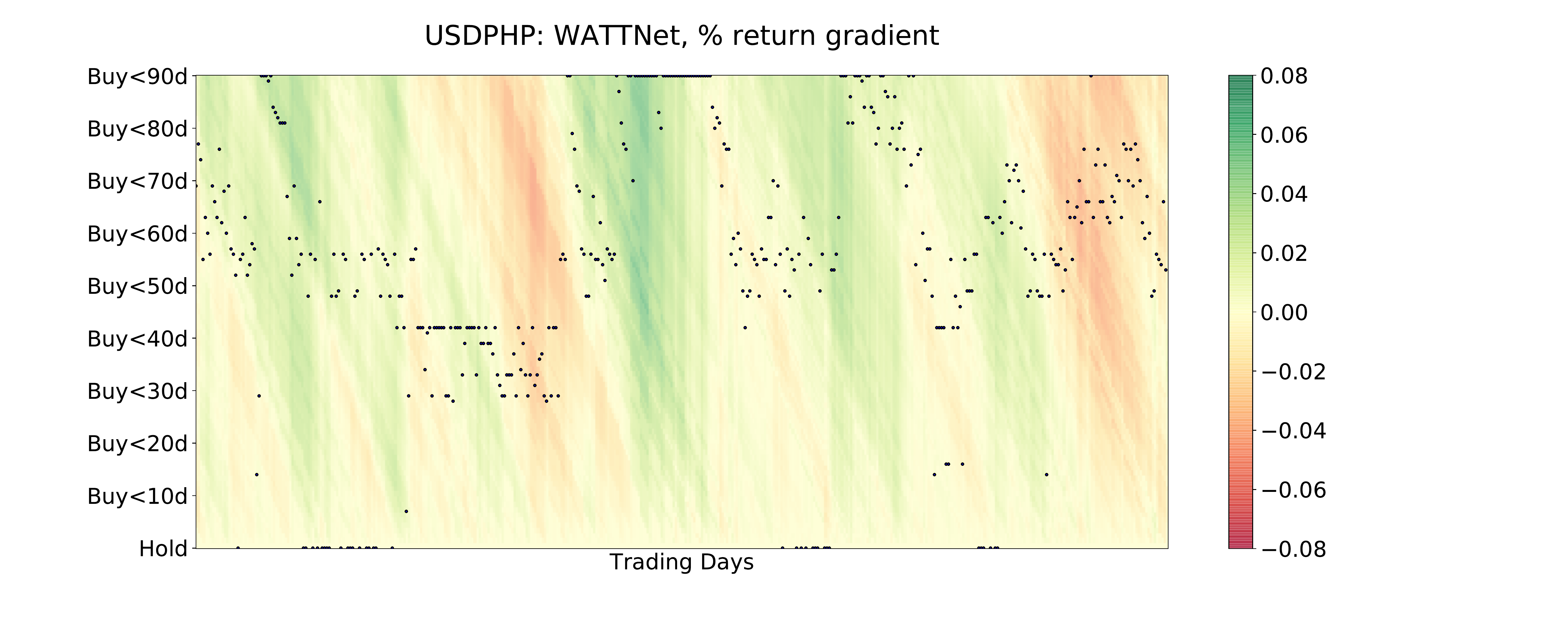}
    \label{fig:umap}
    \caption{USDPHP tenor actions}
\end{figure}
\begin{figure}[!h]
   \centering
   \includegraphics[scale=0.2]{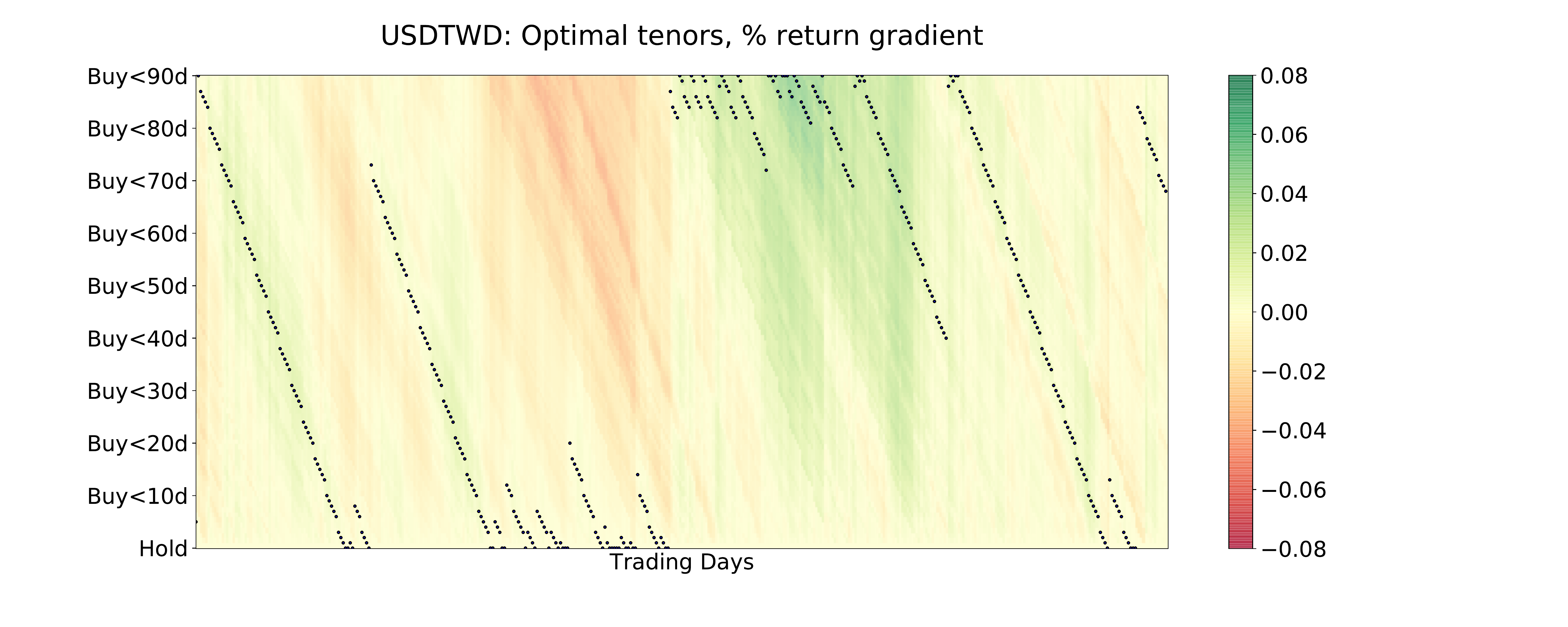}
    \includegraphics[scale=0.2]{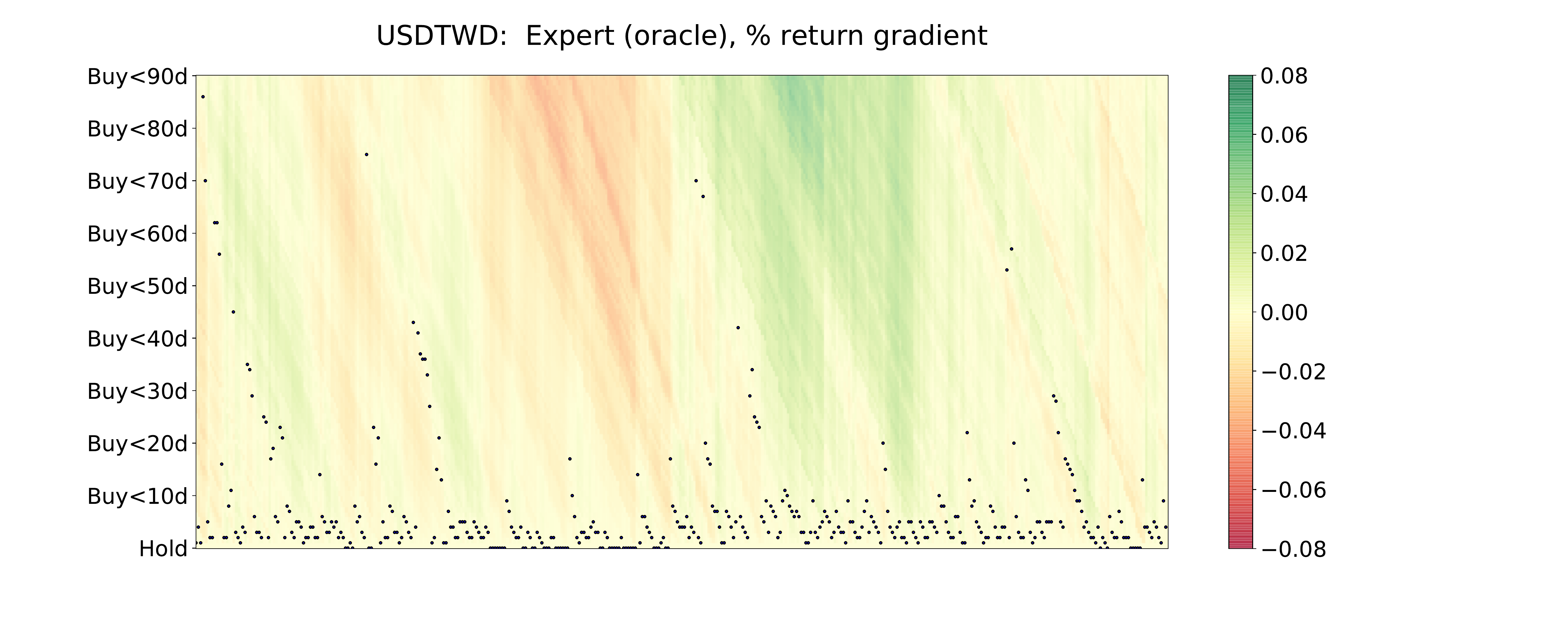}
    \includegraphics[scale=0.2]{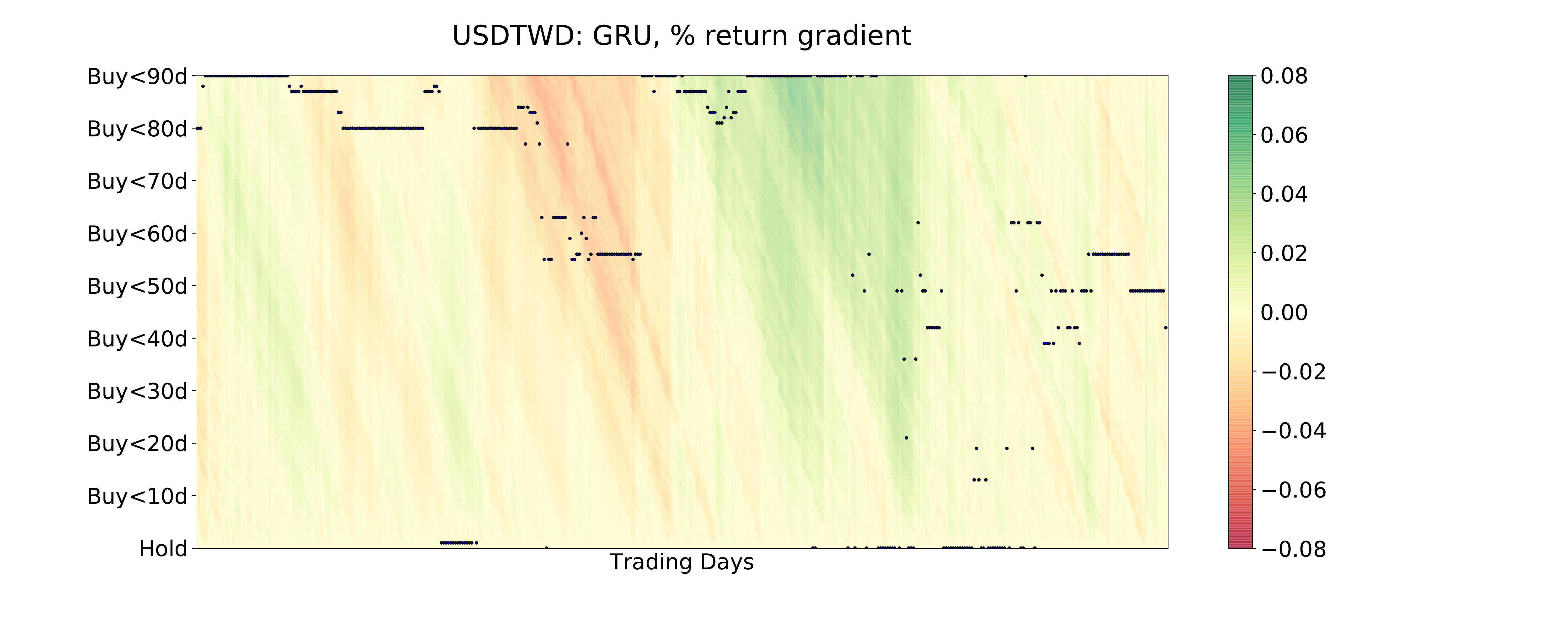}
    \includegraphics[scale=0.2]{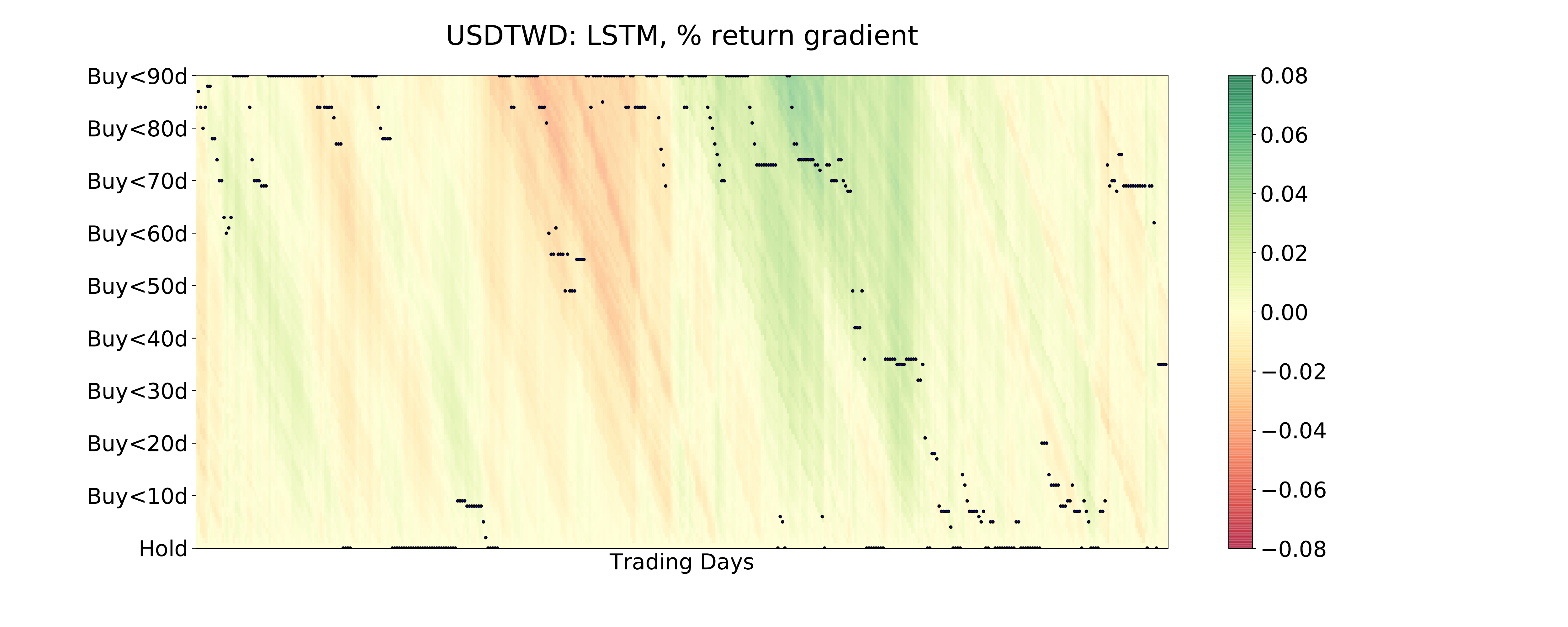}
    \includegraphics[scale=0.2]{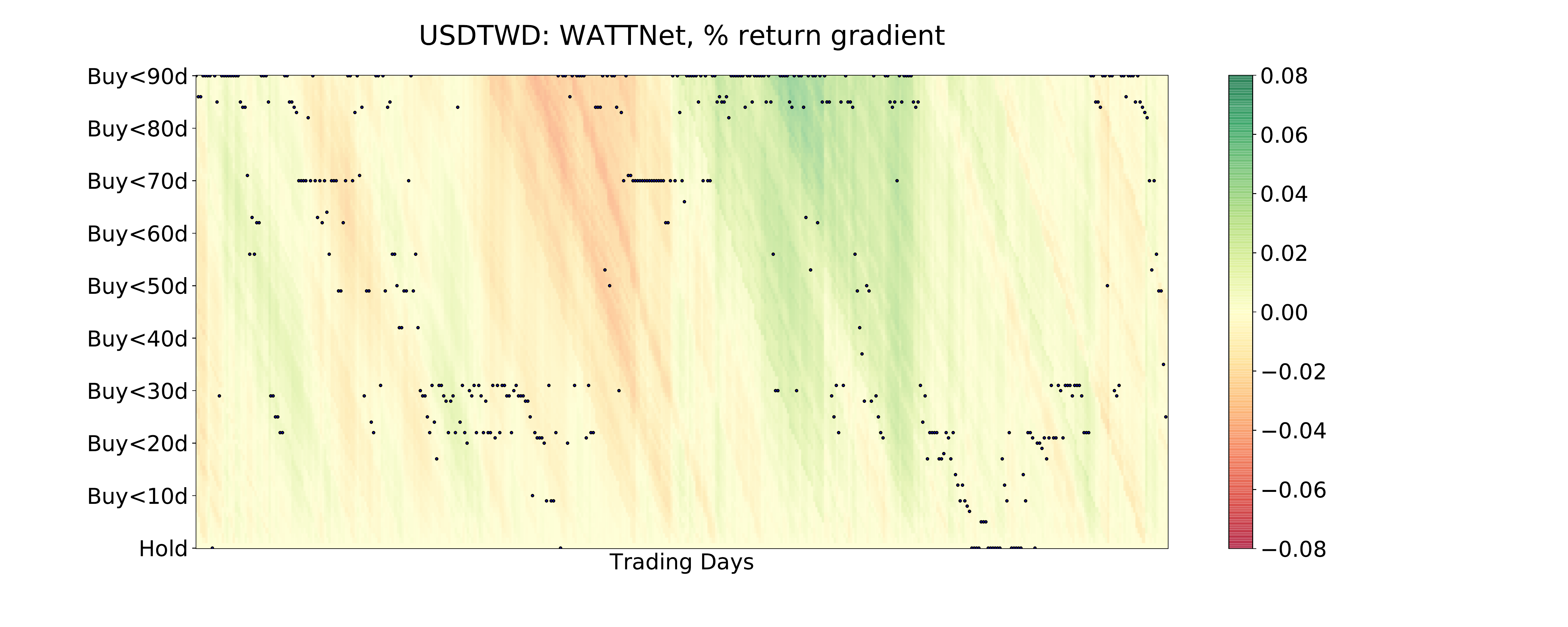}
    \label{fig:umap}
    \caption{USDTWD tenor actions}
\end{figure}

\begin{figure*}[!h]
   \centering
    \includegraphics[scale=0.26]{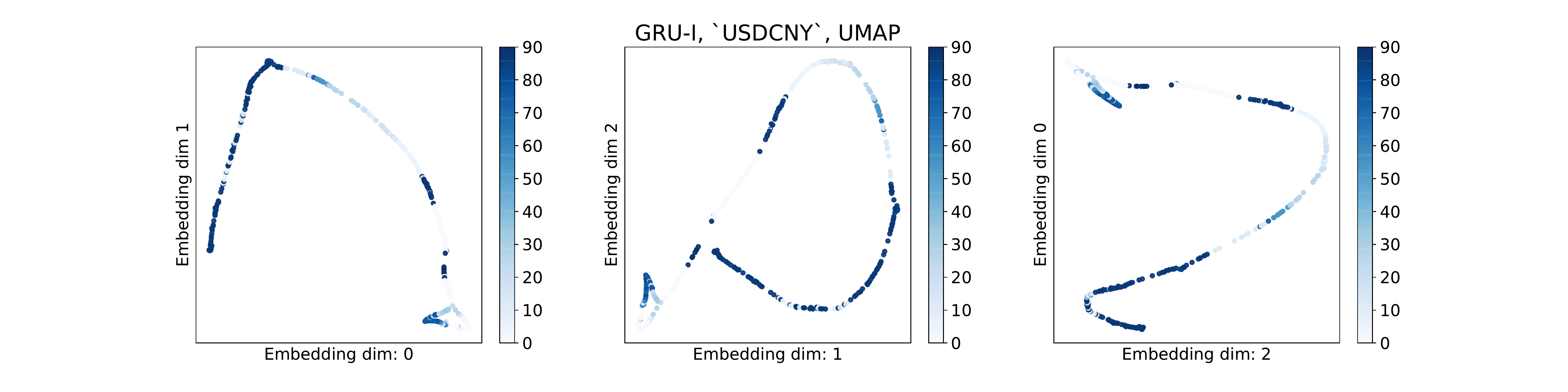}
    \includegraphics[scale=0.26]{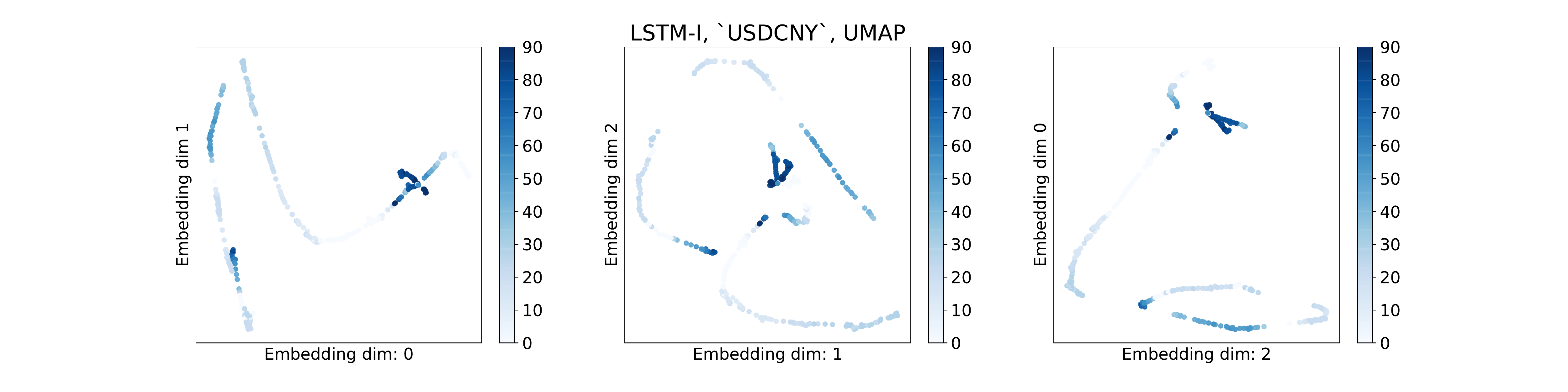}
    \includegraphics[scale=0.26]{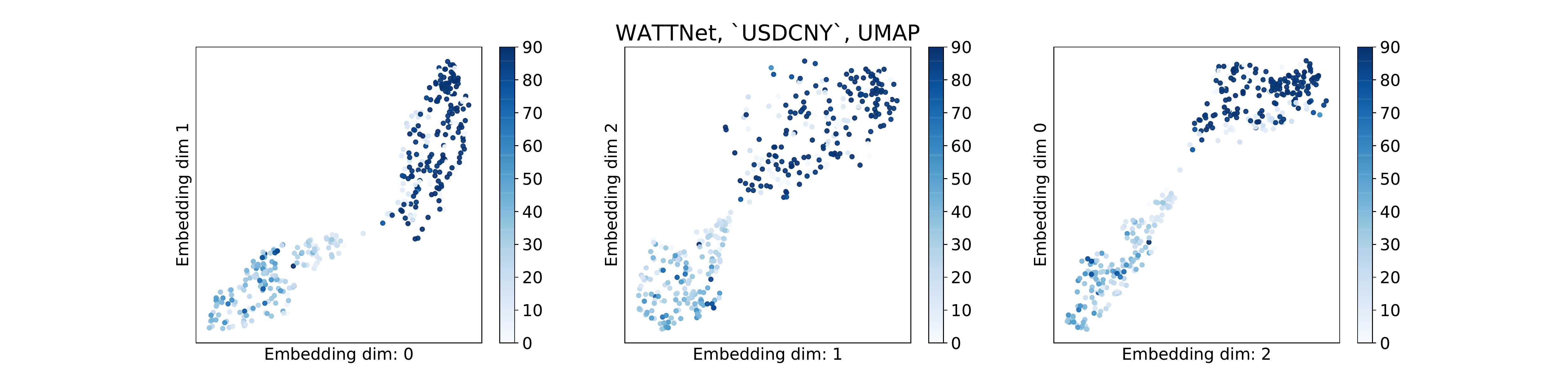}
    \label{fig:umap}
        \caption{UMAP of latent representation - USDCNY}
\end{figure*}
\begin{figure*}[!h]
   \centering
    \includegraphics[scale=0.26]{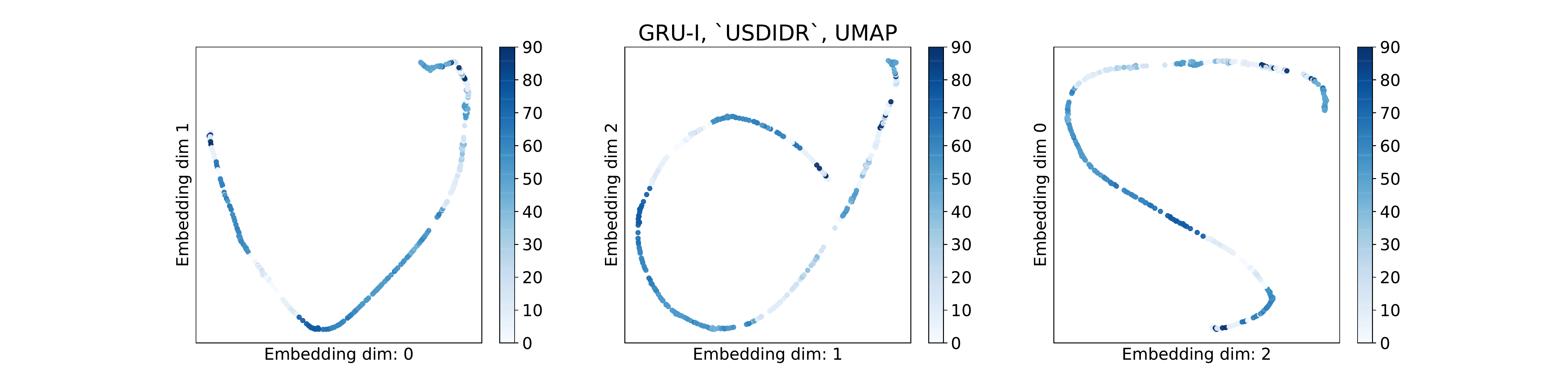}
    \includegraphics[scale=0.26]{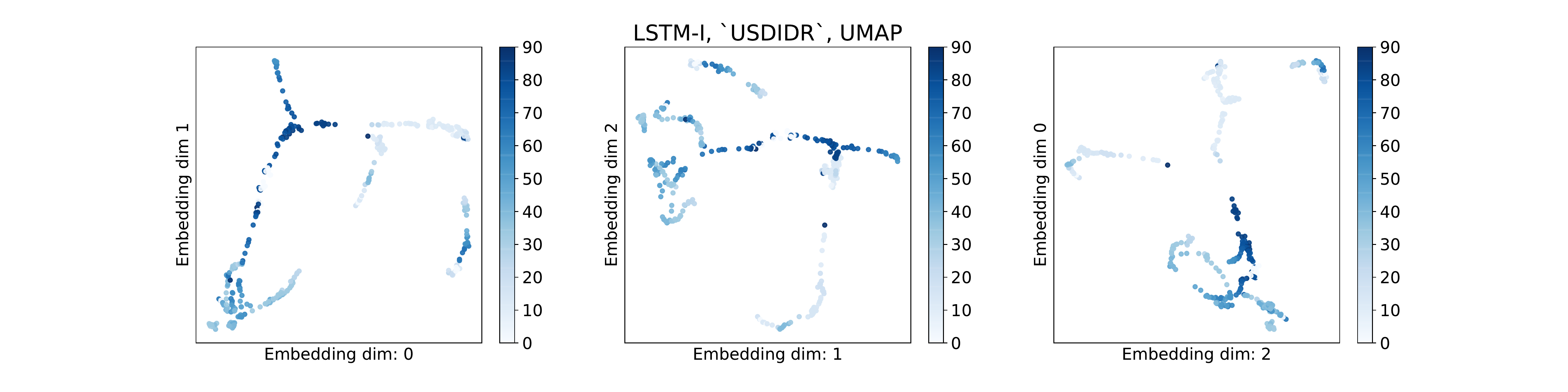}
    \includegraphics[scale=0.26]{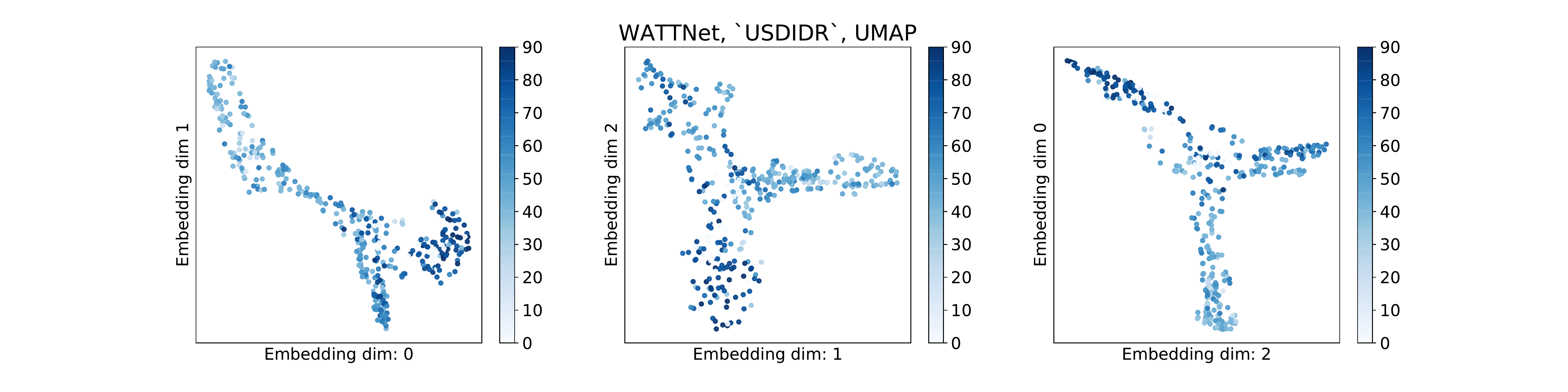}
    \label{fig:umap}
    \caption{UMAP of latent representation - USDIDR}
\end{figure*}
\begin{figure*}[!h]
   \centering
    \includegraphics[scale=0.26]{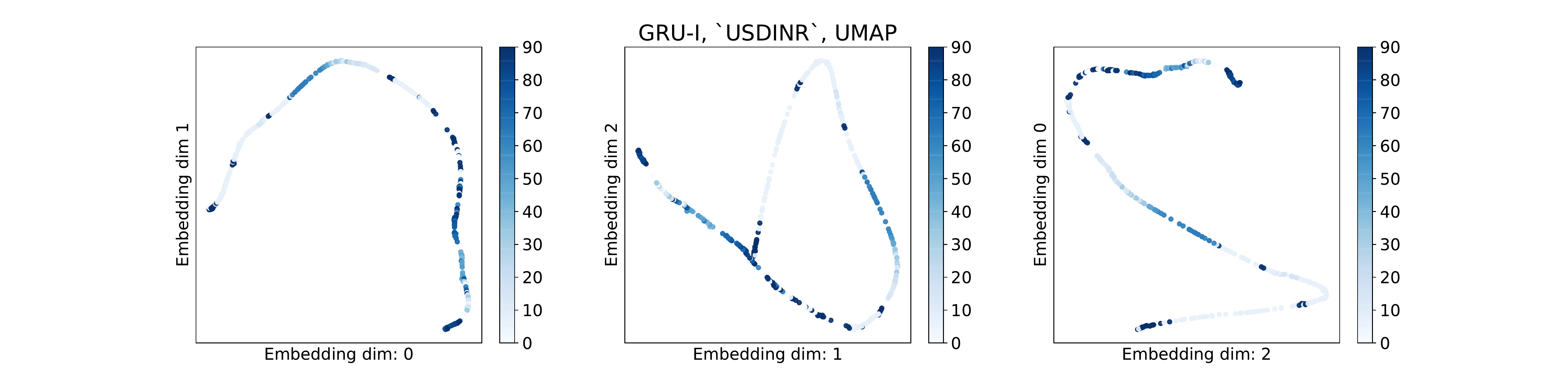}
    \includegraphics[scale=0.26]{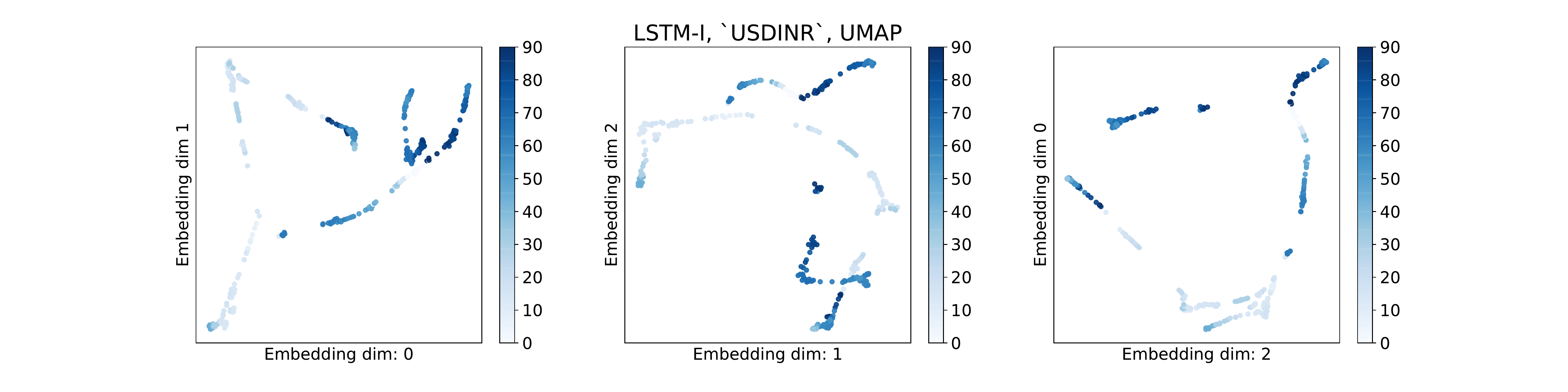}
    \includegraphics[scale=0.26]{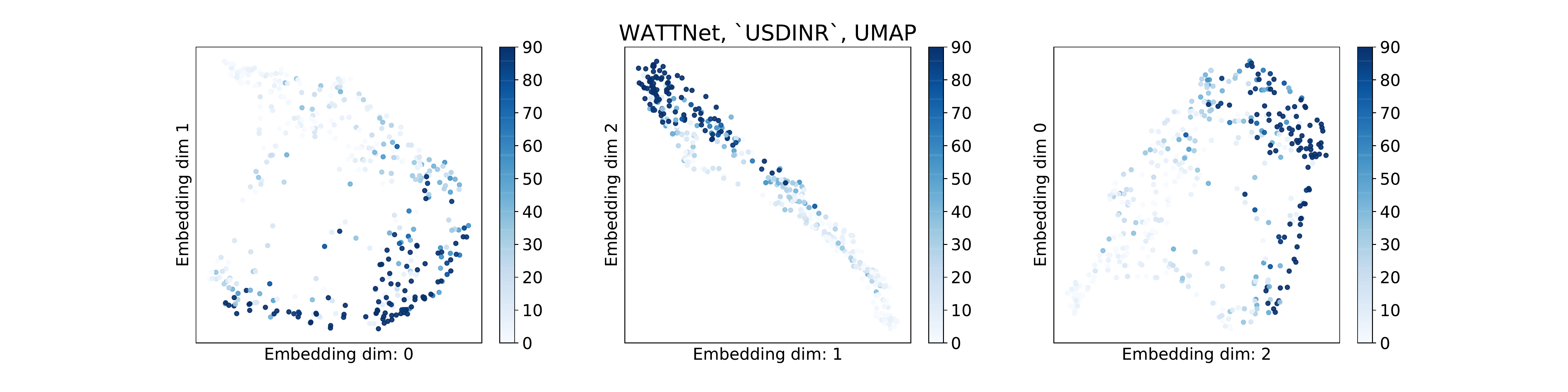}
    \label{fig:umap}
    \caption{UMAP of latent representation - USDINR}
\end{figure*}
\begin{figure*}[!h]
   \centering
    \includegraphics[scale=0.26]{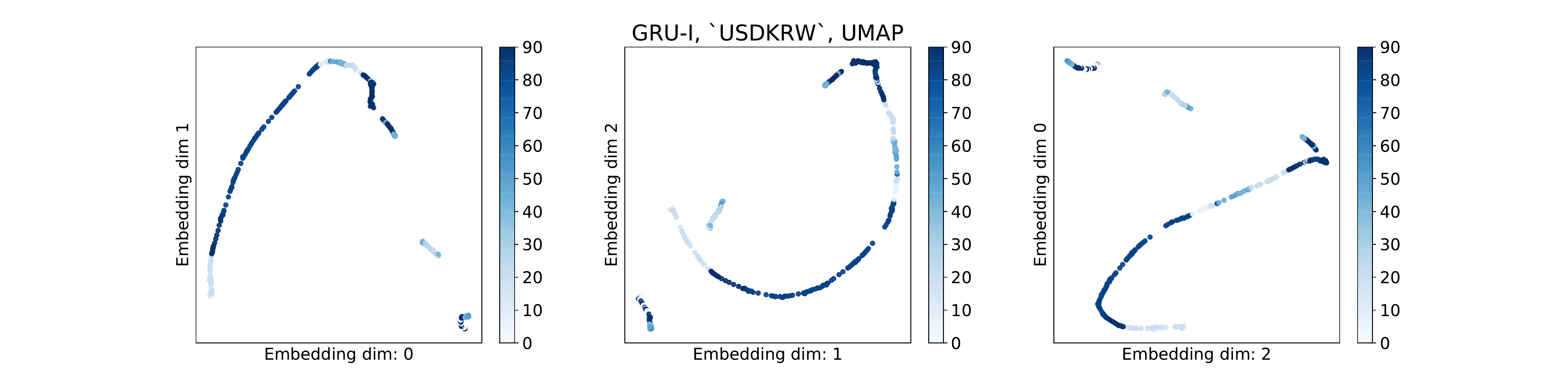}
    \includegraphics[scale=0.26]{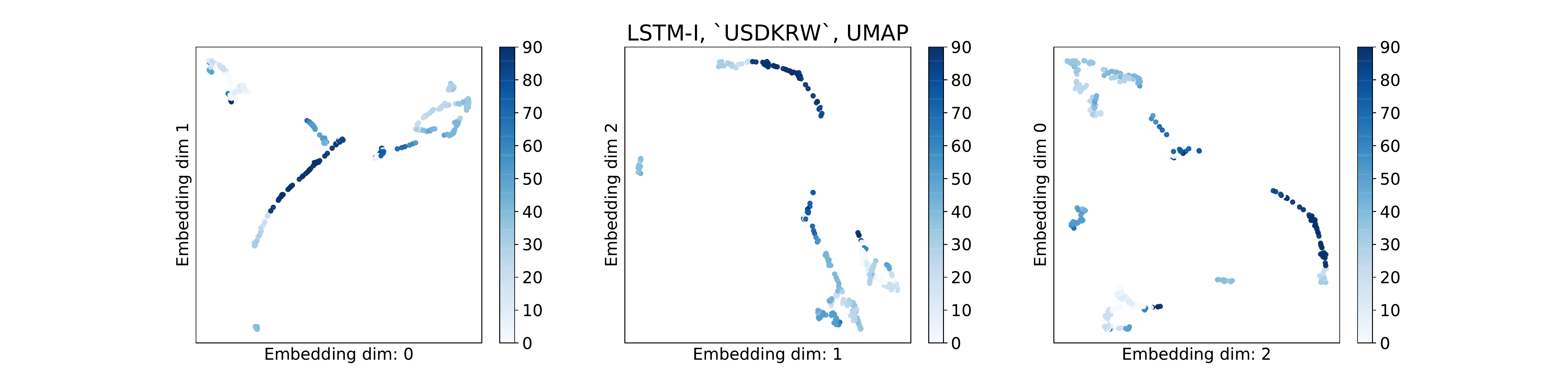}
    \includegraphics[scale=0.26]{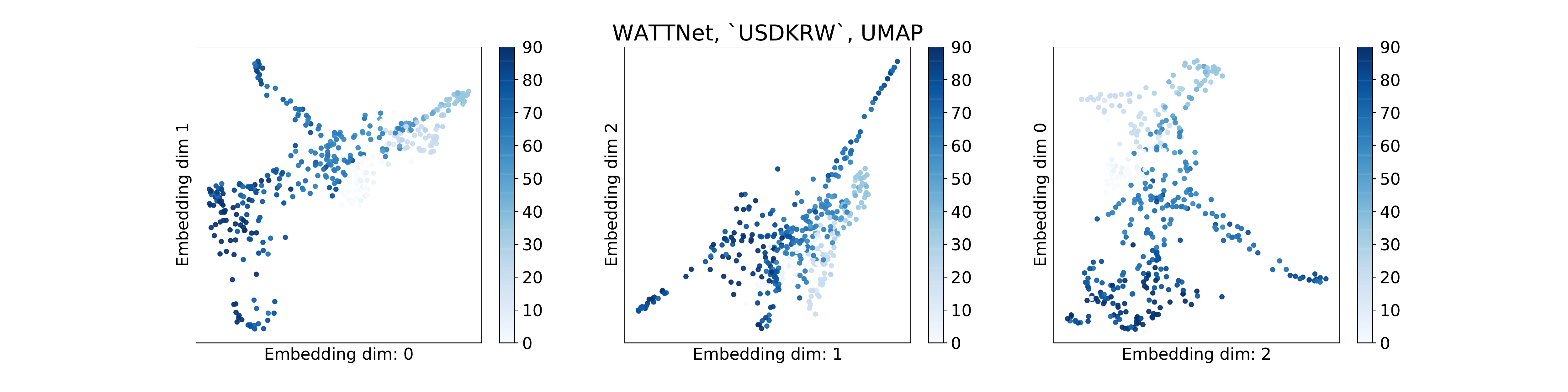}
    \label{fig:umap}
    \caption{UMAP of latent representation - USDKRW}
\end{figure*}
\begin{figure*}[!h]
   \centering
    \includegraphics[scale=0.26]{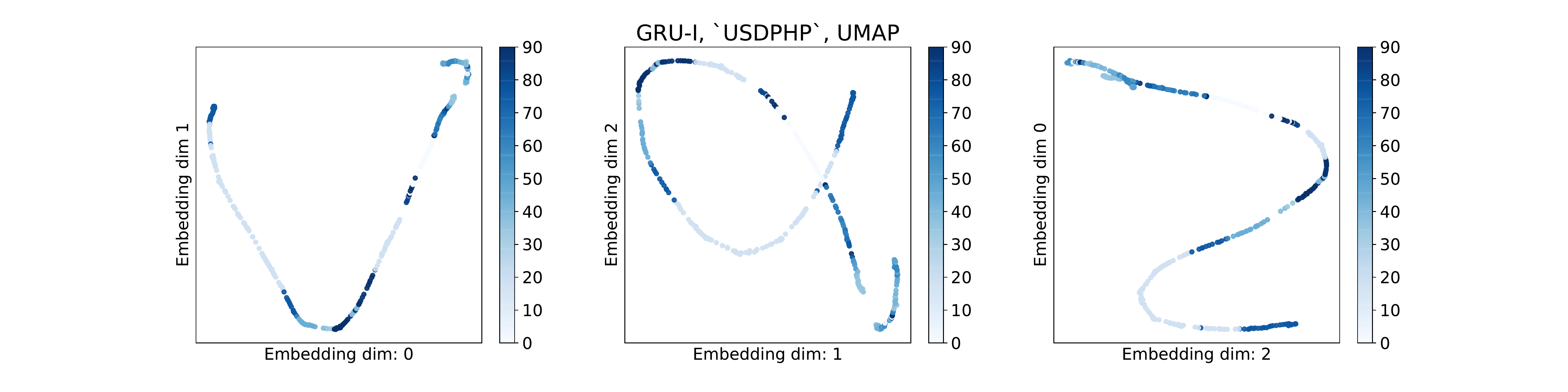}
    \includegraphics[scale=0.26]{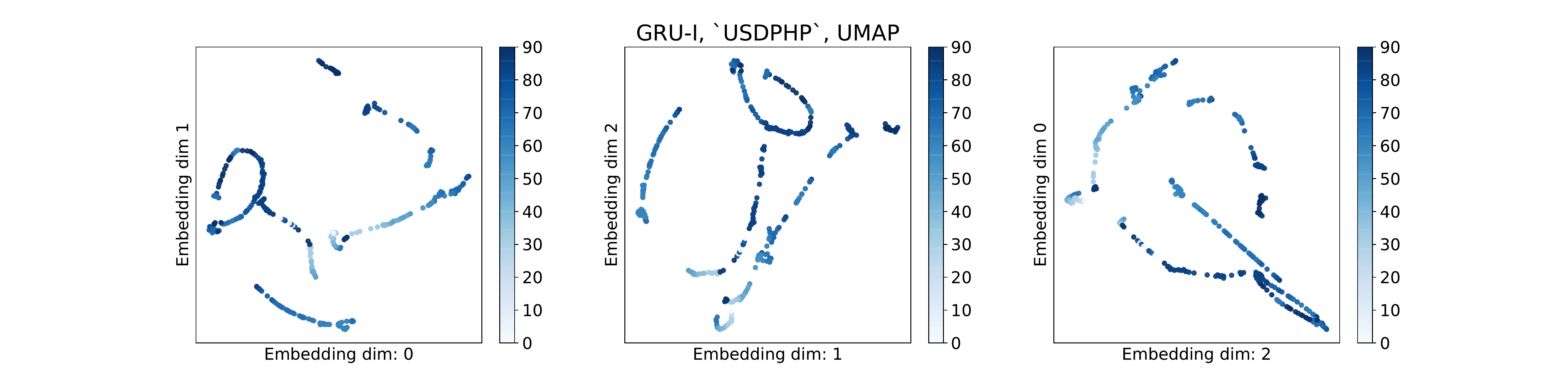}
    \includegraphics[scale=0.26]{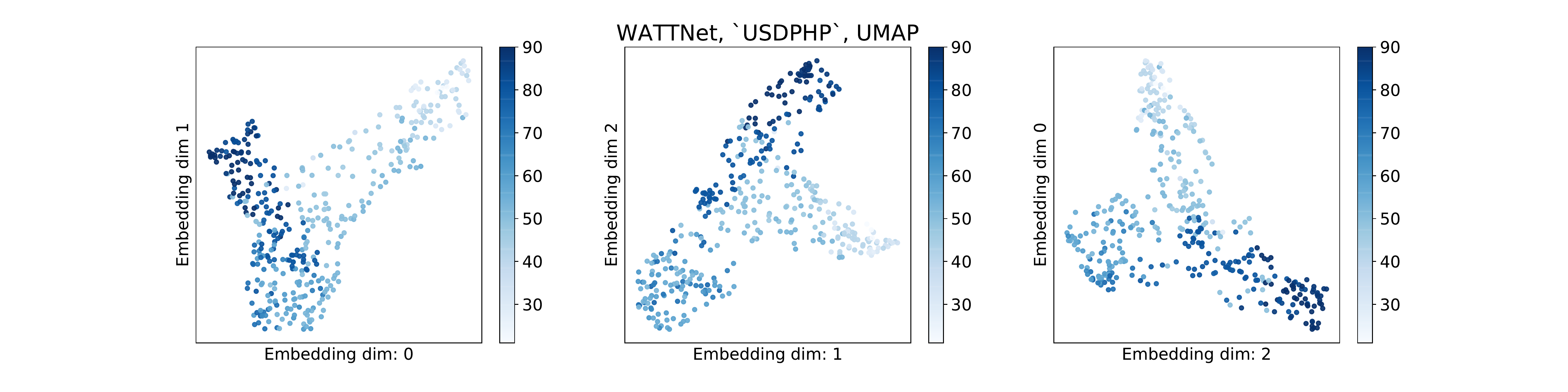}
    \label{fig:umap}
    \caption{UMAP of latent representation - USDPHP}
\end{figure*}
\begin{figure*}[!h]
   \centering
    \includegraphics[scale=0.26]{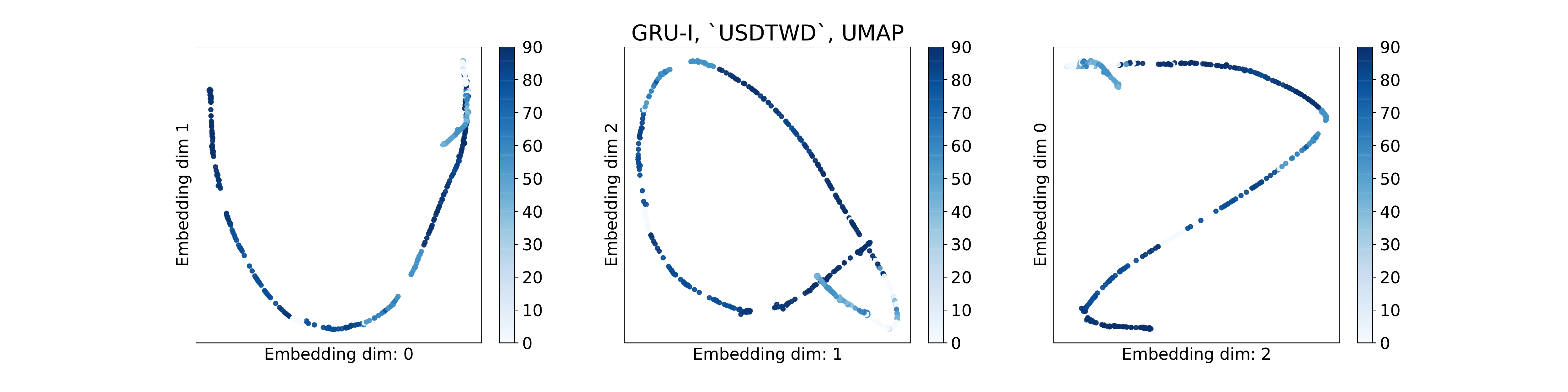}
    \includegraphics[scale=0.26]{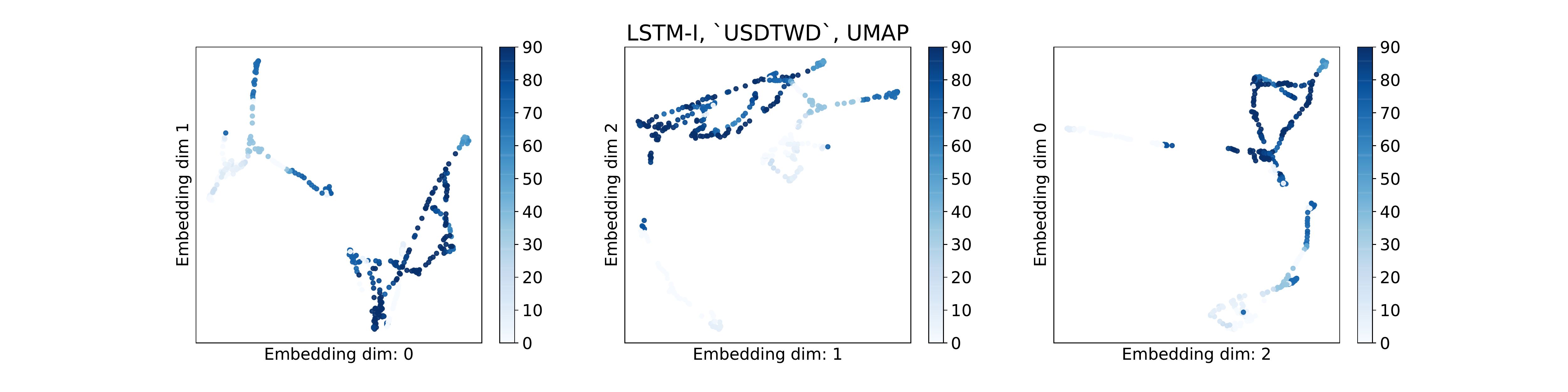}
    \includegraphics[scale=0.26]{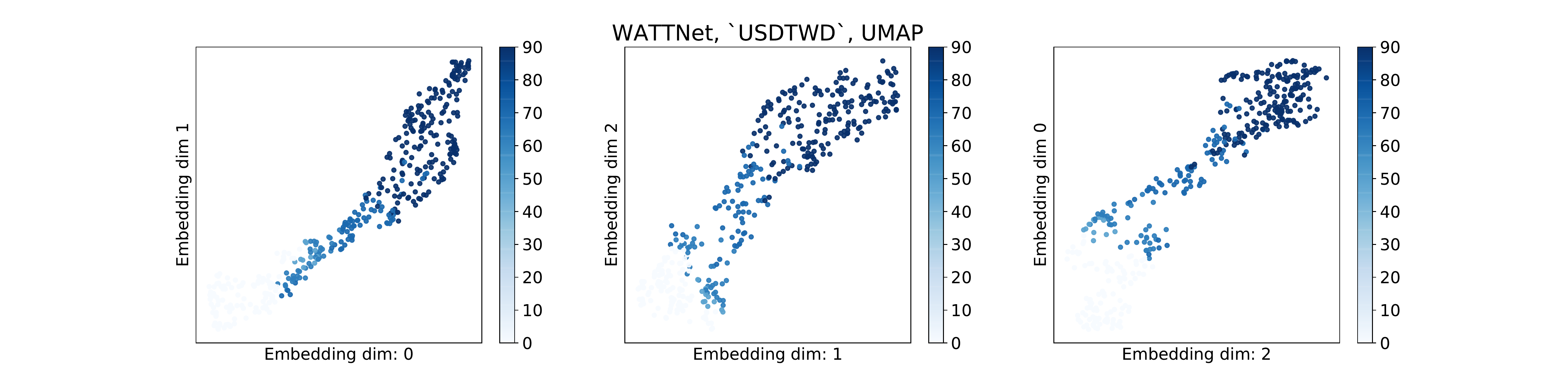}
    \label{fig:umap}
    \caption{UMAP of latent representation - USDTWD}
\end{figure*}

\end{document}